\documentclass{article}

\usepackage[final]{neurips_2023}

\usepackage[utf8]{inputenc} 
\usepackage[T1]{fontenc}    
\usepackage{url}            
\usepackage{booktabs}       
\usepackage{amsfonts}       
\usepackage{nicefrac}       
\usepackage{microtype}      
\usepackage{xcolor}         

\usepackage{graphicx}
\definecolor{mydarkblue}{rgb}{0,0.08,0.45}
\usepackage[colorlinks=true, linkcolor=mydarkblue, urlcolor=mydarkblue, citecolor=mydarkblue, filecolor=mydarkblue]{hyperref}
\usepackage{subfigure,caption,subcaption,wrapfig}
\usepackage{tcolorbox}
\usepackage{wrapfig}

\title{Colour versus Shape Goal Misgeneralization in Reinforcement Learning: A Case Study}

\author{%
  Karolis Ramanauskas\\
  Department of Computer Science\\
  University of Bath\\
  Bath, United Kingdom\\
  \texttt{kr711@bath.ac.uk} \\
  \And
  {\"O}zg{\"u}r~{\c{S}}im{\c{s}}ek \\
  Department of Computer Science\\
  University of Bath \\
  Bath, United Kingdom \\
  \texttt{o.simsek@bath.ac.uk} \\
}

\begin{document}

\maketitle
\newcommand{\kr}[1]{\textcolor{blue}{#1}}

\newcommand{\oz}[1]{\textcolor{green}{#1}}

\begin{abstract}

We explore colour versus shape goal misgeneralization originally demonstrated by \citet{di2022goal} in the Procgen Maze environment, where, given an ambiguous choice, the agents seem to prefer generalization based on colour rather than shape.
After training over 1,000 agents in a simplified version of the environment and evaluating them on over 10 million episodes, we conclude that the behaviour can be attributed to the agents learning to detect the goal object through a specific colour channel.
This choice is arbitrary.
Additionally, we show how, due to underspecification, the preferences can change when retraining the agents using exactly the same procedure except for using a different random seed for the training run.
Finally, we demonstrate the existence of outliers in out-of-distribution behaviour based on training random seed alone.
\end{abstract}

\section{Introduction}

Goal misgeneralization in reinforcement learning is a type of failure observed when an agent is deployed in an environment that is different from the environment in which it was trained.  
Specifically, goal misgeneralization is said to occur when an agent pursues a goal competently but that goal is not the one intended by the system designer~\citep{di2022goal, shah2022goal}.
For example, an agent trained to collect gold coins may fail to collect them when the coins are placed differently in the environment but still efficiently navigate to certain locations while successfully avoiding enemies.   
Goal misgeneralization is highly relevant to AI safety. Advanced AI systems that competently pursue unintended goals can pose significant risks to humanity~\citep{russell2019human}.

We focus on colour versus shape misgeneralization, following an experiment conducted by \cite{di2022goal}, where the agent was trained (from pixel observations) in different mazes to reach a yellow, line-shaped object.
When the agent was tested in mazes that did not contain a yellow line but instead contained both a red line and a yellow gem, the agent generally pursued the yellow gem rather than the red line.
It may be argued that this behaviour is expected given that colour is a feature directly observed by the agent while shape is a latent one that has to be learned~\citep{scimeca2021shortcut}.
Is that what happened? Or are there other factors at play? Should we expect more complex AI systems, such as self-driving cars, to also exhibit such behaviour, perhaps by acting differently when driving in an area with red school buses instead of yellow ones?

Our contributions are as follows:
\begin{itemize}
    \item We reproduce the goal misgeneralization observed by \cite{di2022goal} in a simpler environment  (\autoref{sec:simplifying-the-environment}) that allows  extensive behavioural experimentation.
    \item We show that a change as simple as the random seed used in training the agent can change the preference of the agent from colour to shape (\autoref{sec:the-first-100-training-runs}).
    \item We show that this behaviour stems from the arbitrary and underspecified choice of RGB colour channel through which the agent learns to detect the goal object (\autoref{sec:colour-vs-shape}).
    \item We demonstrate the existence of outliers, on a scale of 1 in 500, in the ways that agents learn to solve a task (\autoref{sec:outliers}). We discuss the implications of such outliers for large-scale AI models.
\end{itemize}

\section{Methodology}

\paragraph{Original environment.}
We started by trying to reproduce the results by \cite{di2022goal} by using the same code\footnote{\url{https://github.com/jbkjr/objective-robustness-failures}} and hyper-parameters.
The agent uses PPO~\citep{schulman2017proximal} and the IMPALA network architecture~\citep{espeholt2018impala} with no recurrent elements.
The environment is based on the Procgen Maze~\citep{cobbe2020leveraging} but is slightly modified by replacing the original goal object (cheese) by a yellow line.
The only reward signal is $+$10 received upon reaching the goal, which terminates the episode. 
An episode terminates also after 500 steps even if the goal has not been reached.
The maze layout, maze size (ranging from 3$\times$3 to 25$\times$25), goal position, and background textures are randomized throughout training.
\autoref{fig:downsampling} shows some examples.
The different instances of the environment are called \textit{levels}.
The actions are the four cardinal direction moves.\footnote{Procgen environments have 15 actions but the remaining 11 actions do nothing in Maze.
The agents quickly learn to use only the four active actions.}

After training, the agent is tested in unseen levels, where the yellow line is replaced with two new objects in random locations -- a yellow gem and a red line.
The goal is now ambiguous. And yet the agent reliably goes for the yellow gem, revealing that it learned to detect the object by colour rather than shape.
This was observed in multiple training runs using different random seeds.
If the designer of the system intended the goal to be a line, goal misgeneralization occurred.

\paragraph{Simplifying the environment.}
\label{sec:simplifying-the-environment}

\begin{wrapfigure}{r}{0.48\textwidth}
    \centering
    \subfigure{
        \includegraphics[width=0.22\textwidth]{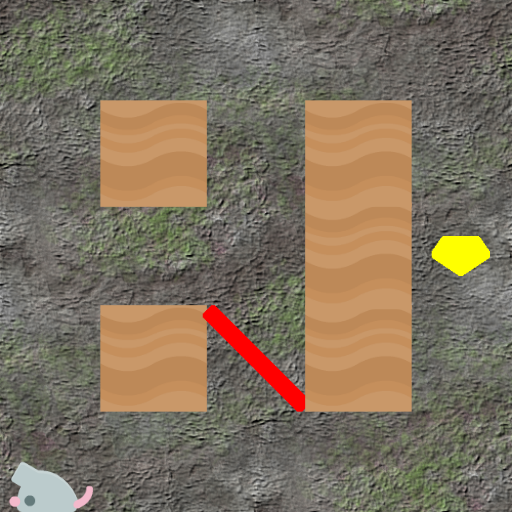}
    }
    \subfigure{
        \includegraphics[width=0.22\textwidth]{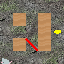}
    }

    \captionsetup{belowskip=-0.3cm}
    \caption{The simplified Maze environment in human and agent view. The size is always 5$\times$5; there are no dead ends; goal assets have been redesigned. 
    The shapes are visible in agent view.}
    \label{fig:simple-maze}
\end{wrapfigure}

Observations are downsampled from the 512$\times$512 pixel human view to 64$\times$64 before being given to the agent.
After some experimentation, we noticed that the downsampling procedure made line objects invisible about 50\% of the time but the gem objects only about 20\% of the time.
Additionally, the yellow line and the yellow gem are slightly different shades of yellow.
We present full details and images in  \autoref{appendix:disappearing-objects}.
We simplified the environment as follows to eliminate these differences between the goal objects (see \autoref{fig:simple-maze}):
\begin{itemize}
\item We created new line and gem assets of all eight combinations of RGB colours in their pure form, i.e. either 0 or 255 for each of the three channels, as shown in \autoref{appendix:original-and-our-environment-assets}.
\item We always used a 5$\times$5 maze with no outside padding.
\item We revised the maze generation algorithm to not have dead ends. This prevents cases where reaching one object without going over the other object is impossible.
\item We make either object in the test environment end the episode but with different rewards.
\end{itemize}

The simplified environment has the welcome side effect that a single training run takes about 40 GPU minutes instead of 40 GPU hours, mainly due to fewer agent-environment interactions needed to train the agent.
A speed up of 60 times helps run more comprehensive experiments on these agents.
The last two changes allow for the systematic testing of many trained agents.

With these changes, we obtained the same result as those reported by \cite{di2022goal} on the first agent that we trained. Specifically, an agent we trained to seek out a yellow line preferred a yellow gem to a red line in the test environment. We then performed a large number of further experiments, which we explain in the next section.\footnote{Code available here: \url{https://github.com/KarolisRam/colour-shape-goal-misgeneralization}}

\paragraph{Measuring preferences and capabilities.}

We train 100 agents using different random seeds. This impacts the weight initializations of the neural networks and the training environment levels. Each agent is tested on the same set of 1,000 held-out levels. Most results are shown as scatter plots of preferences and capabilities because goal misgeneralization is about the interaction between preferences and capabilities.
We care the most about cases where different preferences can appear while capabilities remain intact.
For example, a training procedure that produces two agents that both act capably but differ in their out-of-distribution goals would be of considerable concern if observed in advanced AI systems.

To measure capabilities, we use the mean episode length. Even an agent acting randomly usually gets to a goal object before the episode times out. On average, a trained agent takes 4 steps to reach a goal while a random agent takes 96 steps. In all plots, the axis for mean episode length ranges between 0 and 100, representing the range between a competent agent and a random one.
Any marks appearing on the border show an episode length of 100 or more, indicating performance worse than the one obtained by acting randomly.\footnote{The mean is not a perfect measure of capabilities. For example, some agents are fully capable on more than half of the test levels while getting stuck in the remaining levels. This could give a mean of 200 and a median of 4. We observed that this rarely happens in practice.}

We measure preferences by the proportion of test runs in which the agent picked up a particular object before the other one. We established a baseline for preferences as follows. 
When an agent is competently navigating to a specific object, it may step on the other object along the way, without seeking to do so. We therefore tested an agent in an environment with a yellow line and an invisible object,  observing that the agent stepped on the invisible object approximately 20\% of the time on its way to the yellow line. Thus, we consider picking an object 80\% of the time as the baseline for the full preference for that object.

\section{Results}

\subsection{The first 100 training runs}
\label{sec:the-first-100-training-runs}

\begin{figure}[t]
  \centering
  \subfigure{\includegraphics[width=0.32\textwidth]{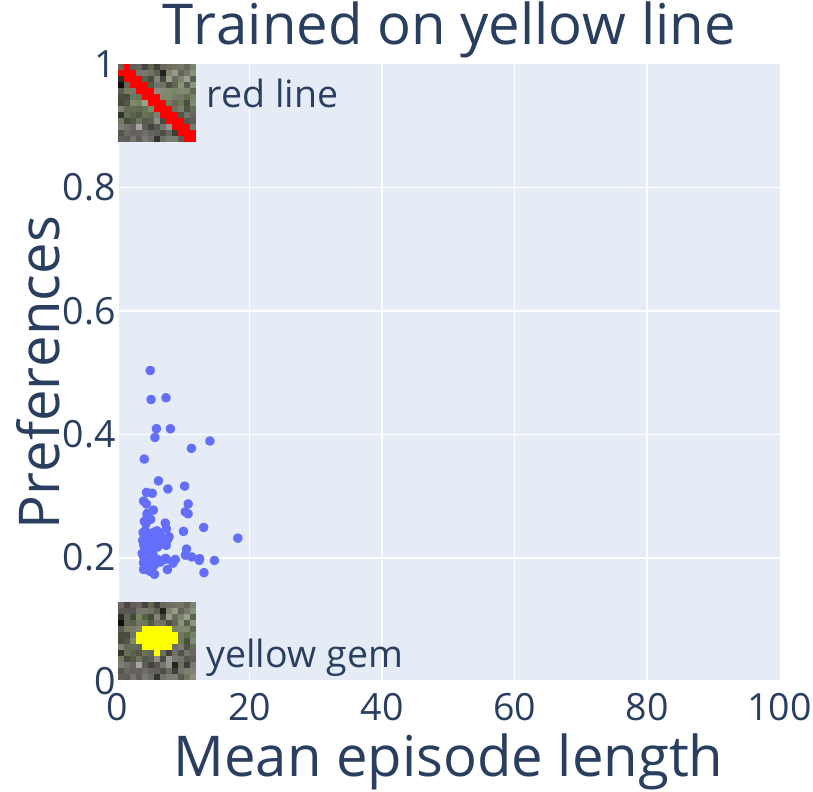}}
  \subfigure{\includegraphics[width=0.32\textwidth]{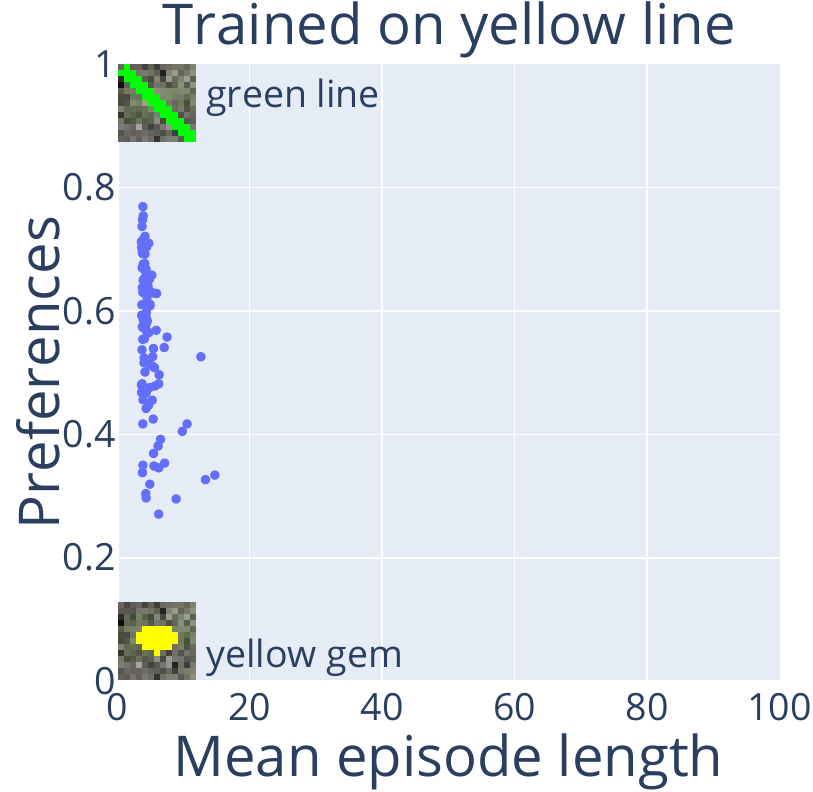}}
  \subfigure{\includegraphics[width=0.32\textwidth]{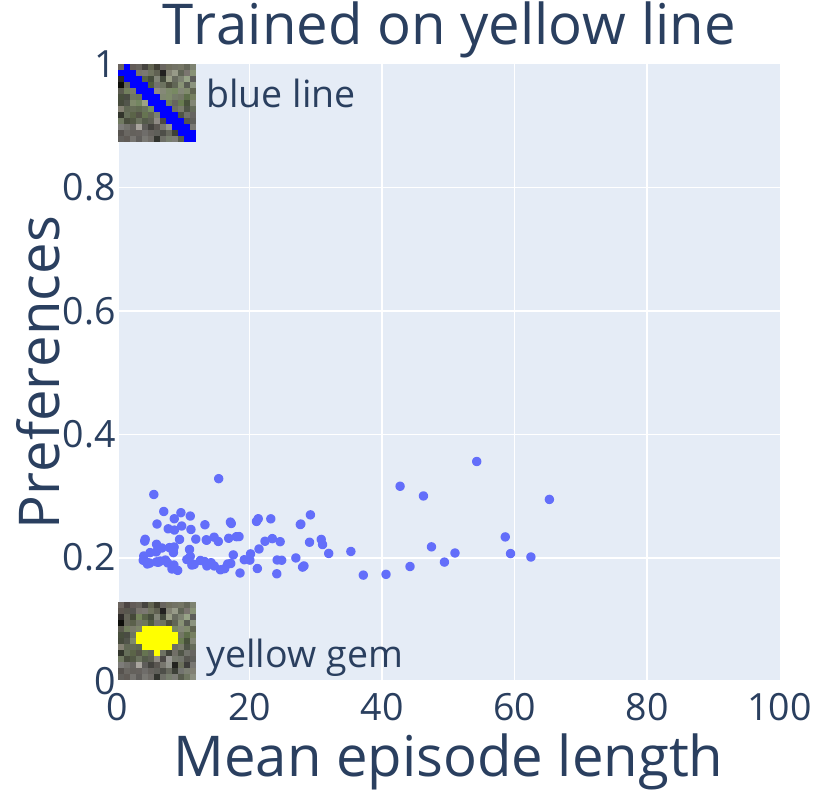}}

  \subfigure{\includegraphics[width=0.32\textwidth]{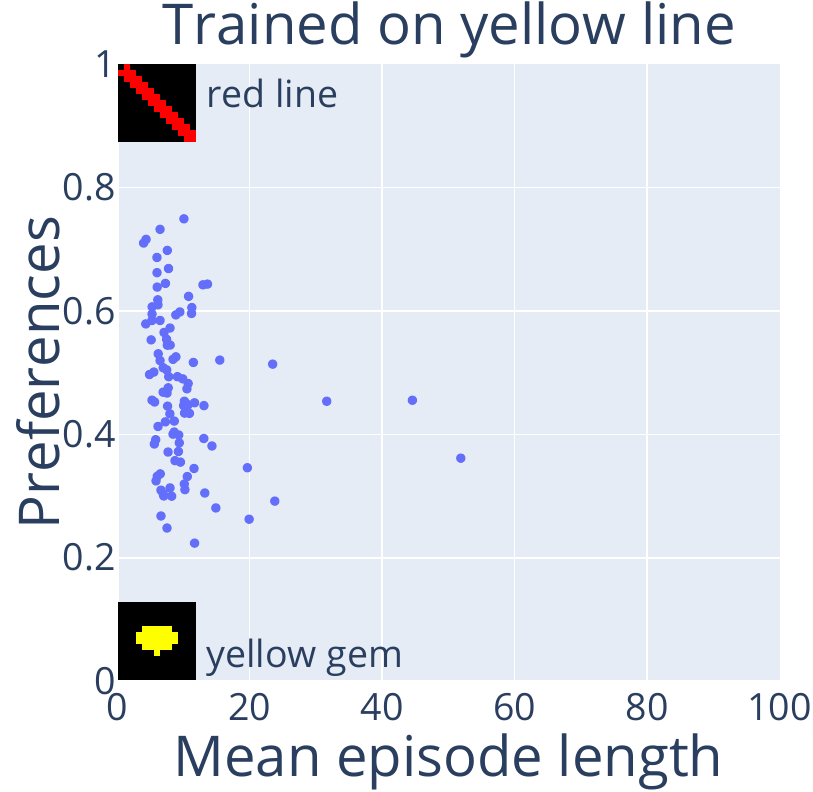}}
  \subfigure{\includegraphics[width=0.32\textwidth]{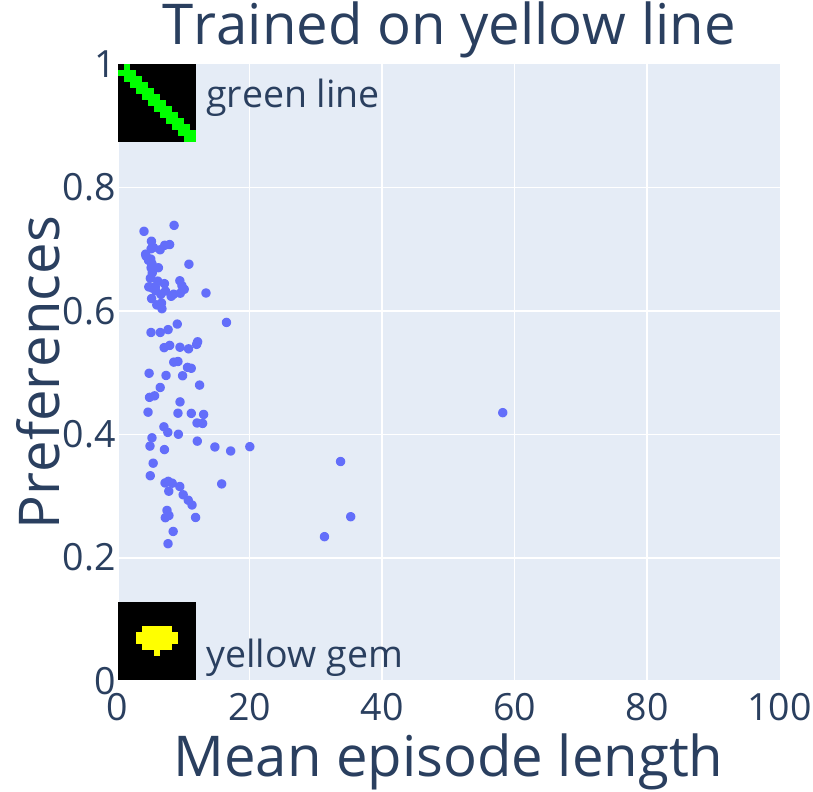}}
  \subfigure{\includegraphics[width=0.32\textwidth]{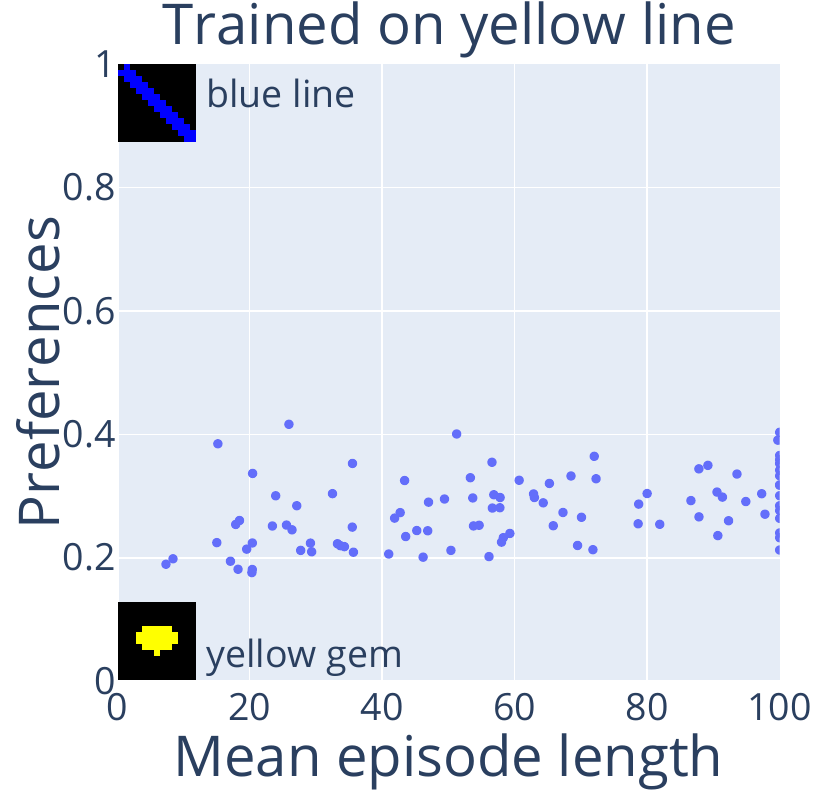}}
      \captionsetup{belowskip=-0.3cm}
  \caption{The top row shows the results of training 100 agents to reach a yellow line on a textured background and then testing them in an environment with the red/green/blue line versus the yellow gem.
  The bottom row shows the same after replacing the textures with black backgrounds during both training and testing.
  Each of the 100 dots in a plot represents a single trained agent. Only the random seed is changed between runs.
  The vertical axis shows the fraction of time the agent chooses the first object.
  The horizontal axis shows a proxy for retained capabilities -- mean episode lengths.}
  \label{fig:train-yellow-line-test-red-green-blue-lines-vs-yellow-gem}
\end{figure}

After confirming that goal misgeneralization happens in our simplified Maze environment, we ran this same training experiment 100 times, varying only the random seed used to initialize the weights of the network and the environment levels. The goal in these experiments is a yellow line. 
We trained each agent for 10 million steps to ensure all of them converged.
We then tested each trained agent on the same set of 1,000 test levels and measured the preference for the red line versus the yellow gem and the mean episode length.
We show the results as a scatter plot on the top left of \autoref{fig:train-yellow-line-test-red-green-blue-lines-vs-yellow-gem}.
We see that most agents keep their capabilities and most strongly prefer the yellow gem.
However, we see several agents having no preference for either object yet keeping their capabilities intact.
 
We ran two further experiments, where we replaced the red line first with a green line, and then with a blue line.
The results are shown on the top row of \autoref{fig:train-yellow-line-test-red-green-blue-lines-vs-yellow-gem}.
In the case of the green line, agents mostly keep their capabilities, but many agents prefer the green line to the yellow gem.
However, other agents still preferred the yellow gem to the green line.
Why did a different colour induce a preference for the shape?
Our next clue is the capability loss in some agents when the line was blue.

Agents do not see things in the same way that humans do.
Image observations usually come as three 2D matrices of numbers between 0 and 255, representing the three colour channels: red, green and blue.
Yellow is a combination of red and green colours and a lack of blue.
This is still somewhat similar to human trichromacy~\citep{horiguchi2013human}, but not a perfect analogy, as humans do not have conscious access to the three separate colours.
Humans also cannot distinguish monochromatic yellow from yellow produced by mixing red and green light~\citep{chittka2022mind}.\footnote{An analogy in music would be hearing note D exactly like notes C and E played together, which we do not. In the case of the agents, they would hear D (yellow) as C + E (red + green), and not as a single tone (colour).}
A closer analogy might be the more separate red and yellow fields in the retina of the pigeon, which are used to view objects at different distances~\citep{bloch1983specialization}.
The agent can detect the line using any of the three channels.
We show what the different lines look like to the agent in \autoref{fig:rgb-splits}.

The question remains of why the red line of the original experiment did not exhibit the same range of behaviours as the green line.
We believe this is because the different backgrounds are not uniformly distributed among the three colour channels, which we show in detail in \autoref{appendix:procgen-maze-background-texture-analysis}.
We hypothesize that in the case of the green line versus the yellow gem, the agents that learned to detect the yellow line primarily via the green channel prefer the green line.
In contrast, the agents that used the red channel to detect the yellow line preferred the yellow gem.
If this is true, the goal misgeneralization in this case happens mostly through colour.

\subsection{Removing background textures}

\begin{figure}[t]
    \centering
    \subfigure{
        \includegraphics[width=0.45\textwidth]{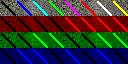}
    }
    \subfigure{
        \includegraphics[width=0.45\textwidth]{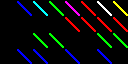}
    }
    \captionsetup{belowskip=-0.3cm}
    \caption{The top row shows how humans perceive the lines. The bottom three rows show how the agents see them in three channels: red, green and blue.
    }
    \label{fig:rgb-splits}
\end{figure}

To show an even simpler case of goal misgeneralization, we changed the background to pure black.
The black line is entirely invisible in this case while the yellow line is visible only in the red and green channels. The lines on a black background and their RGB splits are shown in \autoref{fig:rgb-splits}.

We train 100 new agents and test each one on 1,000 levels in 14 different experimental setups, all of which are shown in \autoref{fig:train-yellow-line-black-background-01} and \autoref{fig:train-yellow-line-black-background-02} in \autoref{appendix:all-results}.
Here we focus on the most interesting results.
The red and the green line versus yellow gem look much more similar than in the experiments with textures, as shown in the first two plots on the bottom row of \autoref{fig:train-yellow-line-test-red-green-blue-lines-vs-yellow-gem}.
This gives some credence to the hypothesis that backgrounds were causing that difference.
It also shows how a different type of goal misgeneralization can occur by simply retraining the agent.

\begin{figure}[t]
    \centering
    \subfigure{
        \includegraphics[width=0.47\textwidth]{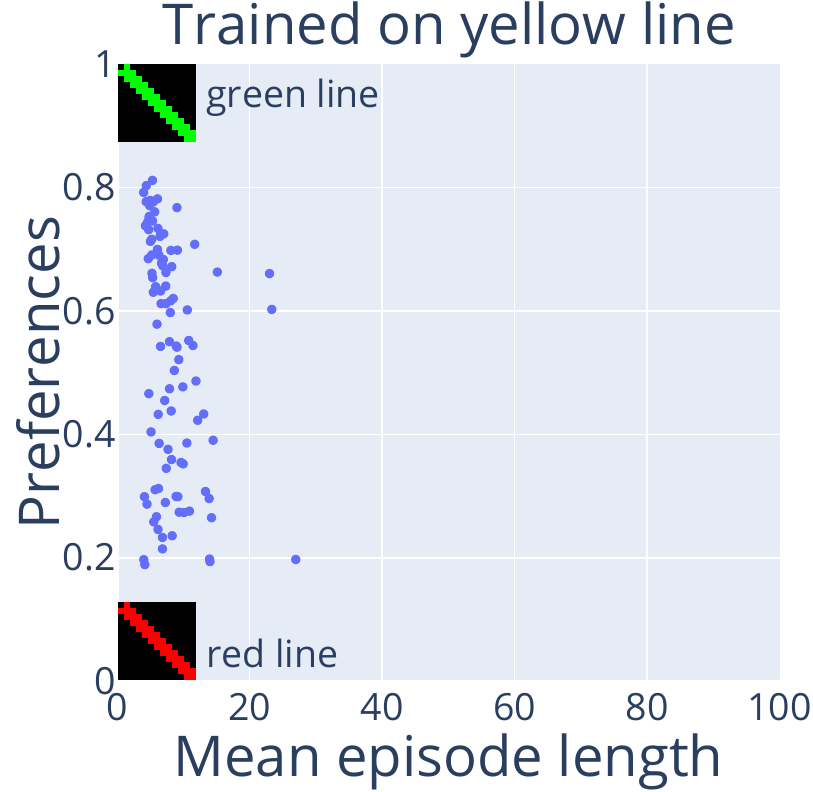}
    }
    \subfigure{
        \includegraphics[width=0.49\textwidth]{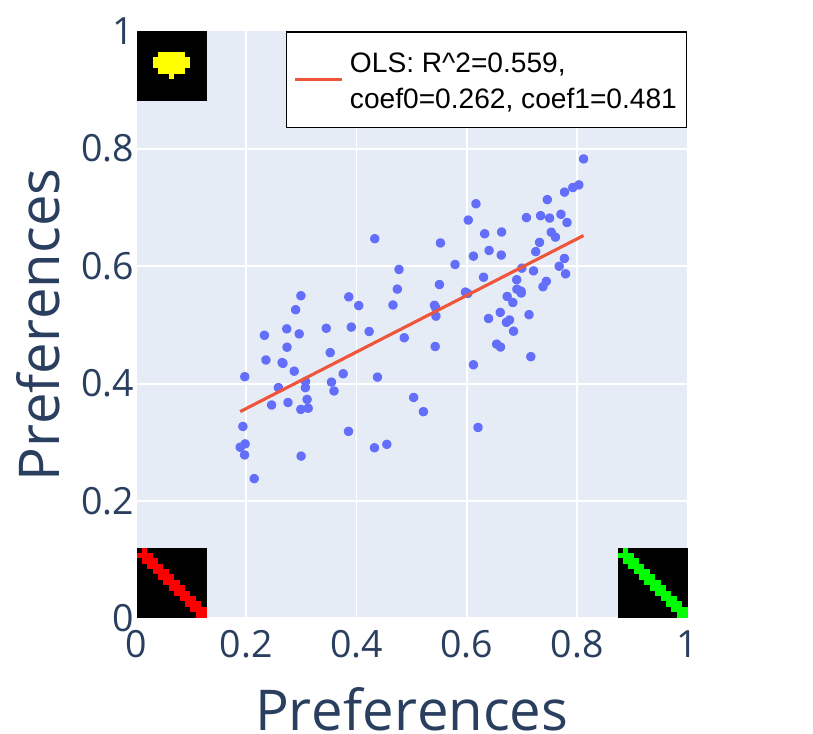}
    }

    \caption{Left -- testing on the green line versus the red line. 
    Right -- OLS regression of that preference as an independent variable versus the original set-up of the yellow gem versus the red line as the dependent variable.
    A clear correlation exists, indicating that the seeming preference for shape comes from the arbitrary choice of colour channel through which to detect a goal object.}
    \label{fig:red-line-green-line-and-OLS}
\end{figure}

Next, we test the agents on the red line versus the green line and show a roughly even split between agents preferring each line, as shown in \autoref{fig:red-line-green-line-and-OLS}.
Many agents do not show a strong preference, indicating learning through both red and green channels.
One interesting observation is that the agents that prefer the red line show some capability loss, while the ones that prefer the green line do not.
This asymmetry may be due to the colours of the maze walls or maybe even the mouse.

\paragraph{Colour versus shape.}
\label{sec:colour-vs-shape}

Do agents that most strongly prefer the green line versus the red line also prefer the yellow gem versus the red line?
If so, this further proves goal misgeneralization happening mainly through the agent's arbitrary choice of colour channel to learn through.
We perform an OLS regression on the preferences of each agent and show the results in \autoref{fig:red-line-green-line-and-OLS}.
The relationship is reasonably strong (R\textsuperscript{2} of 0.56) but less robust than we expected.
Something else seems to be happening, especially when considering the outliers on the chart.
We will call agents by the training seeds used in their training runs.
For example, Agent-8894 is the agent closest to the bottom right corner on the right plot in \autoref{fig:red-line-green-line-and-OLS}.
It prefers green line over red line over yellow gem.
It would be informative to understand how exactly this particular agent learned to perform the task.
Perhaps it learned to detect the shape and prefers green colour, but not as strongly as it prefers the line shape.

It might be that these observations are relevant only for toy problems because real-world images rarely have pure single or double-channel colours.
However, it would be interesting to test how different self-driving car models perceive red, yellow, and green colours on a black background.
It might be an important thing to distinguish well while driving.\footnote{This idea is not too far fetched. We have already seen them get confused by stickers on road signs or ghost white lines on the road.}

\subsection{Outliers}
\label{sec:outliers}

After noticing outliers in many of the scatter plots, we decided to investigate them further.
We trained 512 agents to reach a white line on black backgrounds.
The line is visible in all three channels, which adds more possibilities to learn the same underspecified task in different ways.
As usual, only the random seed was changed between the 512 training runs.
We then measured the preferences and capabilities of each agent in 11 different two-object variations, 1,000 levels for each variation.
This produced a dataset with 512 $\times$ 22 data points.
We used several clustering and outlier detection tools and also simple sorting by primary and derivative features of this dataset.
We found various interesting outliers, many of them unique among the 512 agents.
For example, Agent-439 was the only one out of 512 to strongly prefer white gem over the yellow line while being highly capable (bottom left dot in the bottom left plot of \autoref{fig:train-white-line-black-background}).
Meanwhile, Agent-2875 performed worse than a random agent when both lines were single channel (red, green or blue).
It also really liked purple lines.
A longer list of outlier behaviours can be found in \autoref{appendix:list-of-outliers}.

Could similar outliers exist in frontier AI models, and how would we know?
What does this mean for model behaviour attribution, when rare outcomes can happen based on the training random seed alone, that is, luck?
This might even have some parallels in human behaviour, where we sometimes observe extreme deviations from typical patterns, and individuals may receive diagnoses such as OCD, ADHD, or autism to describe their unique characteristics and challenges.
In the Maze agents, the random seed changes the neural network weight initialization and the training levels, which could loosely represent the genes and the environment, both of which interact to produce the outcomes.
We try to detect these conditions as early as possible in humans to be able to intervene if needed.
Should we be trying to look for signs of outlier behaviour in frontier AI models early in their training?

\section{Related work}

\paragraph{Behaviour attribution.}

\citet{hilton2020understanding} use various interpretability techniques to understand the behaviour of Procgen Coinrun agents.
\citet{mcgrath2022acquisition} analyzed the chess concepts learned by AlphaZero~\citep{silver2018general} through behavioural analysis and other methods.
\citet{atrey2019exploratory} show how saliency maps are not enough for attributing agent behaviour and that counterfactual experiments should be performed.

\paragraph{Goal misgeneralization.}

\citet{di2022goal} introduced initial empirical examples, while \citet{shah2022goal} added more, including hypothetical future ones and showed how the problem is different from misspecification.
\citet{turner2023understanding} explored an object versus location misgeneralization in a similar Procgen Maze environment.
Some image classification failures can also be interpreted as goal misgeneralization~\citep{shah2022goal}.
Examples include classifying a wolf as a husky if there is snow in the background ~\citep{ribeiro2016should} or classifying skin lesions as malignant if they were photographed next to a ruler~\citep{narla2018automated}.

\paragraph{Underspecification.}

\citet{d2022underspecification} shows how retraining the same machine learning model using a different random seed can lead to the same test set performance but different out-of-distribution performance.
They do this in multiple domains and input modalities.
We show this in a deep reinforcement learning setting with goal misgeneralization.
\citet{sellam2021multiberts} demonstrate a similar underspecification phenomenon in larger scale language models, including visible outliers on their Winogender bias correlation plot.
This indicates the possibility of different random seed based goal misgeneralization and outlier behaviours in frontier AI models.

\section{Conclusion}
\label{sec:conclusion}

We have shown how part of the colour versus shape goal misgeneralization behaviour in Procgen Maze can be attributed to the RGB colour encoding scheme.
We also showed that this behaviour can change based on the random seed used for training.
These observations raise the question of how far such findings of model behaviour attribution transfer beyond the toy examples.
Since we see underspecification happening in larger models, we expect to also see empirical examples of goal misgeneralization being discovered in such models.
It is possible that outliers in out-of-distribution behaviour exist in large-scale models based on the training random seed alone.
However, it would be infeasible to run such experiments on frontier AI models.
Thus, we should also be careful when making claims about behaviour attribution -- they might be true for that particular checkpoint of the trained model but might not hold anymore if it was simply retrained using a different random seed.

A potential limitation of our work is that the simplified environment might not be diverse enough and no meaningful generalizations are learned.
Specifically, we would expect more complex vision-based deep reinforcement learning agents to show colour versus shape preference, even if it did not happen in the case of Procgen Maze.
In the future, we hope to see more work of this kind -- taking a specific example of interesting behaviour and trying to deeply understand how it happens.

\begin{ack}
This work was supported by the UKRI Centre for Doctoral Training in Accountable, Responsible and Transparent AI (ART-AI) [EP/S023437/1] and the University of Bath. This research made use of Hex, the GPU Cloud in the Department of Computer Science at the University of Bath. We thank the reviewers for the useful feedback. We also thank Rachael Bedford, Dan Beechey, Christoph Bumiller, Sophia Jones, Gabija Juce, Anssi Kanervisto, Stephanie Milani, Owen Parsons and Tom Smith for helpful discussions and reading earlier versions of this work. We thank members of the Bath Reinforcement Learning Lab and the SANDPIT Lab for their feedback.
\end{ack}

\small
\bibliography{references}

\begin{thebibliography}{19}
\providecommand{\natexlab}[1]{#1}
\providecommand{\url}[1]{\texttt{#1}}
\expandafter\ifx\csname urlstyle\endcsname\relax
  \providecommand{\doi}[1]{doi: #1}\else
  \providecommand{\doi}{doi: \begingroup \urlstyle{rm}\Url}\fi

\bibitem[Di~Langosco et~al.(2022)Di~Langosco, Koch, Sharkey, Pfau, and Krueger]{di2022goal}
L.~L. Di~Langosco, J.~Koch, L.~D. Sharkey, J.~Pfau, and D.~Krueger.
\newblock Goal misgeneralization in deep reinforcement learning.
\newblock In \emph{International Conference on Machine Learning}, pages 12004--12019. PMLR, 2022.

\bibitem[Shah et~al.(2022)Shah, Varma, Kumar, Phuong, Krakovna, Uesato, and Kenton]{shah2022goal}
R.~Shah, V.~Varma, R.~Kumar, M.~Phuong, V.~Krakovna, J.~Uesato, and Z.~Kenton.
\newblock Goal misgeneralization: Why correct specifications aren't enough for correct goals.
\newblock \emph{arXiv preprint arXiv:2210.01790}, 2022.

\bibitem[Russell(2019)]{russell2019human}
S.~Russell.
\newblock \emph{Human compatible: Artificial intelligence and the problem of control}.
\newblock Penguin, 2019.

\bibitem[Scimeca et~al.(2021)Scimeca, Oh, Chun, Poli, and Yun]{scimeca2021shortcut}
L.~Scimeca, S.~J. Oh, S.~Chun, M.~Poli, and S.~Yun.
\newblock Which shortcut cues will dnns choose? a study from the parameter-space perspective.
\newblock \emph{arXiv preprint arXiv:2110.03095}, 2021.

\bibitem[Schulman et~al.(2017)Schulman, Wolski, Dhariwal, Radford, and Klimov]{schulman2017proximal}
J.~Schulman, F.~Wolski, P.~Dhariwal, A.~Radford, and O.~Klimov.
\newblock Proximal policy optimization algorithms.
\newblock \emph{arXiv preprint arXiv:1707.06347}, 2017.

\bibitem[Espeholt et~al.(2018)Espeholt, Soyer, Munos, Simonyan, Mnih, Ward, Doron, Firoiu, Harley, Dunning, et~al.]{espeholt2018impala}
L.~Espeholt, H.~Soyer, R.~Munos, K.~Simonyan, V.~Mnih, T.~Ward, Y.~Doron, V.~Firoiu, T.~Harley, I.~Dunning, et~al.
\newblock Impala: Scalable distributed deep-rl with importance weighted actor-learner architectures.
\newblock In \emph{International conference on machine learning}, pages 1407--1416. PMLR, 2018.

\bibitem[Cobbe et~al.(2020)Cobbe, Hesse, Hilton, and Schulman]{cobbe2020leveraging}
K.~Cobbe, C.~Hesse, J.~Hilton, and J.~Schulman.
\newblock Leveraging procedural generation to benchmark reinforcement learning.
\newblock In \emph{International conference on machine learning}, pages 2048--2056. PMLR, 2020.

\bibitem[Horiguchi et~al.(2013)Horiguchi, Winawer, Dougherty, and Wandell]{horiguchi2013human}
H.~Horiguchi, J.~Winawer, R.~F. Dougherty, and B.~A. Wandell.
\newblock Human trichromacy revisited.
\newblock \emph{Proceedings of the National Academy of Sciences}, 110\penalty0 (3):\penalty0 E260--E269, 2013.

\bibitem[Chittka(2022)]{chittka2022mind}
L.~Chittka.
\newblock \emph{The mind of a bee}.
\newblock Princeton University Press, 2022.

\bibitem[Bloch and Martinoya(1983)]{bloch1983specialization}
S.~Bloch and C.~Martinoya.
\newblock Specialization of visual functions for different retinal areas in the pigeon.
\newblock \emph{Advances in vertebrate neuroethology}, pages 359--368, 1983.

\bibitem[Hilton et~al.(2020)Hilton, Cammarata, Carter, Goh, and Olah]{hilton2020understanding}
J.~Hilton, N.~Cammarata, S.~Carter, G.~Goh, and C.~Olah.
\newblock Understanding {RL} vision.
\newblock \emph{Distill}, 5\penalty0 (11):\penalty0 e29, 2020.

\bibitem[McGrath et~al.(2022)McGrath, Kapishnikov, Toma{\v{s}}ev, Pearce, Wattenberg, Hassabis, Kim, Paquet, and Kramnik]{mcgrath2022acquisition}
T.~McGrath, A.~Kapishnikov, N.~Toma{\v{s}}ev, A.~Pearce, M.~Wattenberg, D.~Hassabis, B.~Kim, U.~Paquet, and V.~Kramnik.
\newblock Acquisition of chess knowledge in alphazero.
\newblock \emph{Proceedings of the National Academy of Sciences}, 119\penalty0 (47):\penalty0 e2206625119, 2022.

\bibitem[Silver et~al.(2018)Silver, Hubert, Schrittwieser, Antonoglou, Lai, Guez, Lanctot, Sifre, Kumaran, Graepel, et~al.]{silver2018general}
D.~Silver, T.~Hubert, J.~Schrittwieser, I.~Antonoglou, M.~Lai, A.~Guez, M.~Lanctot, L.~Sifre, D.~Kumaran, T.~Graepel, et~al.
\newblock A general reinforcement learning algorithm that masters chess, shogi, and go through self-play.
\newblock \emph{Science}, 362\penalty0 (6419):\penalty0 1140--1144, 2018.

\bibitem[Atrey et~al.(2019)Atrey, Clary, and Jensen]{atrey2019exploratory}
A.~Atrey, K.~Clary, and D.~Jensen.
\newblock Exploratory not explanatory: Counterfactual analysis of saliency maps for deep reinforcement learning.
\newblock \emph{arXiv preprint arXiv:1912.05743}, 2019.

\bibitem[Turner et~al.(2023)Turner, Grietzer, Mini, and Udell]{turner2023understanding}
A.~Turner, P.~Grietzer, U.~Mini, and D.~Udell.
\newblock Understanding and controlling a maze-solving policy network, 2023.
\newblock URL \url{https://www.alignmentforum.org/posts/cAC4AXiNC5ig6jQnc/understanding-and-controlling-a-maze-solving-policy-network}.

\bibitem[Ribeiro et~al.(2016)Ribeiro, Singh, and Guestrin]{ribeiro2016should}
M.~T. Ribeiro, S.~Singh, and C.~Guestrin.
\newblock " why should i trust you?" explaining the predictions of any classifier.
\newblock In \emph{Proceedings of the 22nd ACM SIGKDD international conference on knowledge discovery and data mining}, pages 1135--1144, 2016.

\bibitem[Narla et~al.(2018)Narla, Kuprel, Sarin, Novoa, and Ko]{narla2018automated}
A.~Narla, B.~Kuprel, K.~Sarin, R.~Novoa, and J.~Ko.
\newblock Automated classification of skin lesions: from pixels to practice.
\newblock \emph{Journal of Investigative Dermatology}, 138\penalty0 (10):\penalty0 2108--2110, 2018.

\bibitem[D'Amour et~al.(2022)D'Amour, Heller, Moldovan, Adlam, Alipanahi, Beutel, Chen, Deaton, Eisenstein, Hoffman, et~al.]{d2022underspecification}
A.~D'Amour, K.~Heller, D.~Moldovan, B.~Adlam, B.~Alipanahi, A.~Beutel, C.~Chen, J.~Deaton, J.~Eisenstein, M.~D. Hoffman, et~al.
\newblock Underspecification presents challenges for credibility in modern machine learning.
\newblock \emph{The Journal of Machine Learning Research}, 23\penalty0 (1):\penalty0 10237--10297, 2022.

\bibitem[Sellam et~al.(2021)Sellam, Yadlowsky, Wei, Saphra, D'Amour, Linzen, Bastings, Turc, Eisenstein, Das, et~al.]{sellam2021multiberts}
T.~Sellam, S.~Yadlowsky, J.~Wei, N.~Saphra, A.~D'Amour, T.~Linzen, J.~Bastings, I.~Turc, J.~Eisenstein, D.~Das, et~al.
\newblock The multiberts: Bert reproductions for robustness analysis.
\newblock \emph{arXiv preprint arXiv:2106.16163}, 2021.

\end{thebibliography}

\normalsize
\clearpage
\appendix

\section{Disappearing objects}
\label{appendix:disappearing-objects}

\begin{figure}[htbp]
    \centering
    \subfigure[Full resolution]{
        \includegraphics[width=0.45\textwidth]{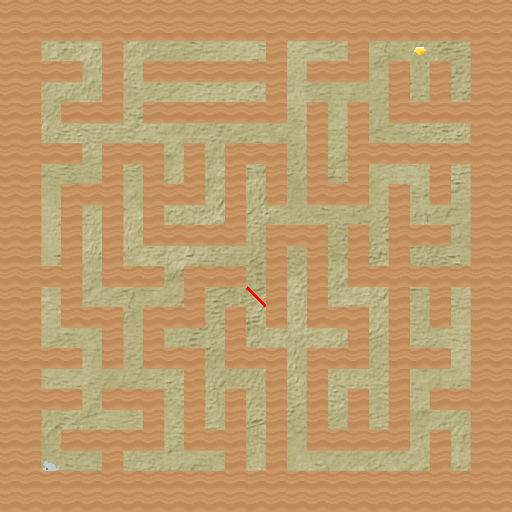}
        \label{fig:correct-human}
    }
    \hspace{0.5cm}
    \subfigure[Correct downsampling (agent view)]{
        \includegraphics[width=0.45\textwidth]{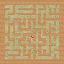}
        \label{fig:correct-agent}
    }
    
    \subfigure[Full resolution]{
        \includegraphics[width=0.45\textwidth]{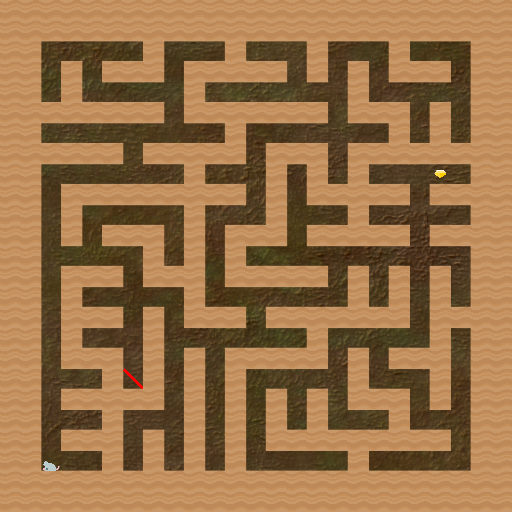}
        \label{fig:incorrect-human}
    }
    \hspace{0.5cm}
    \subfigure[Incorrect downsampling - red line missing]{
        \includegraphics[width=0.45\textwidth]{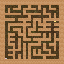}
        \label{fig:incorrect-agent}
    }
    \caption{The observations are downsampled from 512$\times$512 to 64$\times$64 before being given to the agent. Images (a) and (b) show how the downsampled objects can be only 1--2 pixels in size. Images (c) and (d) show how sometimes an object can be invisible to the agent. This happens roughly 50\% of the time for line objects and 20\% of the time for gem objects.}
    \label{fig:downsampling}
\end{figure}

Here we explain the issue of disappearing objects in the original environment used to demonstrate colour versus shape goal misgeneralization by \citet{di2022goal}.
Information is invariably lost when downsampling a pixel-based environment from human view (512$\times$512 pixels) to agent view (64$\times$64 pixels).
In the case of a pixel-based maze with 25$\times$25 grid squares, each square will get downsampled to rectangles with side lengths of 2--3 pixels.
In a sample of 100 randomly selected levels, about half of the time, this will make the line goal object completely invisible to the agent.
The rest of the time, the gem and the line objects are 1--4 coloured pixels, as seen in \autoref{fig:downsampling}.
The gem object is invisible only 20\% of the time.
This makes the result that \textit{the agent prefers the yellow gem} relatively weak -- half the time, the agent does not even see the red line.
This might mean that the agent learns some algorithm, such as “follow the left wall”, which helps it find the goal when it is invisible, as well as learning to navigate to the yellow line.
In addition, the yellow line and the yellow gem are slightly different shades of yellow.

\clearpage
\section{Original and new environment assets}
\label{appendix:original-and-our-environment-assets}

\begin{figure}[htbp]
\begin{tcolorbox}[colback=gray!20,colframe=white]
    \centering
    \subfigure[Original assets]{
        \includegraphics[width=0.75\textwidth]{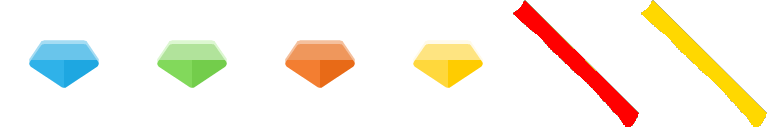}
    }
    
    \subfigure[New assets]{
        \includegraphics[width=0.95\textwidth]{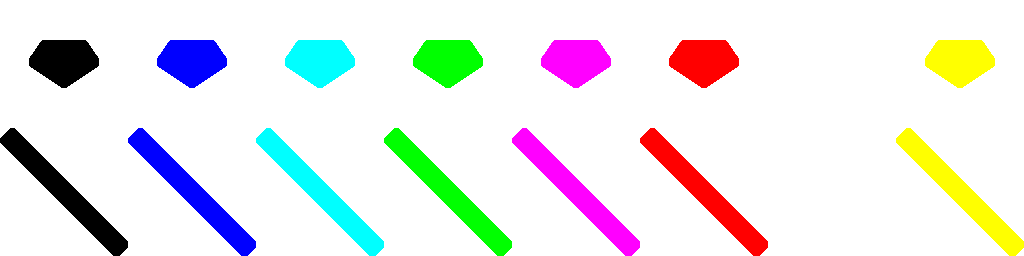}
    }

    \caption{The original assets had slightly irregular shapes for the lines, the red line had a few yellow pixels on it, and the gem colours were not exactly the same as the corresponding line colours. We created new assets with the eight basic colours. A grey background is used in the figure to improve the visibility of white objects.}
    \label{fig:new-assets}
\end{tcolorbox}
\end{figure}

\clearpage
\section{Procgen Maze background texture analysis}
\label{appendix:procgen-maze-background-texture-analysis}

As seen below, Procgen Maze environment texture colours are not entirely random and have some biases towards specific colours. This can make objects more or less visible in certain colour channels, influencing final learned behaviour.

\begin{figure}[htbp]
    \centering
    \subfigure{\includegraphics[width=0.45\textwidth]{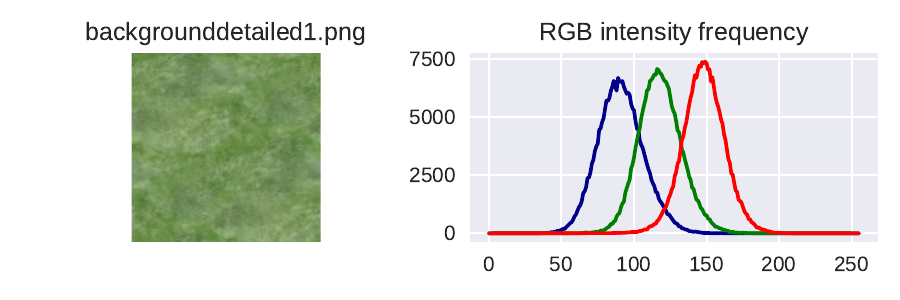}}
    \subfigure{\includegraphics[width=0.45\textwidth]{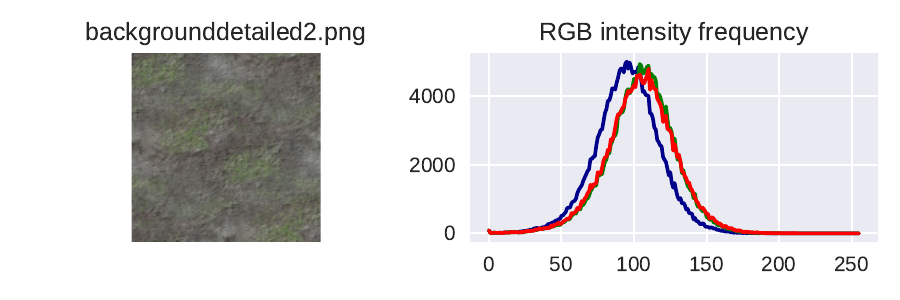}}
    
    \subfigure{\includegraphics[width=0.45\textwidth]{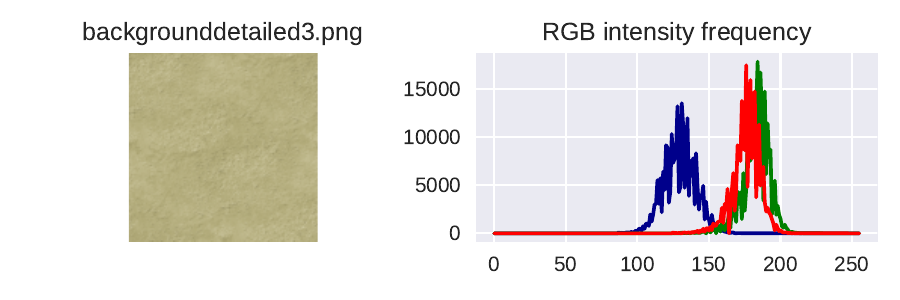}}
    \subfigure{\includegraphics[width=0.45\textwidth]{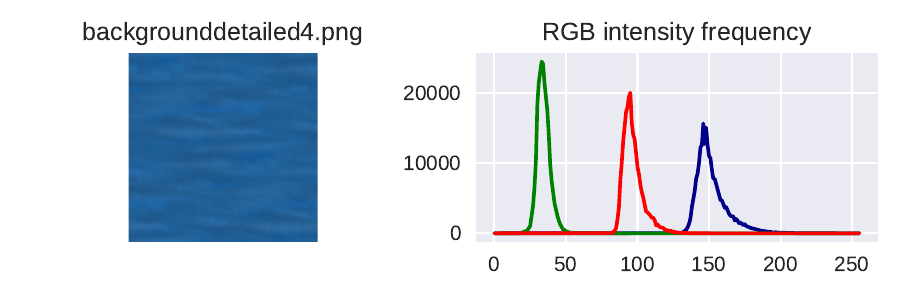}}
    
    \subfigure{\includegraphics[width=0.45\textwidth]{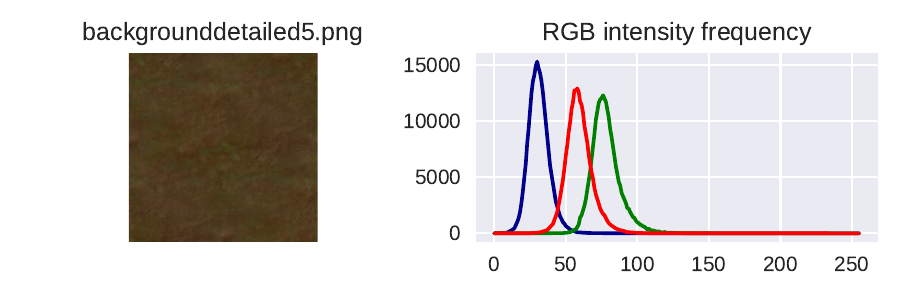}}
    \subfigure{\includegraphics[width=0.45\textwidth]{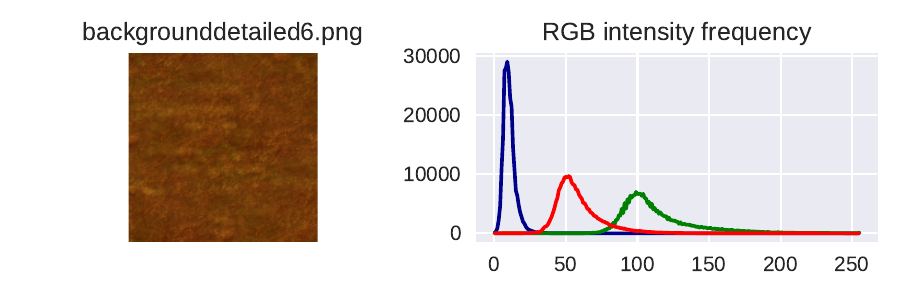}}
    
    \subfigure{\includegraphics[width=0.45\textwidth]{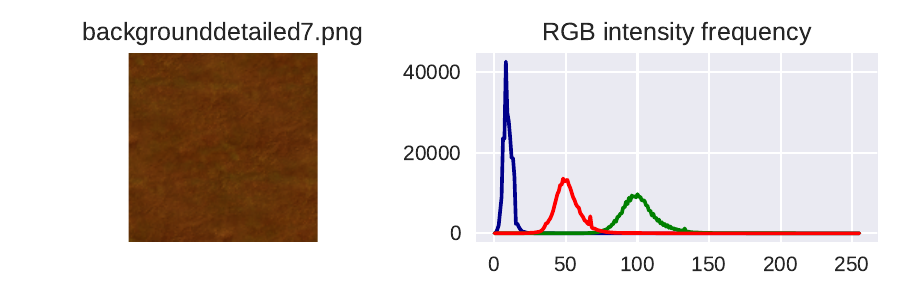}}
    \subfigure{\includegraphics[width=0.45\textwidth]{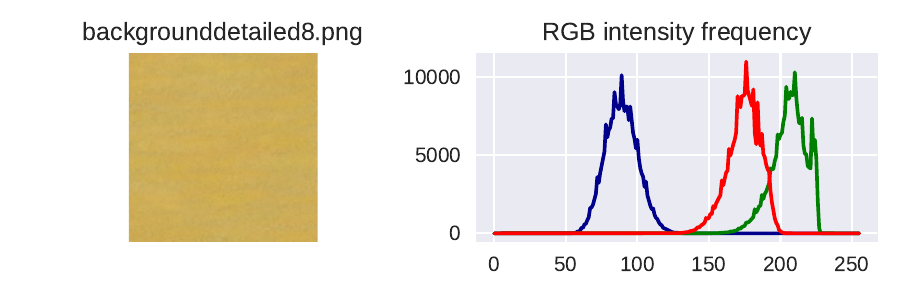}}
    
    \subfigure{\includegraphics[width=0.45\textwidth]{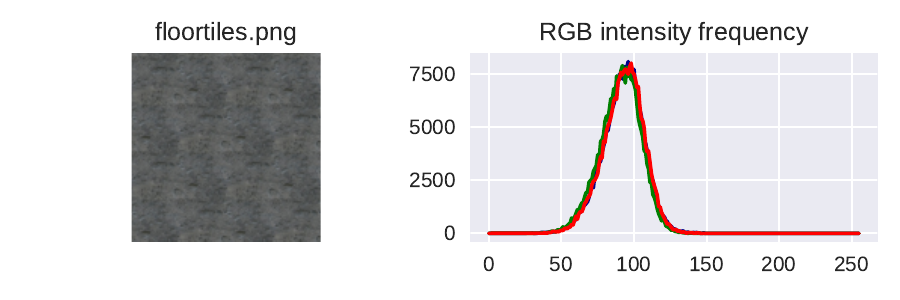}}
    
    \caption{The textures used in the default Procgen Maze environments and their RGB intensity distributions. This might impact goal misgeneralization in different colour and shape preference experiments.}
    \label{fig:background-analysis}
\end{figure}

\clearpage
\section{List of outliers}
\label{appendix:list-of-outliers}

Below is a list of outliers we found among the 512 agents trained to reach a white line on a black background (unless otherwise indicated).

\begin{figure}[htbp]
  \centering
  \subfigure{\includegraphics[width=0.3\textwidth]{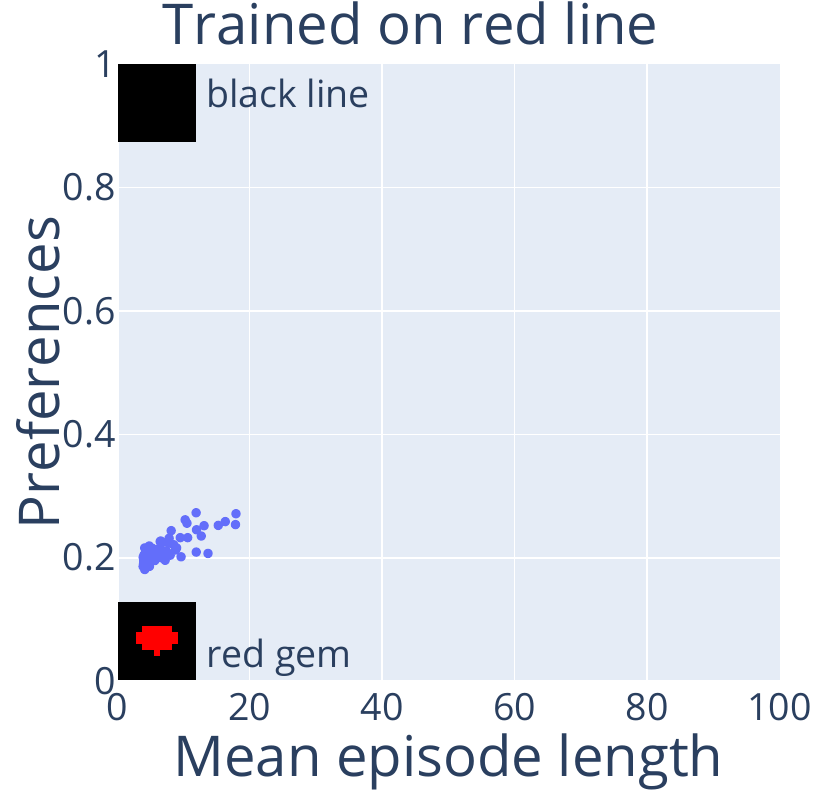}}
  \subfigure{\includegraphics[width=0.3\textwidth]{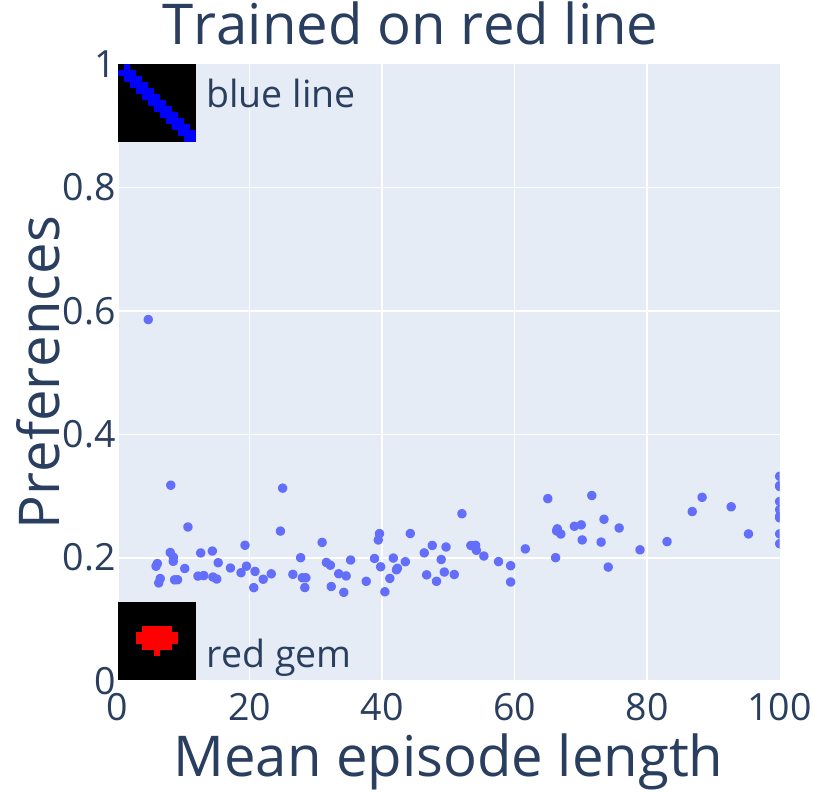}}
  \subfigure{\includegraphics[width=0.3\textwidth]{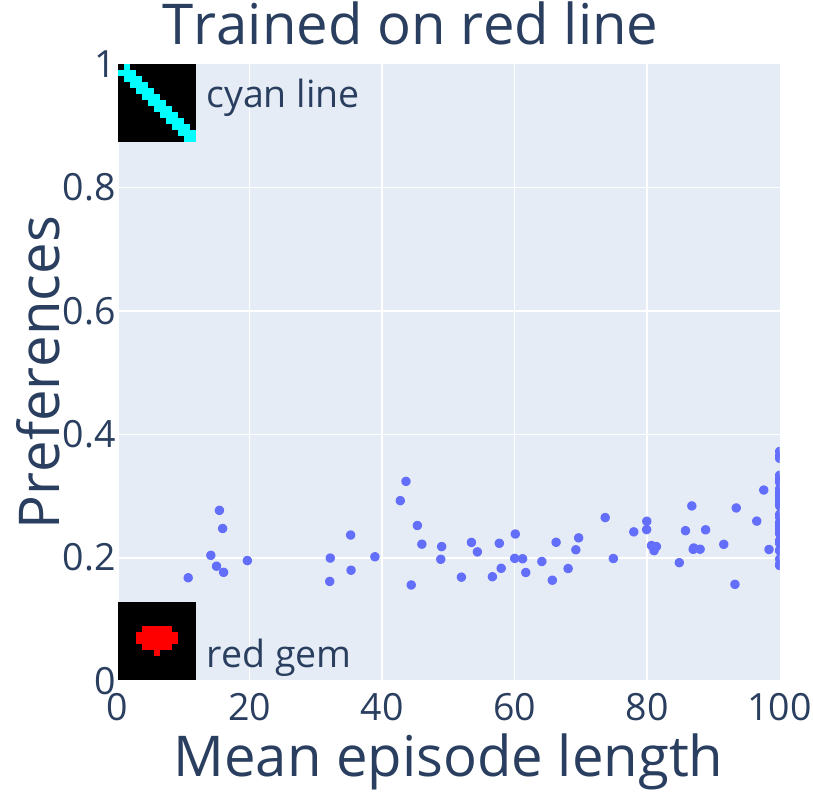}}
  
  \subfigure{\includegraphics[width=0.3\textwidth]{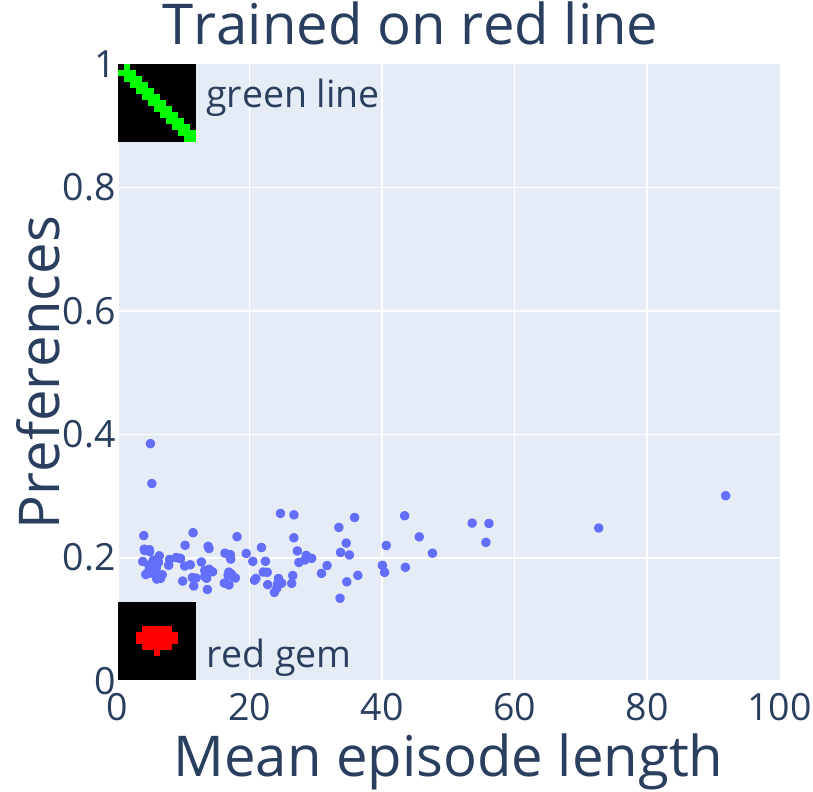}}
  \subfigure{\includegraphics[width=0.3\textwidth]{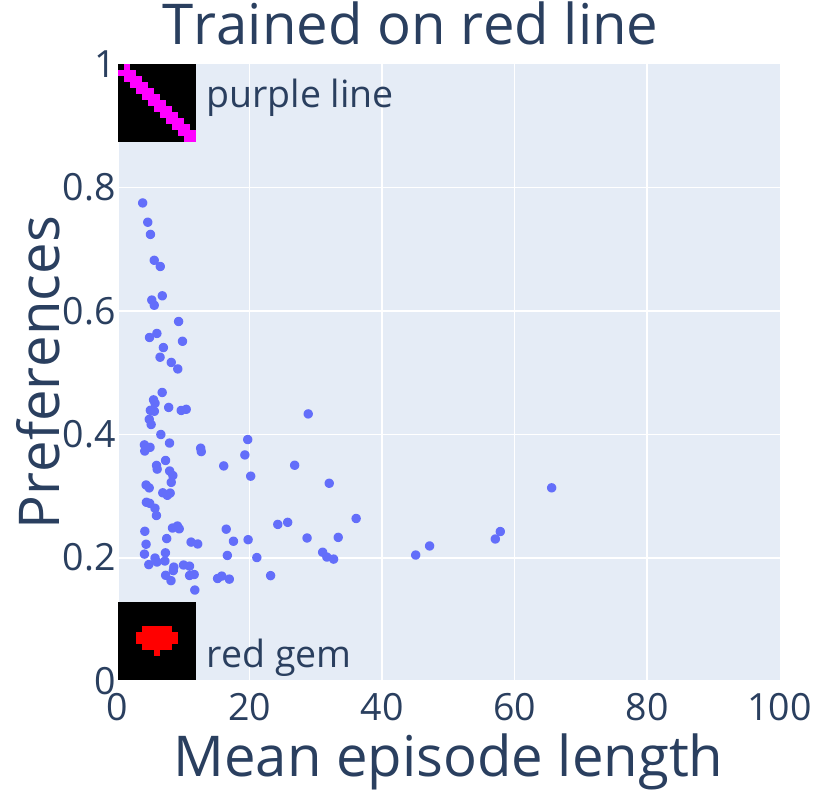}}
  \subfigure{\includegraphics[width=0.3\textwidth]{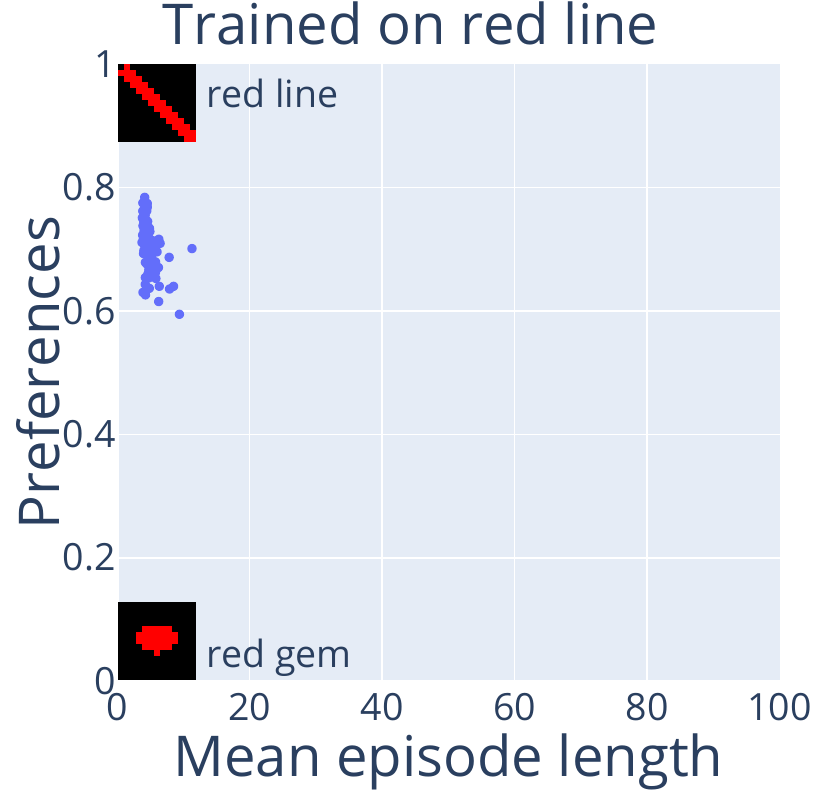}}
  
  \subfigure{\includegraphics[width=0.3\textwidth]{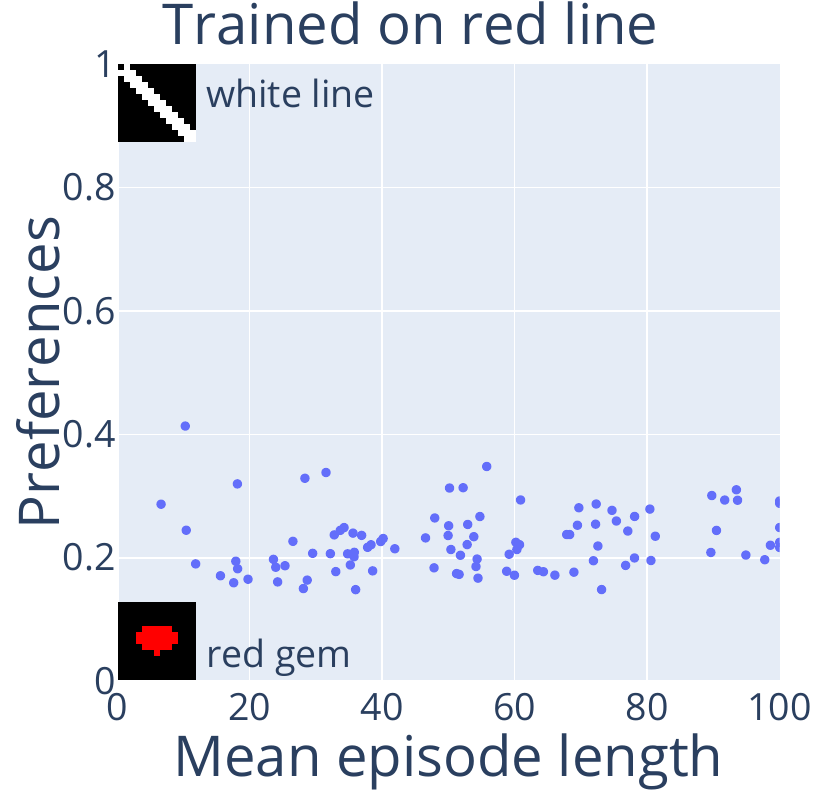}}
  \subfigure{\includegraphics[width=0.3\textwidth]{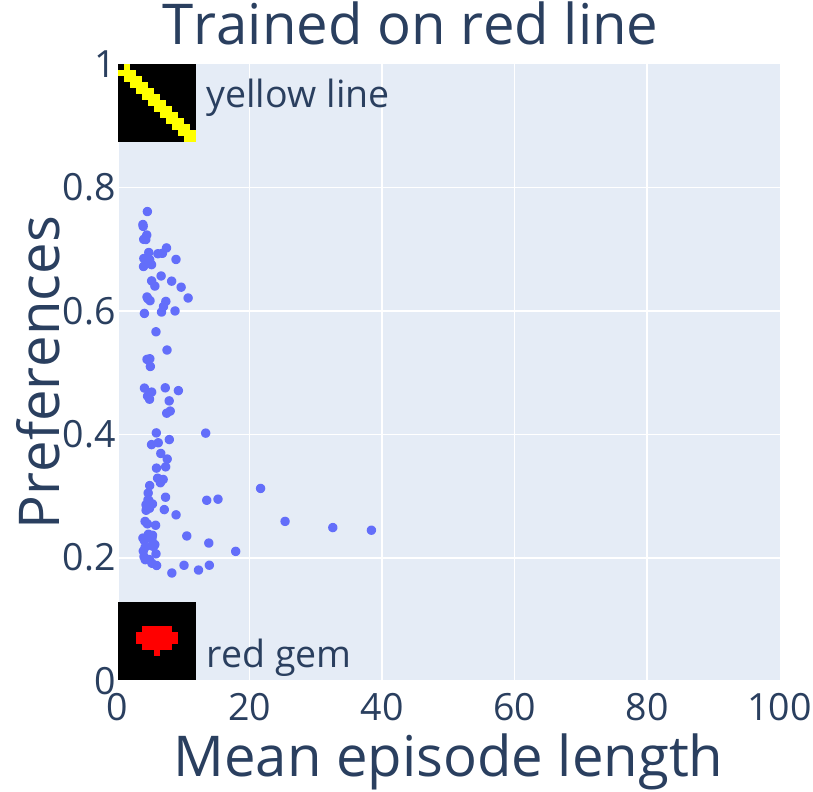}}
  \caption{Trained on the red line with black backgrounds, tested on different colour lines versus red gem.}
  \label{fig:train-red-line-black-background-01}
\end{figure}

\begin{figure}[htbp]
  \centering
  \subfigure{\includegraphics[width=0.3\textwidth]{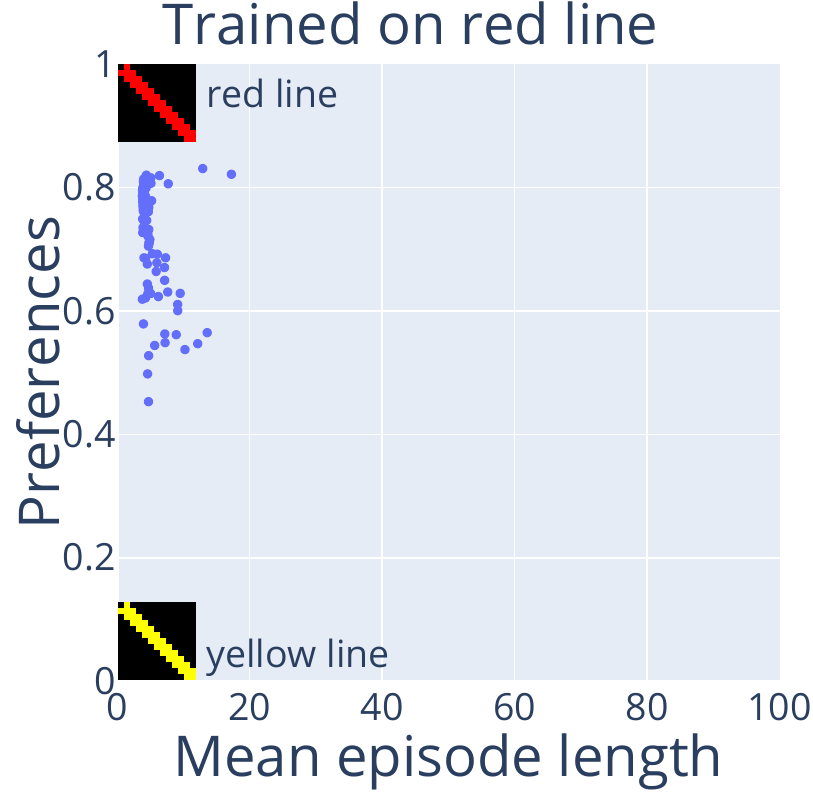}}
  \subfigure{\includegraphics[width=0.3\textwidth]{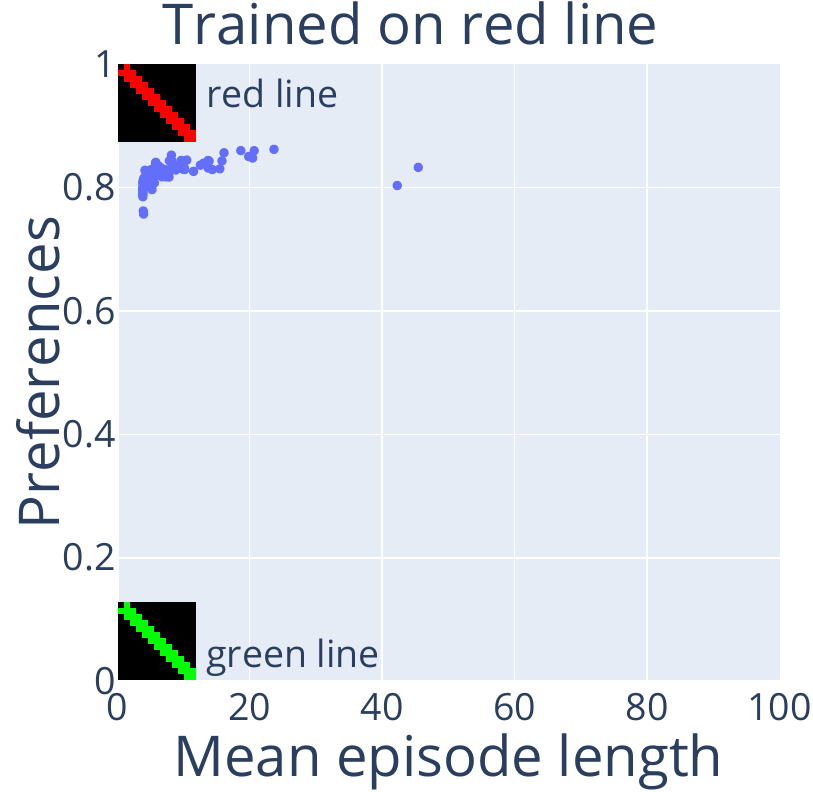}}
  \subfigure{\includegraphics[width=0.3\textwidth]{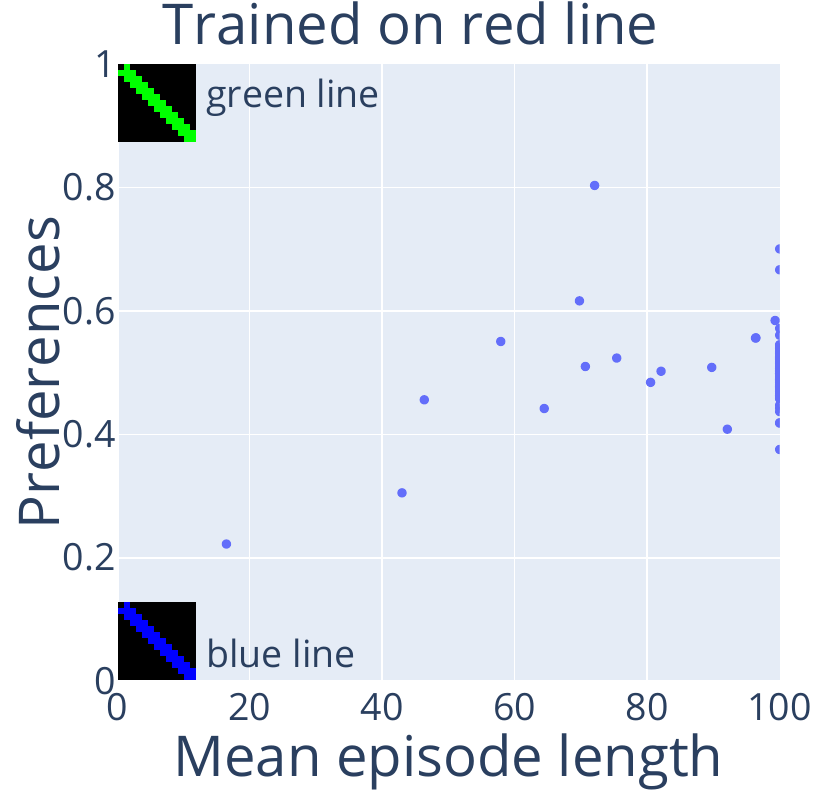}}
  
  \subfigure{\includegraphics[width=0.3\textwidth]{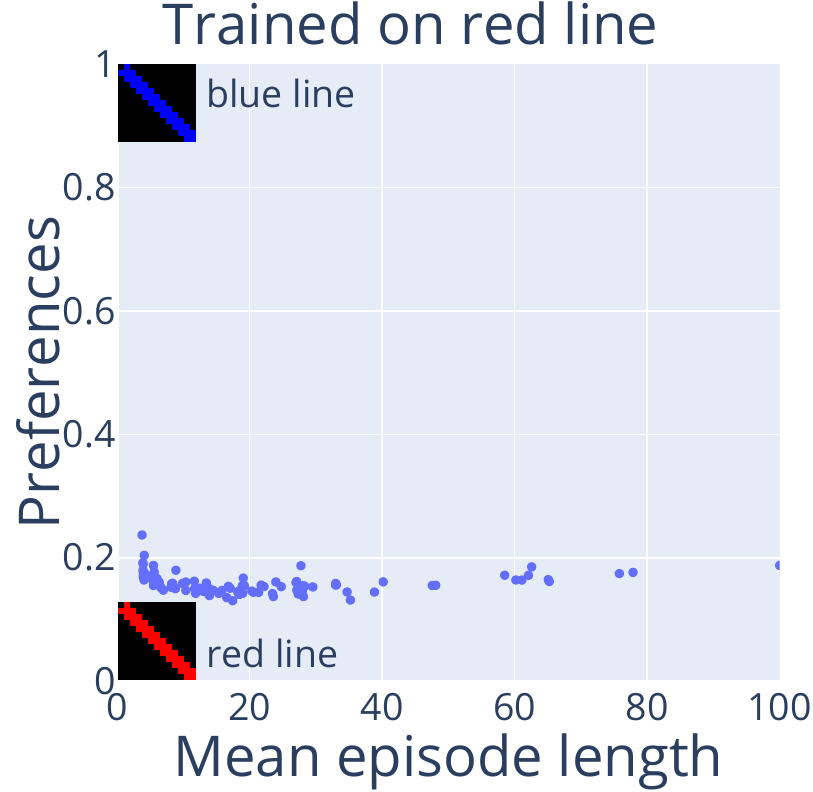}}
  \subfigure{\includegraphics[width=0.3\textwidth]{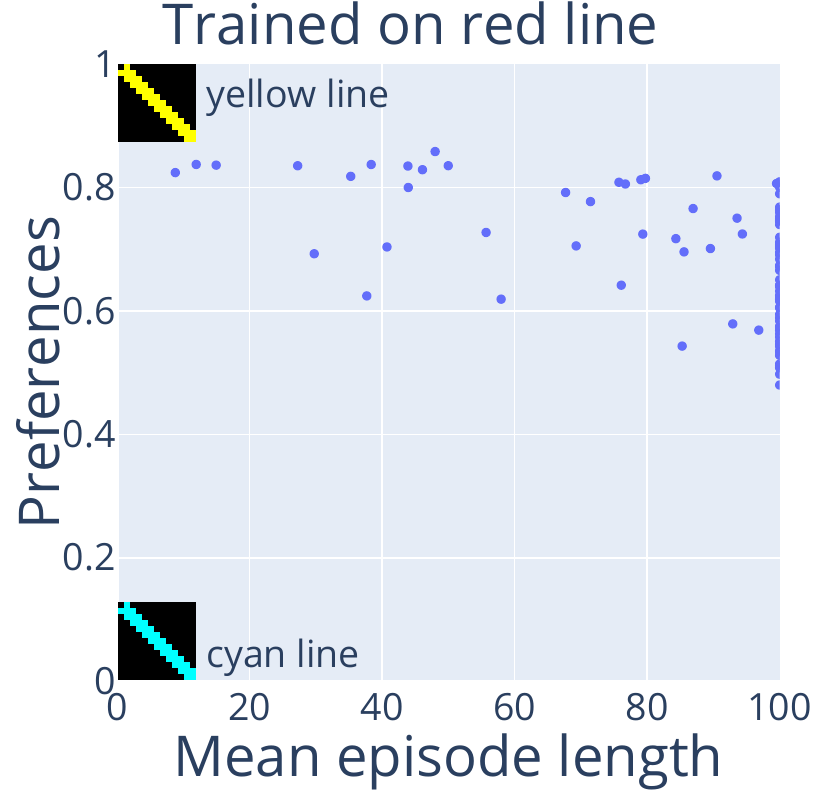}}
  \subfigure{\includegraphics[width=0.3\textwidth]{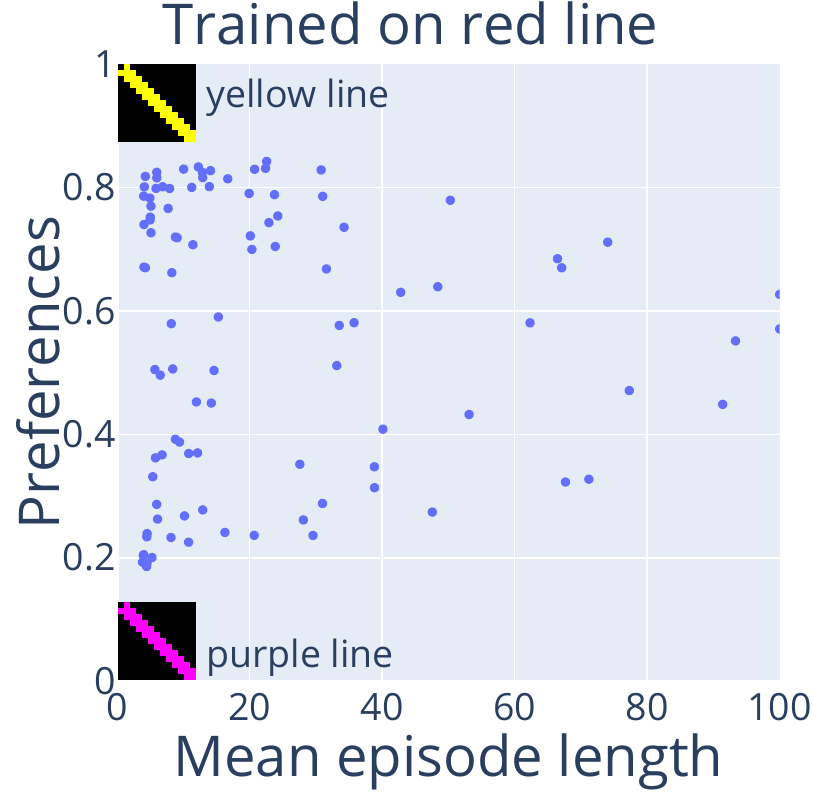}}
  
  \caption{Trained on the red line with black backgrounds, tested on different colour lines versus other different colour lines. This seems to show some avoiding behaviour of red line versus blue line and red line versus green line, indicated by above 80\% preference for the red line in those cases. It may perceive the blue and green lines as walls.}
  \label{fig:train-red-line-black-background-02}
\end{figure}

\begin{figure}[htbp]
  \centering
  \subfigure{\includegraphics[width=0.3\textwidth]{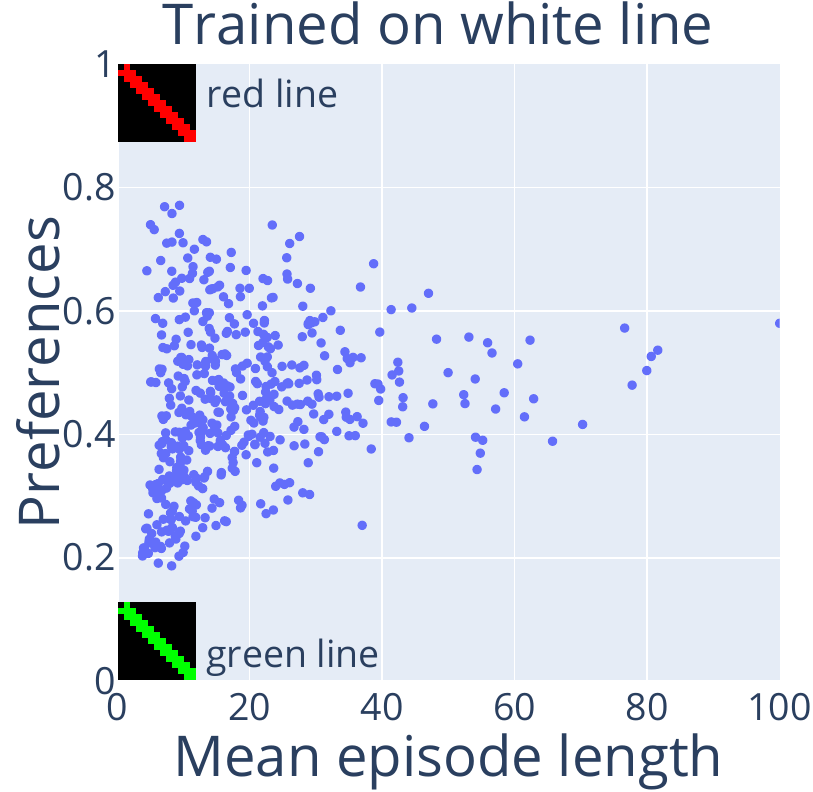}}
  \subfigure{\includegraphics[width=0.3\textwidth]{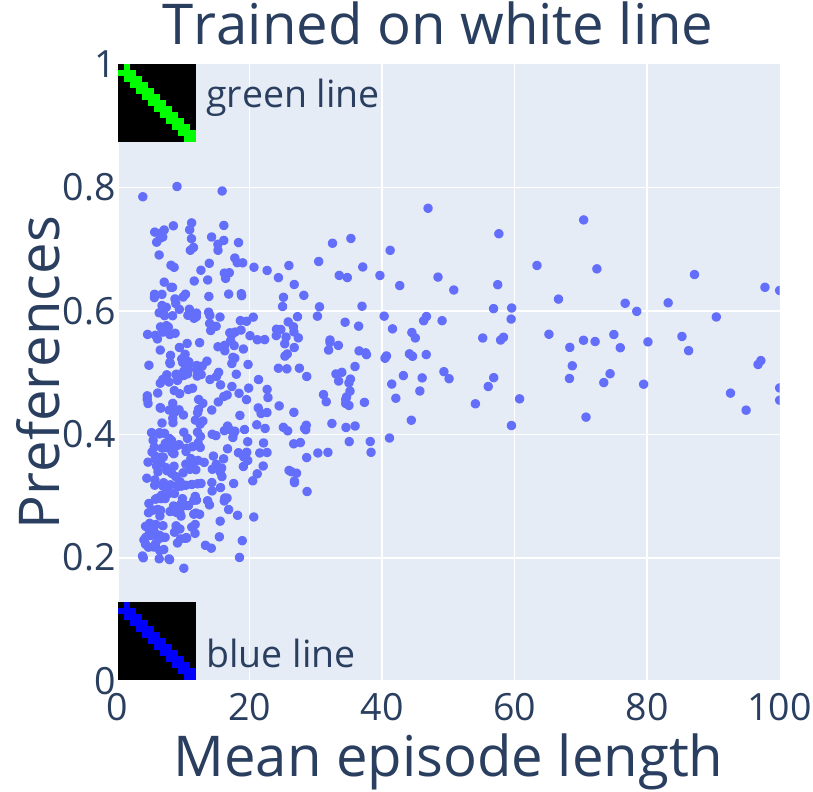}}
  \subfigure{\includegraphics[width=0.3\textwidth]{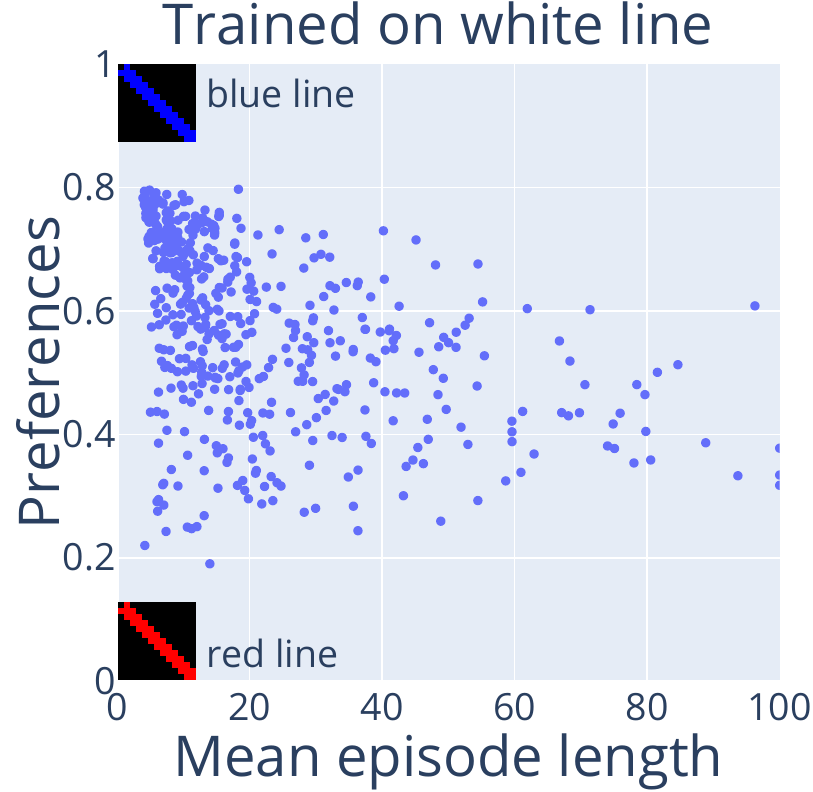}}
  
  \subfigure{\includegraphics[width=0.3\textwidth]{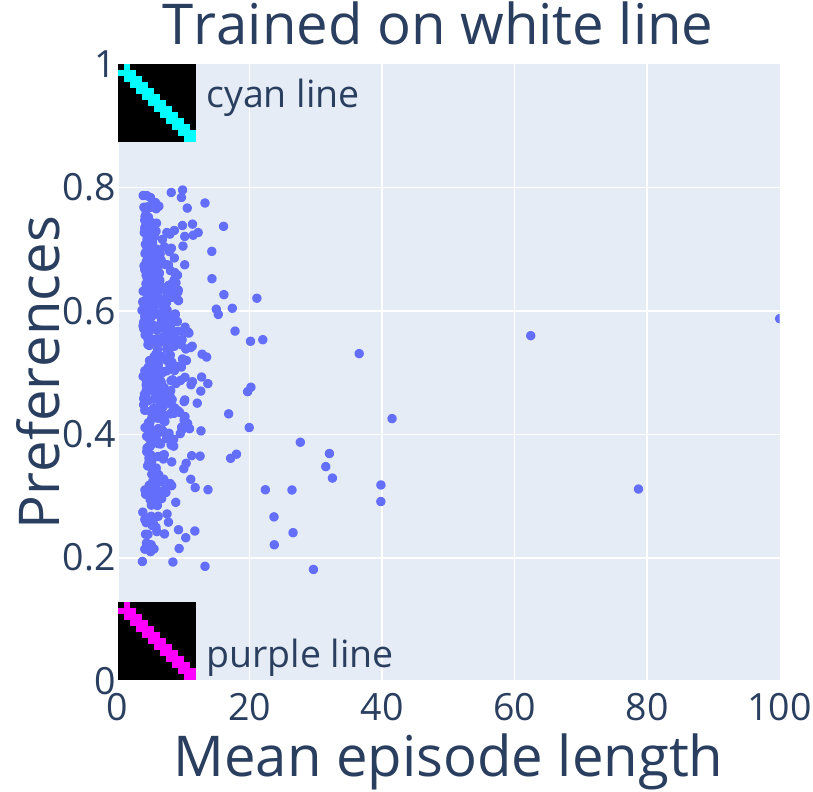}}
  \subfigure{\includegraphics[width=0.3\textwidth]{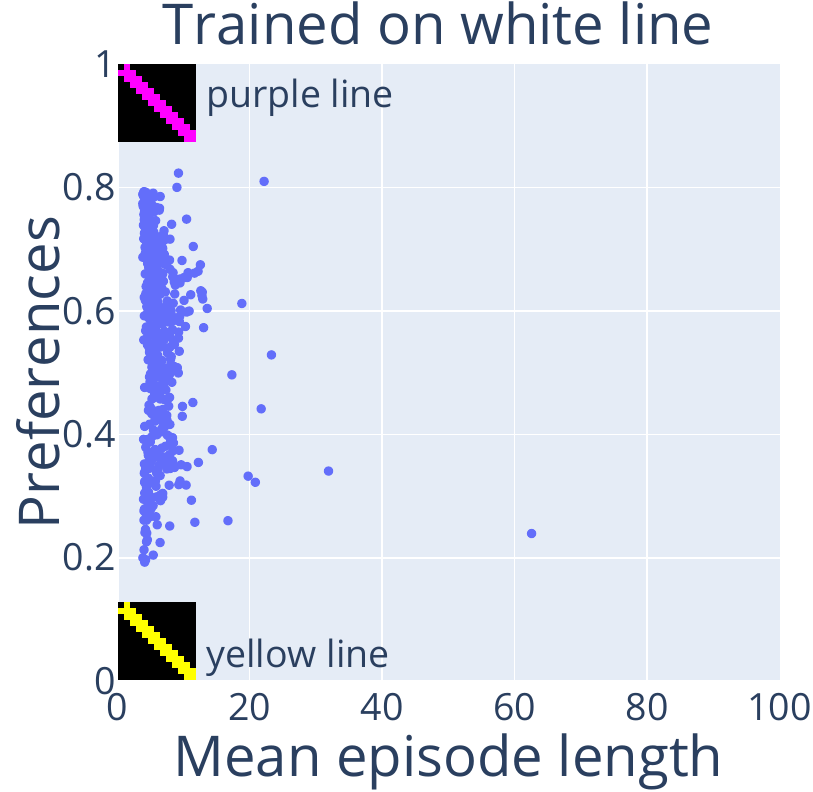}}
  \subfigure{\includegraphics[width=0.3\textwidth]{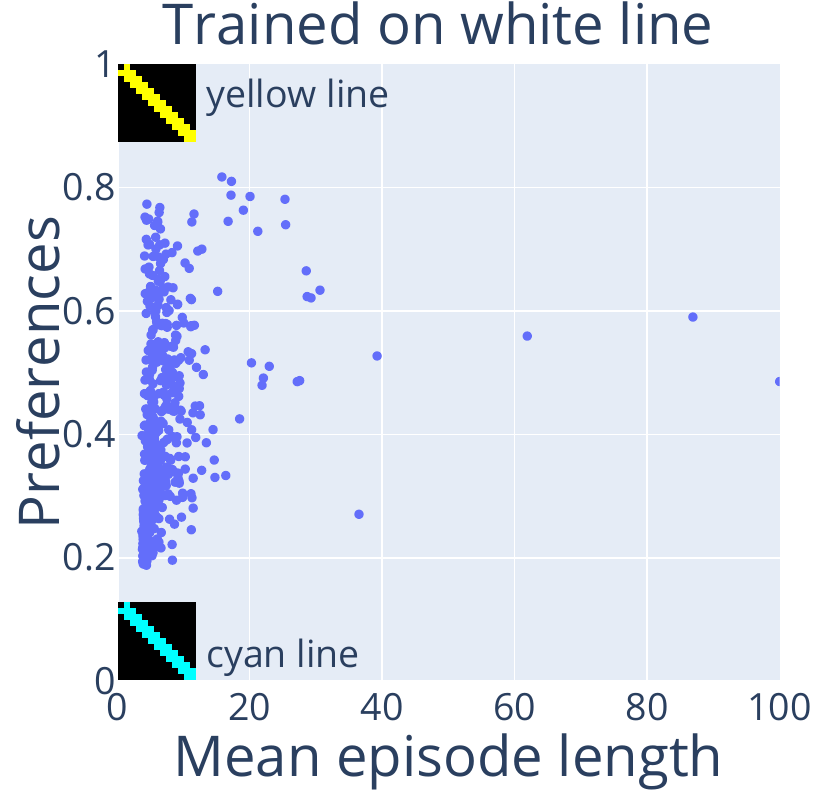}}
  
  \subfigure{\includegraphics[width=0.3\textwidth]{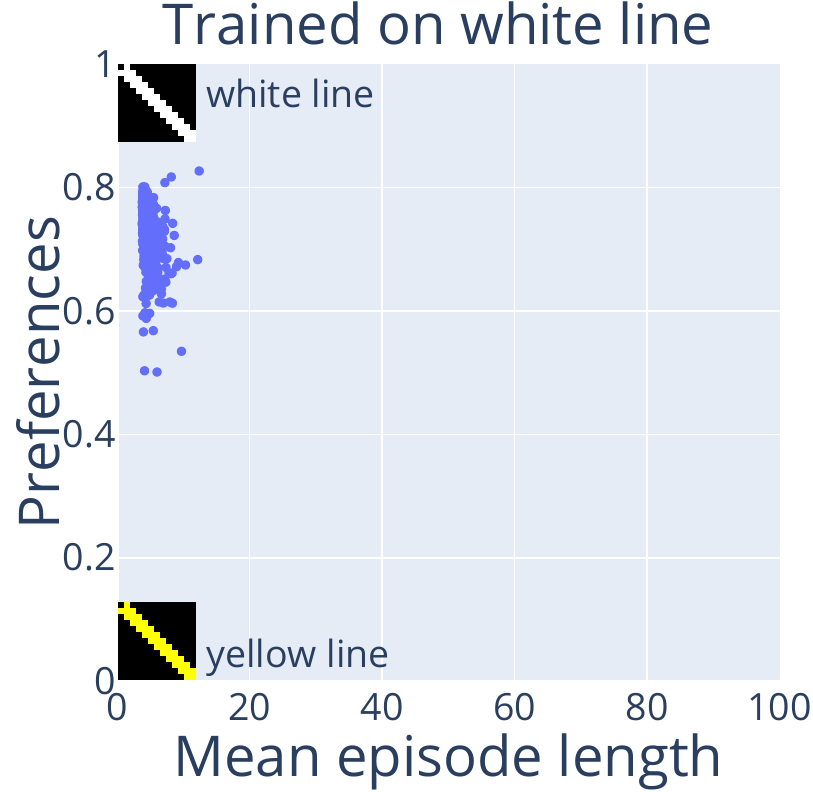}}
  \subfigure{\includegraphics[width=0.3\textwidth]{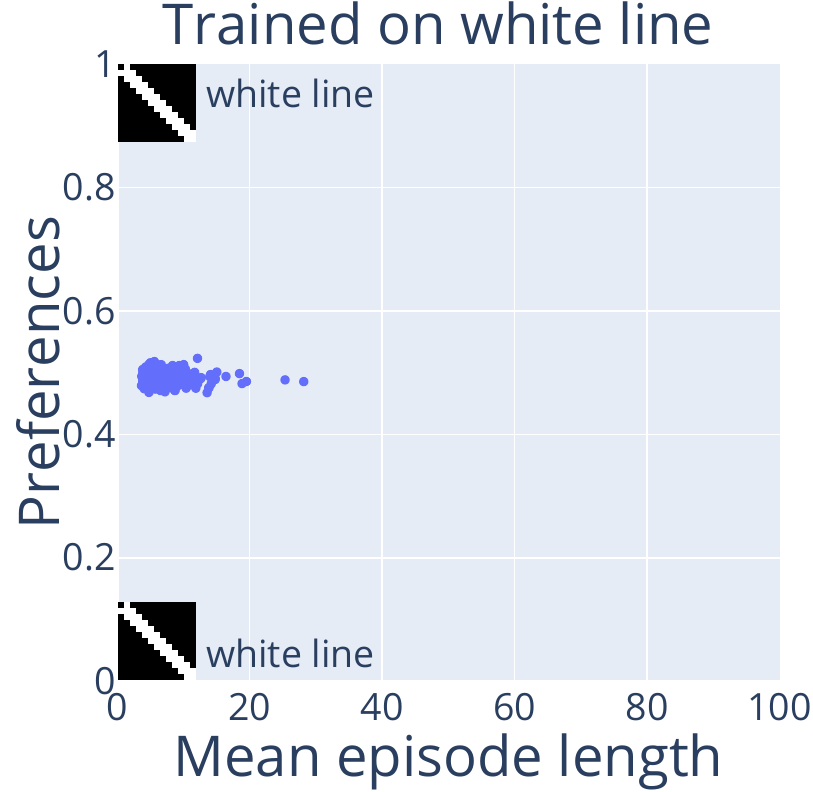}}
  \subfigure{\includegraphics[width=0.3\textwidth]{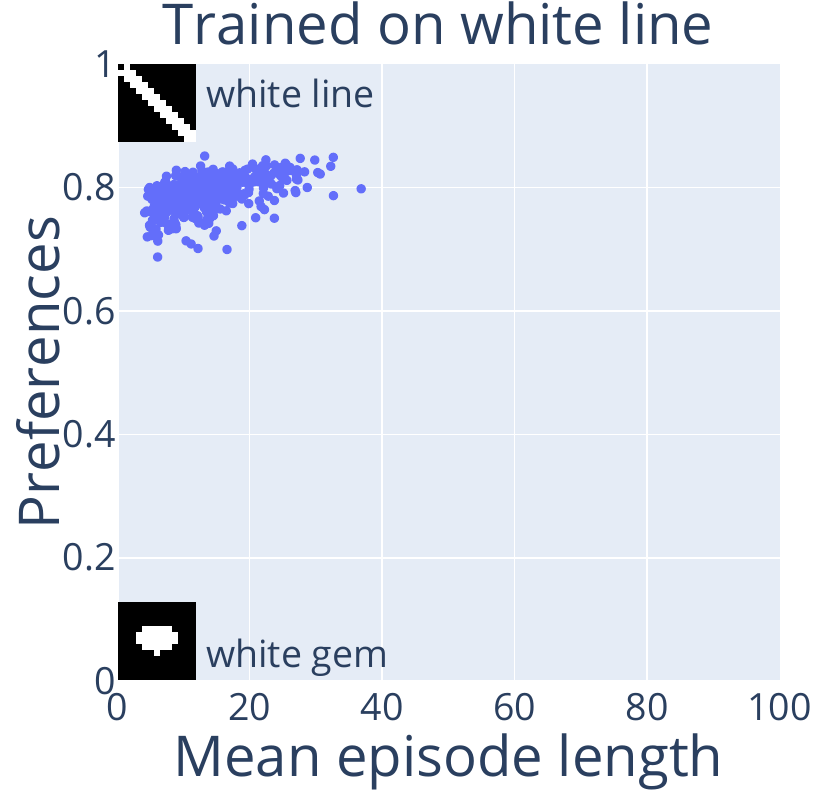}}

  \subfigure{\includegraphics[width=0.3\textwidth]{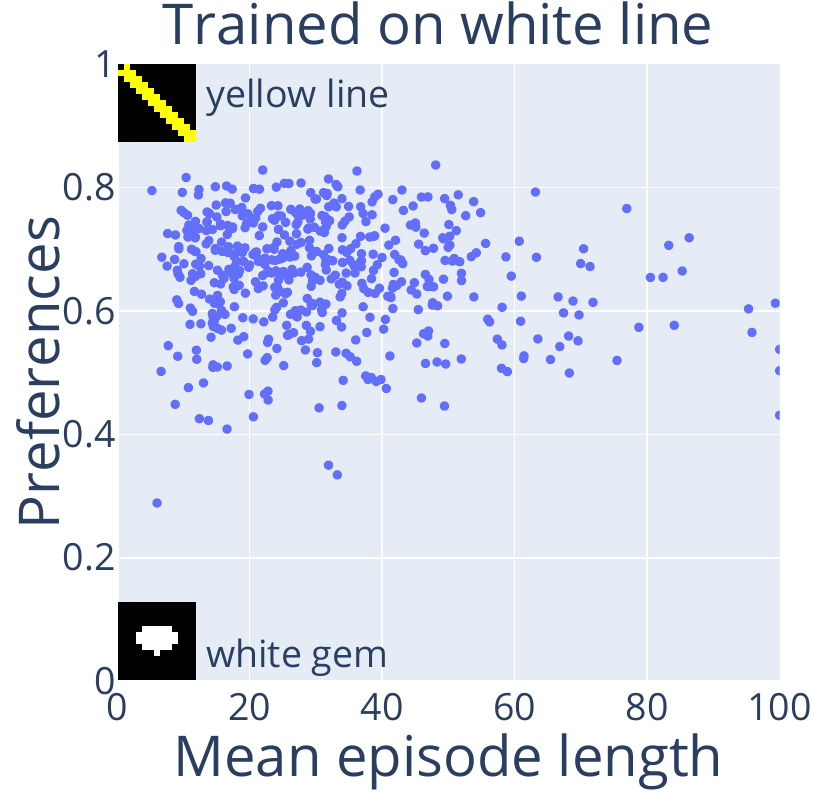}}
  \subfigure{\includegraphics[width=0.3\textwidth]{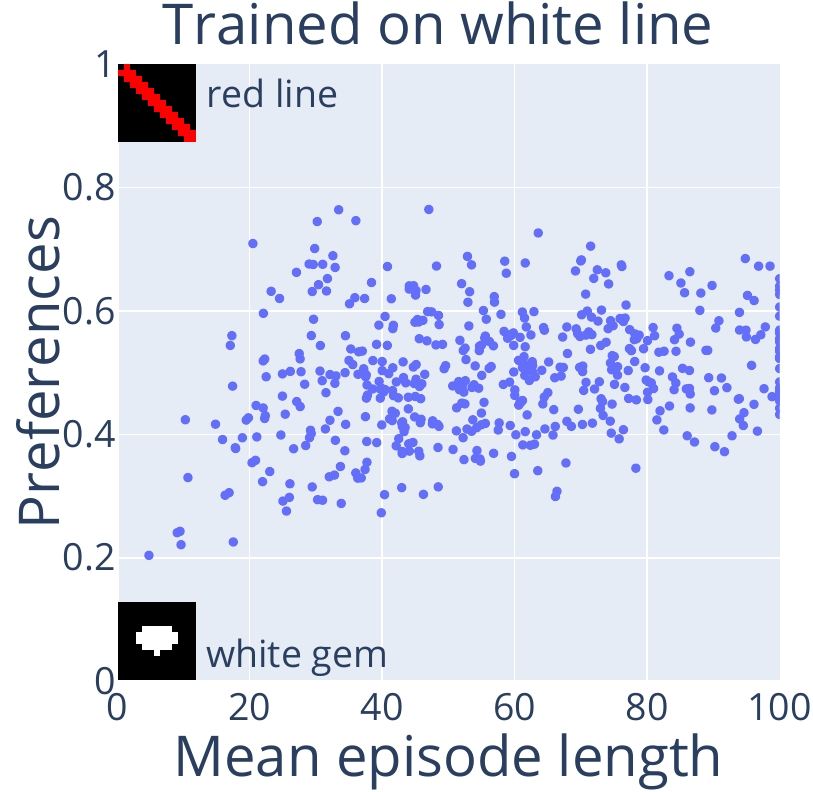}}

  \caption{Trained on the white line with black backgrounds, tested on different colour lines versus other different colour lines. Also, red, yellow and white lines versus white gem, representing one, two and three colour channel match. 512 agents instead of 100.}
  \label{fig:train-white-line-black-background}
\end{figure}

\begin{itemize}
    \item Agent-2875 -- completely breaks to worse than random performance when the lines are single channel (red, green or blue). Really likes purple lines.
    \item Agent-56 and Agent-3824 -- do not prefer the original white line to a yellow line, seemingly ignoring the blue channel (3rd row, left plot of \autoref{fig:train-white-line-black-background}, two bottom dots).
    \item Agent-1278 -- very high preference for the blue line and breaks on the red line versus the green line.
    \item Agent-410 -- very high preference for the blue line, but does not break on the red line versus the green line.
    \item Agent-2505 -- no colour preference, highly capable in each setting.
    \item Agent-9986 -- no colour preference, much less capable on single channel colours.
    \item Agent-9510 -- loses a large portion of capabilities when two white lines are present (3rd row, middle plot of \autoref{fig:train-white-line-black-background}, right-most dot).
    \item Agent-6406 -- breaks transitivity, slightly preferring cyan over yellow over purple over cyan. However, the preferences are between 45--50\%, so this is almost surely just a statistical artefact.
    \item Agent-439 -- only one out of 512 to strongly prefer white gem over yellow line while being highly capable (bottom left plot of \autoref{fig:train-white-line-black-background}, bottom left dot).
    \item Agent-8894 -- (trained on yellow line with black backgrounds) prefers the green line over the red line over the yellow gem (right plot in \autoref{fig:red-line-green-line-and-OLS}, top left dot). It seems to have learned about shape and colour and prefers green, but not as strongly as the line shape.
    \item Agent-1983 -- (trained on red line with black backgrounds) slightly prefers blue line to red gem (top middle plot of \autoref{fig:train-red-line-black-background-01}, top left dot). Also capably pursues a blue line when given a choice between a blue line and a green line (top right plot of \autoref{fig:train-red-line-black-background-02}, bottom left dot). This one is peculiar -- why did it learn to seek a blue line while training to reach a red line?
\end{itemize}

\clearpage
\section{Other experiments}
\label{appendix:other-experiments}

We trained over 1,000 agents but showed the results for only 700 of them.
Here we briefly describe our findings with the remaining 300 agents.
When we replaced the textures with a grey background instead of the black one, making lines visible in all three channels, the asymmetry between the red line and the green line versus the yellow gem disappeared as expected.
However, we observed much more capability loss in this setting.
We tried that by training to reach a yellow line (100 agents) and a red line (100 agents).

When we trained another 100 agents to reach a red line on a black background, we expected there to be no goal misgeneralization because there was only a single channel through which to detect the object. This was not the case. When tested on a yellow line versus a red gem, there was a roughly even split between agents that pursued each object.
We do not know how to explain this and leave it for future research.

\begin{figure}[htbp]
  \centering
  \subfigure{\includegraphics[width=0.3\textwidth]{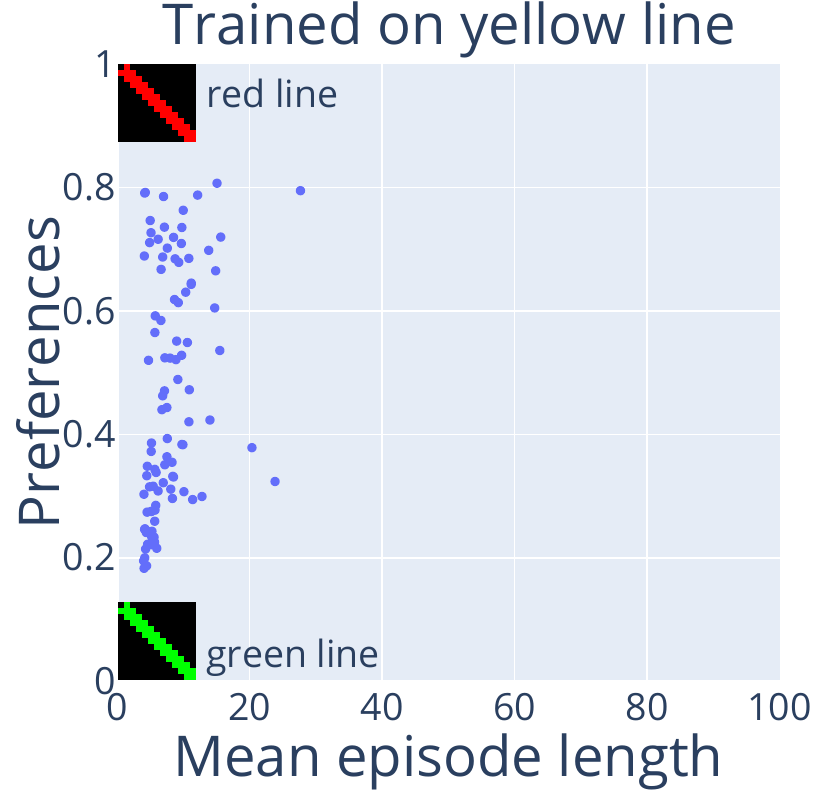}}
  \subfigure{\includegraphics[width=0.3\textwidth]{img/train-yellow-line-black-background-test-green-line-red-line-scatter.pdf}}
  \subfigure{\includegraphics[width=0.3\textwidth]{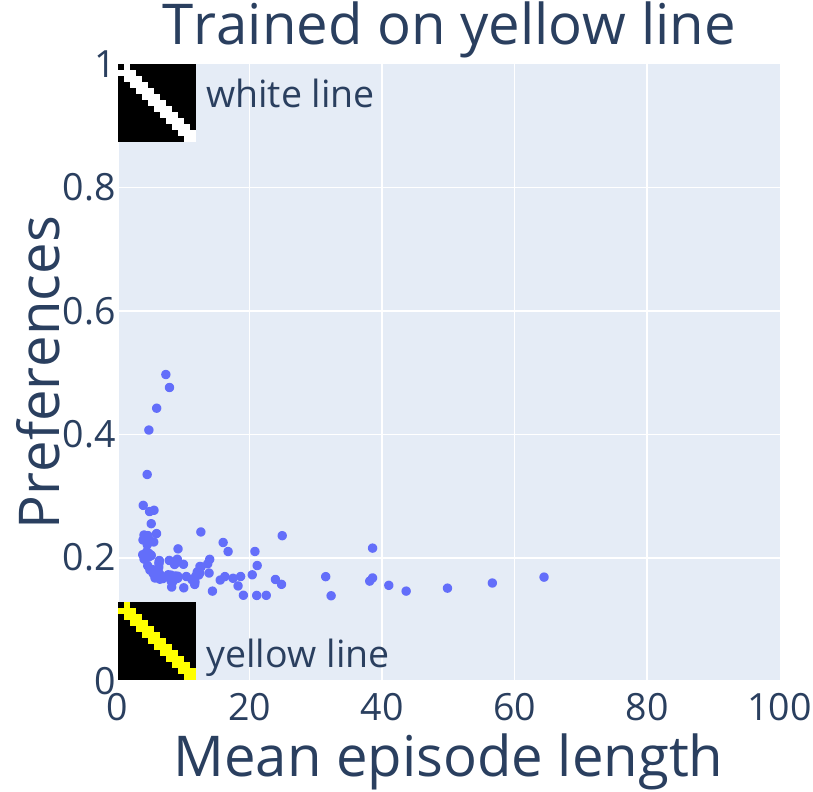}}
  
  \subfigure{\includegraphics[width=0.3\textwidth]{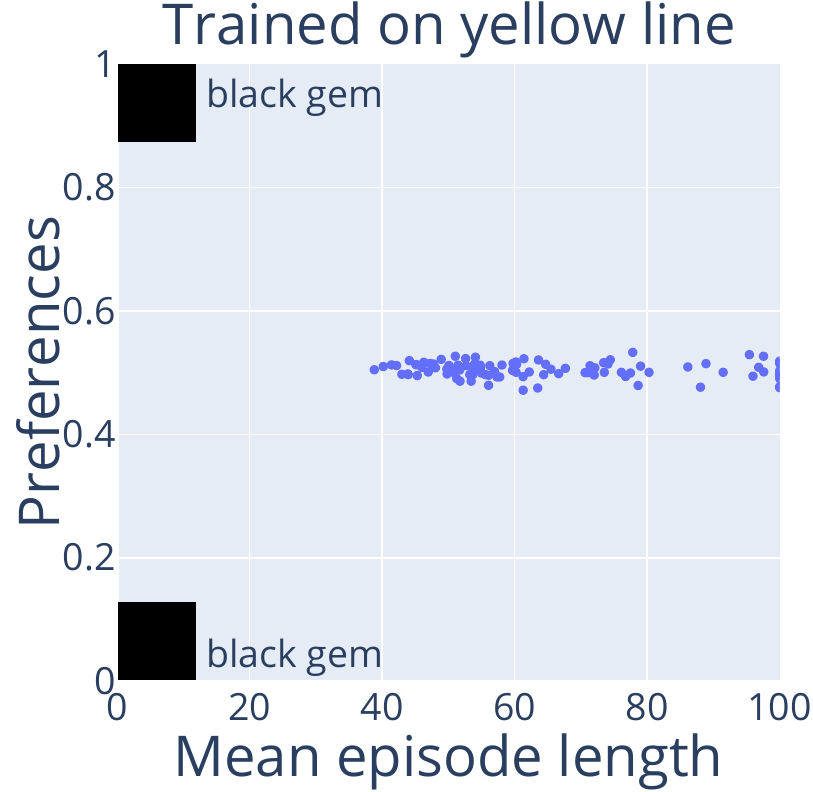}}
  \subfigure{\includegraphics[width=0.3\textwidth]{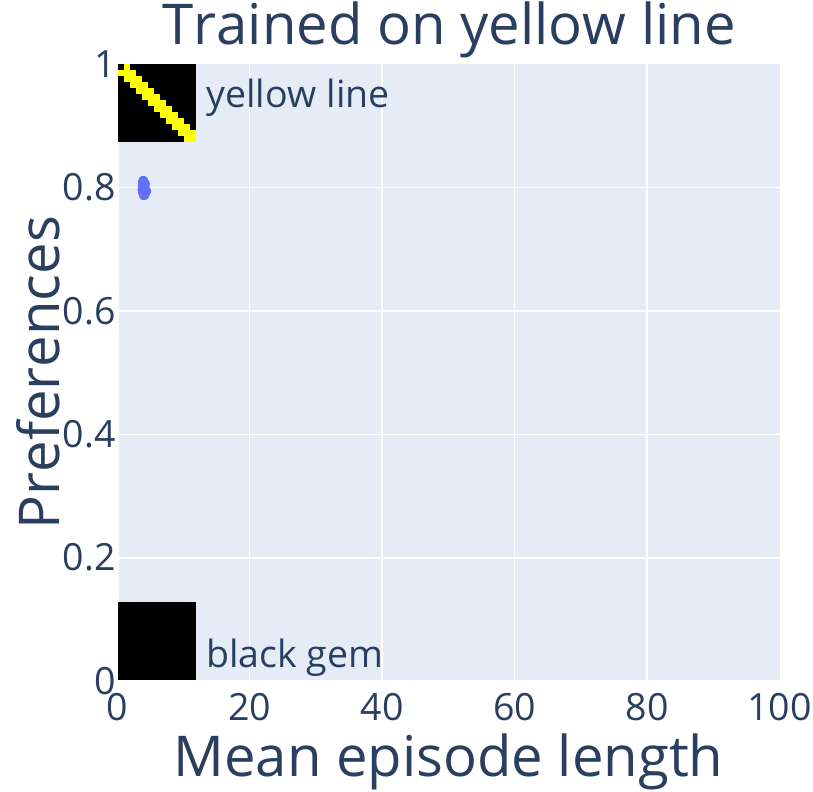}}
  \subfigure{\includegraphics[width=0.3\textwidth]{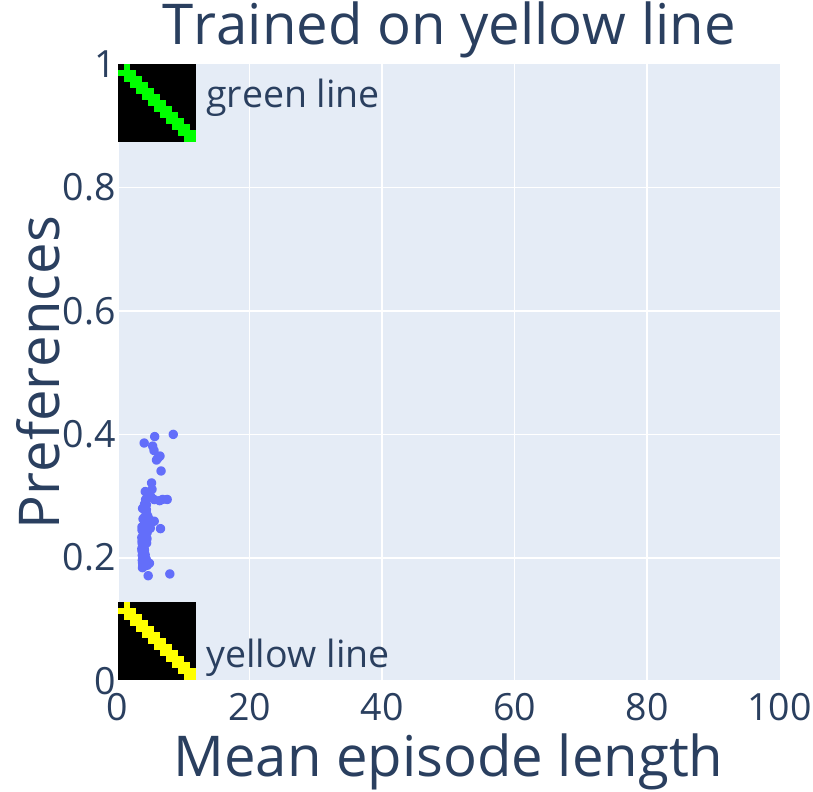}}
  
  \caption{Trained on the yellow line with black backgrounds, tested on different colour lines versus other different colour lines. Also, versus black gem, which is invisible, to test performance on a single yellow line and on a maze with no objects.}
  \label{fig:train-yellow-line-black-background-02}
\end{figure}

We include scatter plots of all these experiments in \autoref{appendix:all-results}.
We believe that interesting further insights can be gained by analysing the plots in detail.
For example, there are some agents that perform much better than random (38 steps on average instead of 96) when both objects are invisible (bottom left plot of \autoref{fig:train-yellow-line-black-background-02}) -- what algorithm could they be using to do that?

\clearpage
\section{All results}
\label{appendix:all-results}

Below we provide full results of all the simplified Maze agents we trained. Some plots are duplicates of plots presented in the main paper; we present them here for easier comparison. All scenarios had 100 agents, each tested on 1,000 episodes, except the agents trained on the white line with black backgrounds, of which there were 512.

\begin{figure}[htbp]
  \centering
  \subfigure{\includegraphics[width=0.3\textwidth]{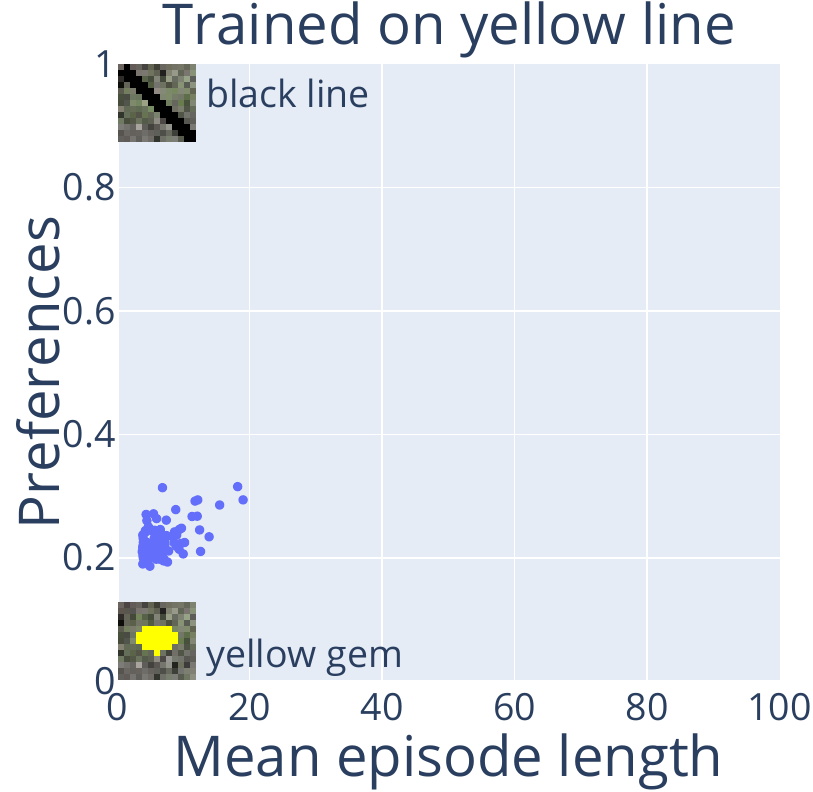}}
  \subfigure{\includegraphics[width=0.3\textwidth]{img/train-yellow-line-test-blue-line-yellow-gem-scatter.pdf}}
  \subfigure{\includegraphics[width=0.3\textwidth]{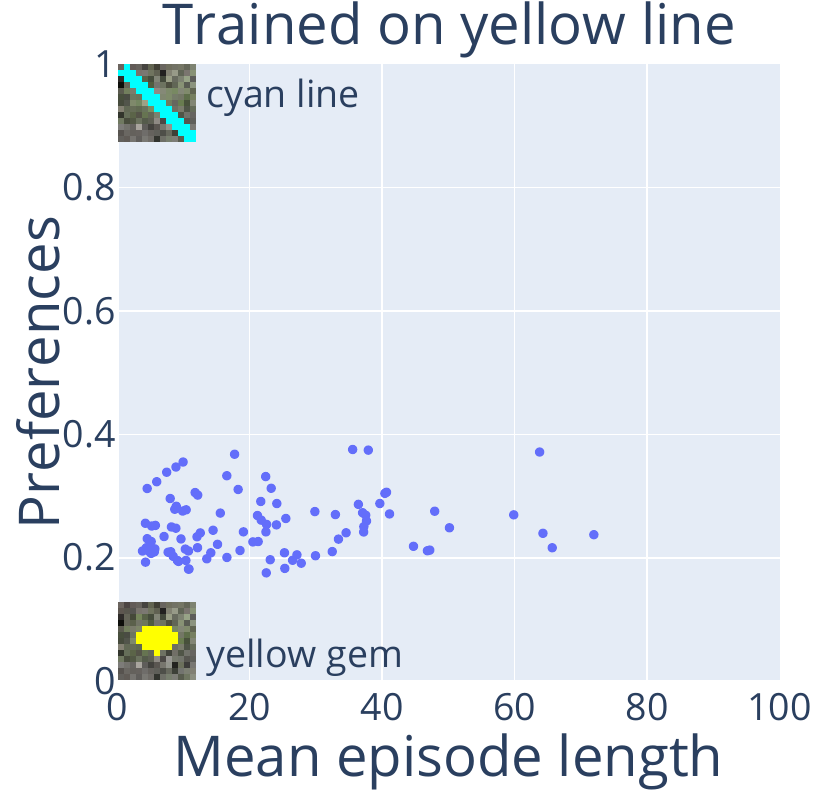}}
  
  \subfigure{\includegraphics[width=0.3\textwidth]{img/train-yellow-line-test-green-line-yellow-gem-scatter.pdf}}
  \subfigure{\includegraphics[width=0.3\textwidth]{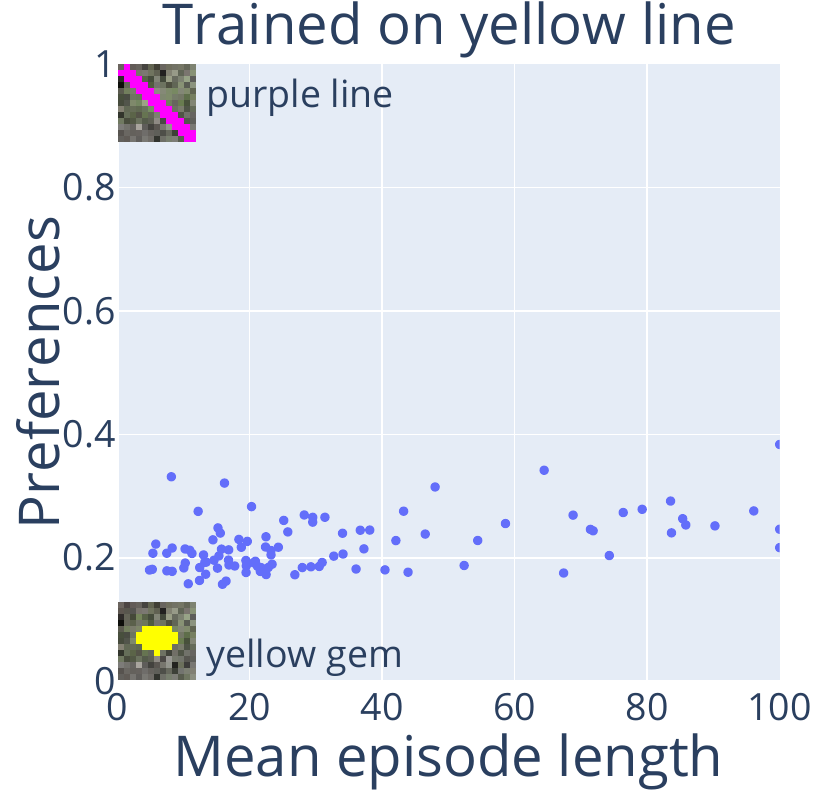}}
  \subfigure{\includegraphics[width=0.3\textwidth]{img/train-yellow-line-test-red-line-yellow-gem-scatter.pdf}}
  
  \subfigure{\includegraphics[width=0.3\textwidth]{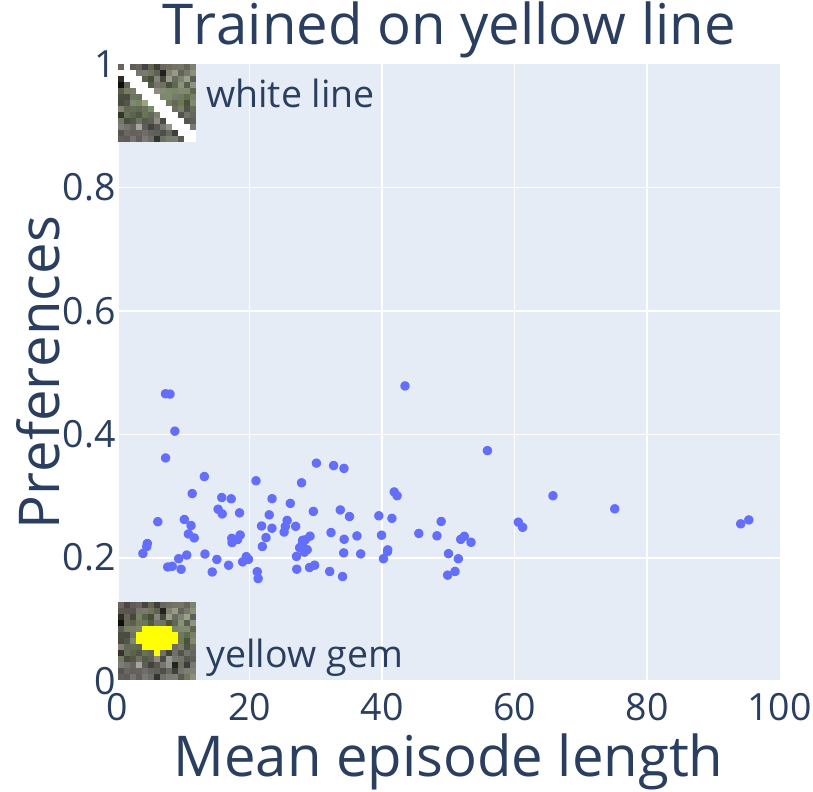}}
  \subfigure{\includegraphics[width=0.3\textwidth]{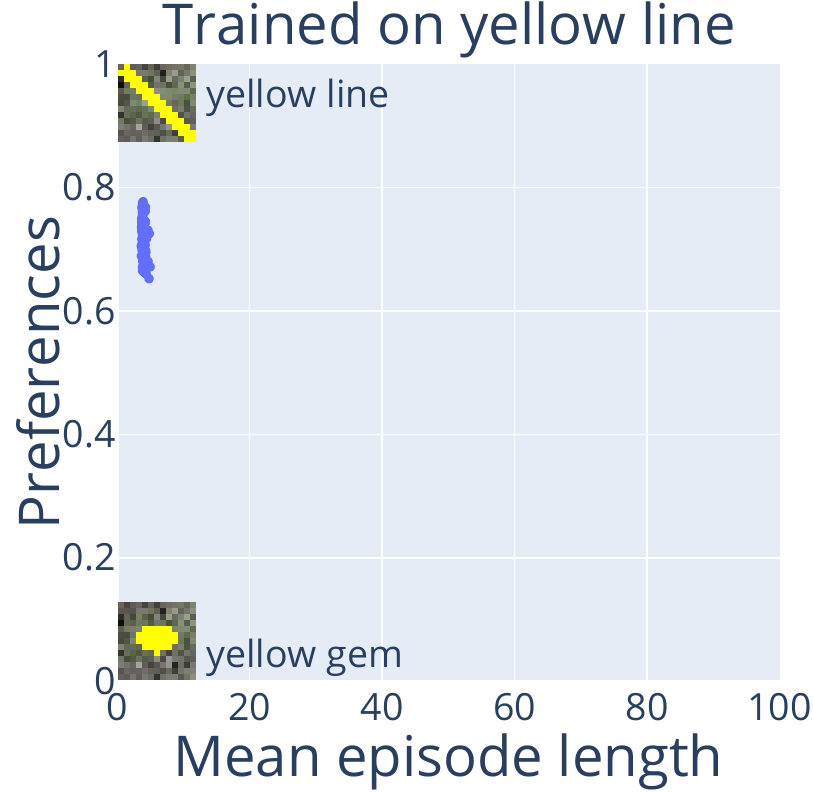}}
  \caption{Trained on the yellow line with multiple textured backgrounds, tested on different colour lines versus yellow gem.}
\end{figure}

\begin{figure}[htbp]
  \centering
  \subfigure{\includegraphics[width=0.3\textwidth]{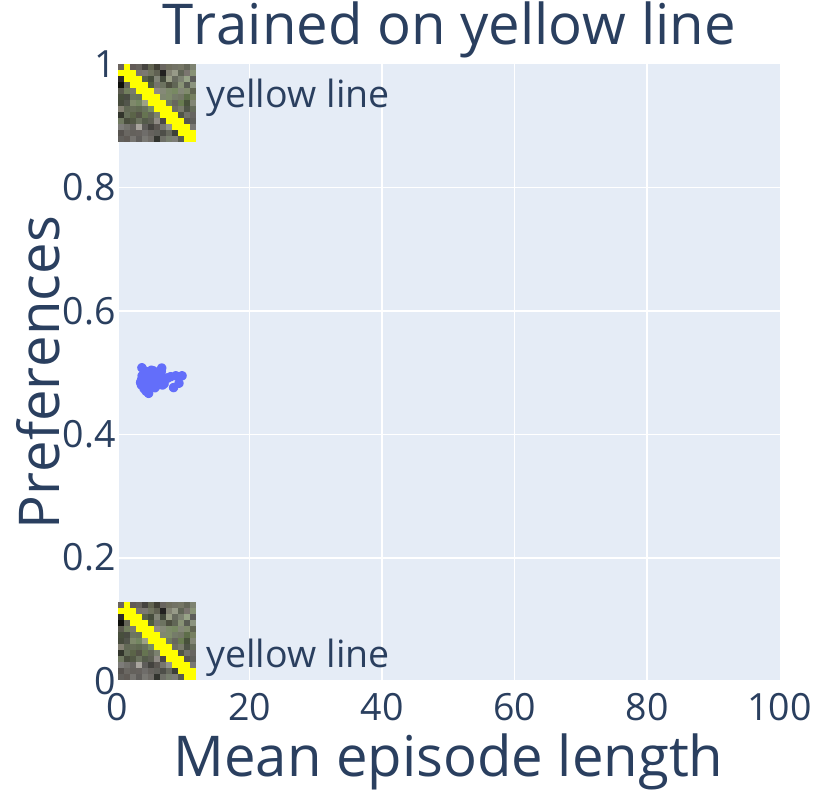}}
  \subfigure{\includegraphics[width=0.3\textwidth]{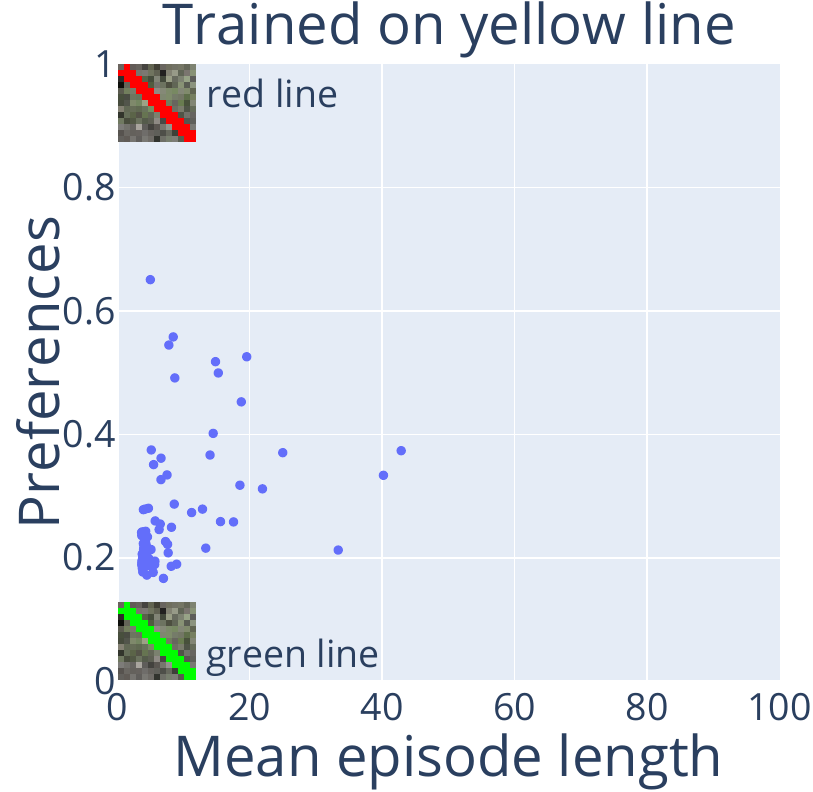}}
  \subfigure{\includegraphics[width=0.3\textwidth]{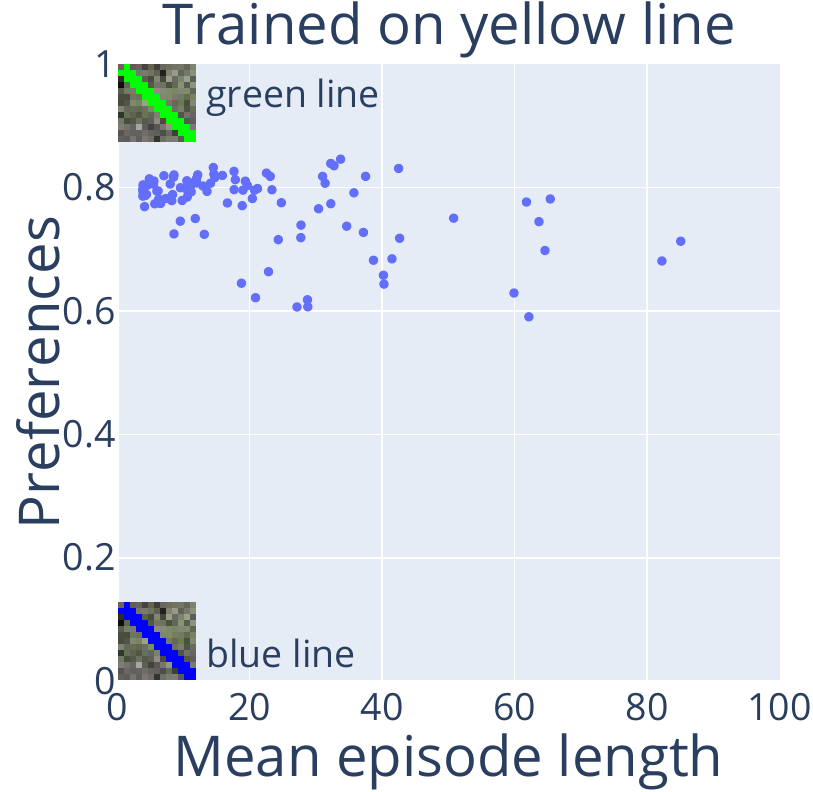}}
  
  \subfigure{\includegraphics[width=0.3\textwidth]{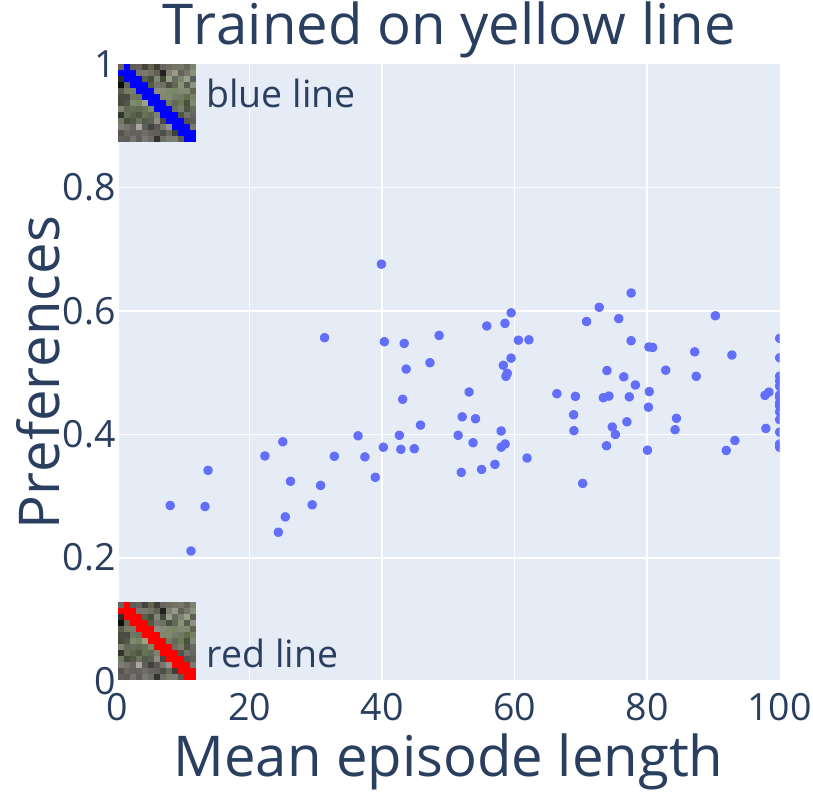}}
  \subfigure{\includegraphics[width=0.3\textwidth]{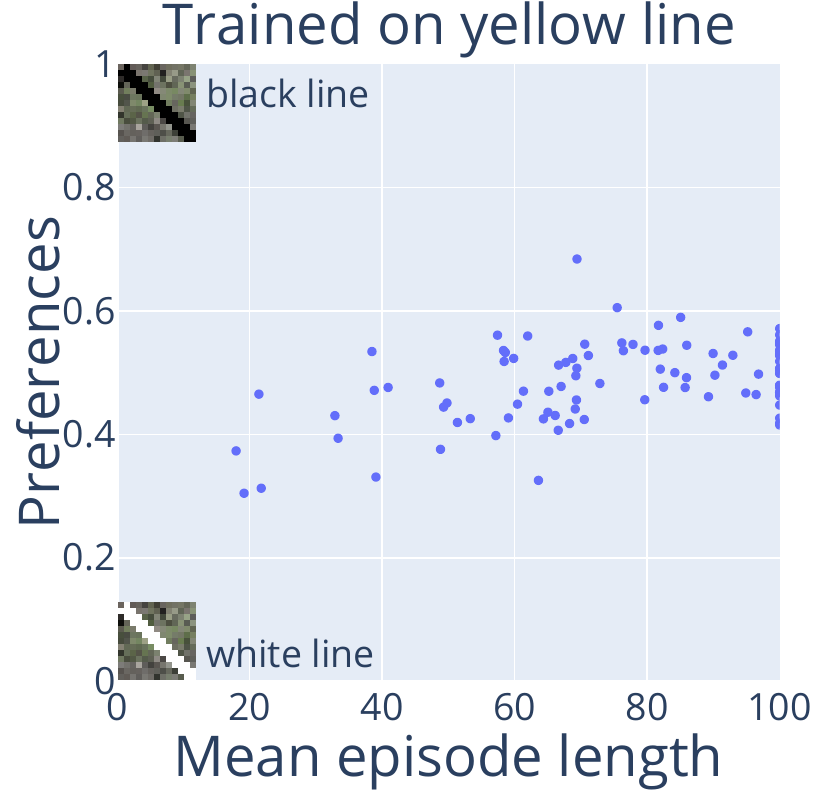}}
  \subfigure{\includegraphics[width=0.3\textwidth]{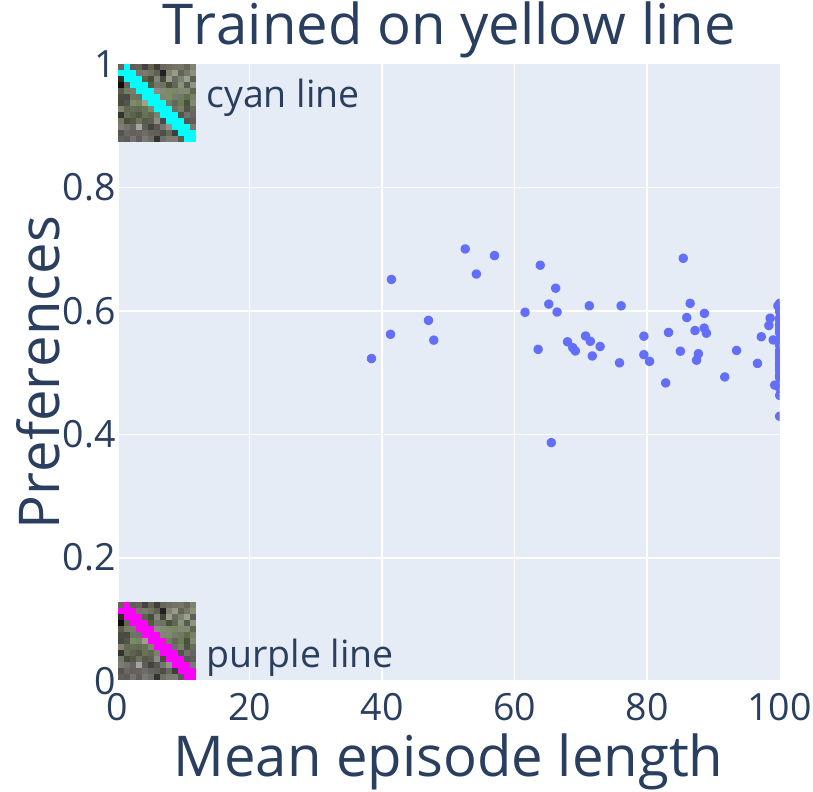}}
  
  \subfigure{\includegraphics[width=0.3\textwidth]{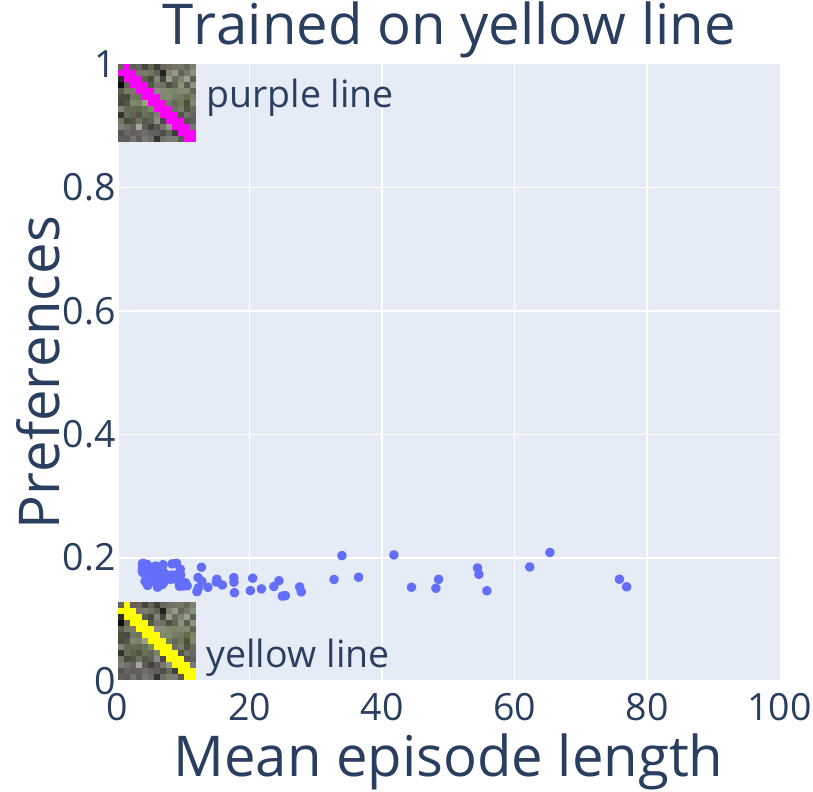}}
  \subfigure{\includegraphics[width=0.3\textwidth]{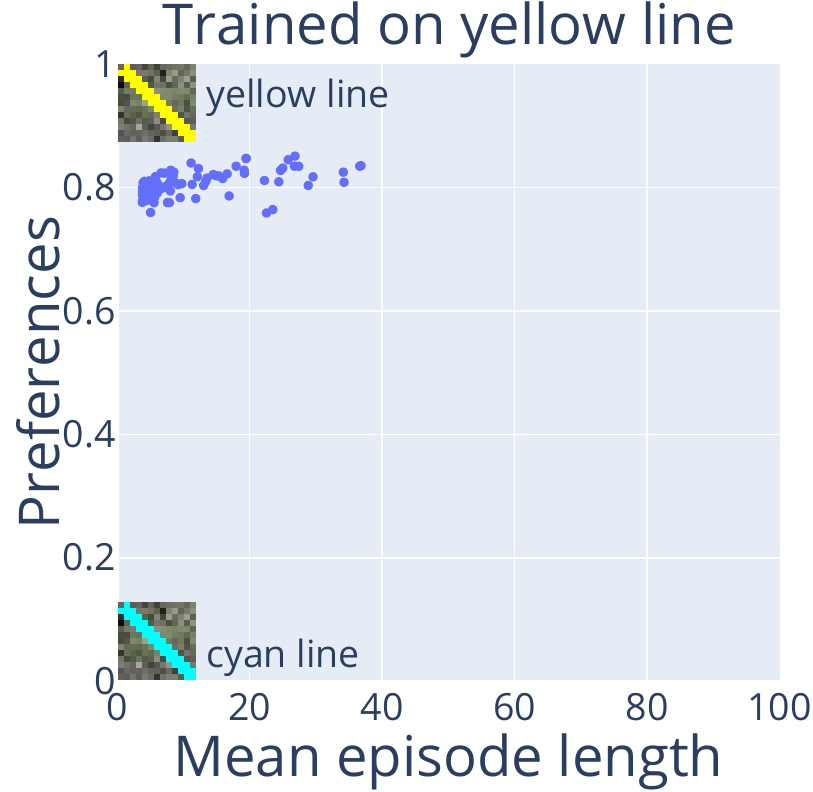}}
  \caption{Trained on the yellow line with multiple textured backgrounds, tested on different colour lines versus other different colour lines.}
\end{figure}

\begin{figure}[htbp]
  \centering
  \subfigure{\includegraphics[width=0.3\textwidth]{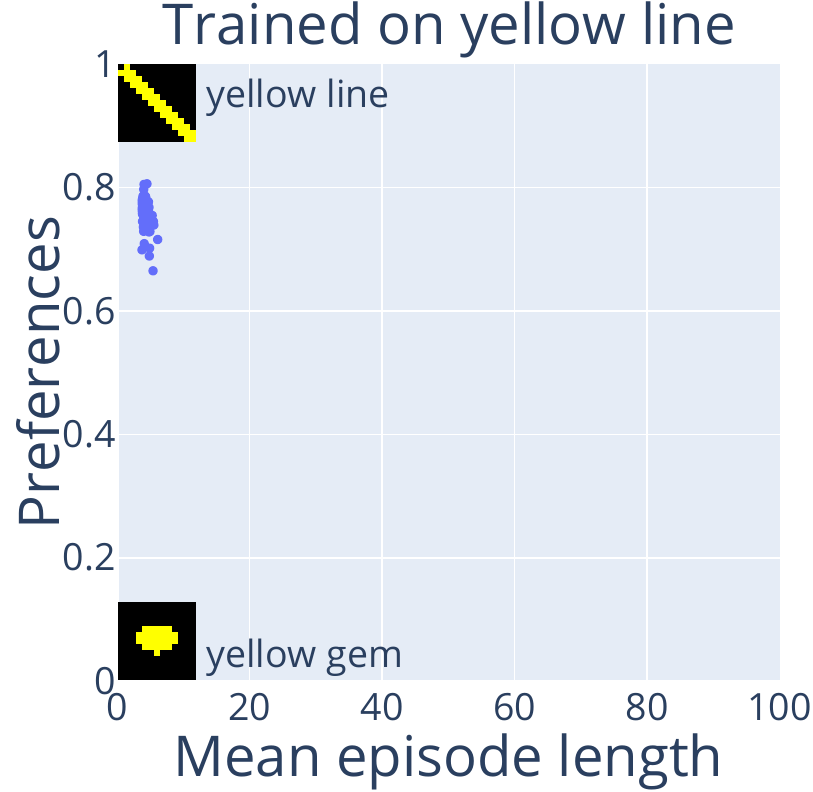}}
  \subfigure{\includegraphics[width=0.3\textwidth]{img/train-yellow-line-black-background-test-red-line-yellow-gem-scatter.pdf}}
  \subfigure{\includegraphics[width=0.3\textwidth]{img/train-yellow-line-black-background-test-green-line-yellow-gem-scatter.pdf}}
  
  \subfigure{\includegraphics[width=0.3\textwidth]{img/train-yellow-line-black-background-test-blue-line-yellow-gem-scatter.pdf}}
  \subfigure{\includegraphics[width=0.3\textwidth]{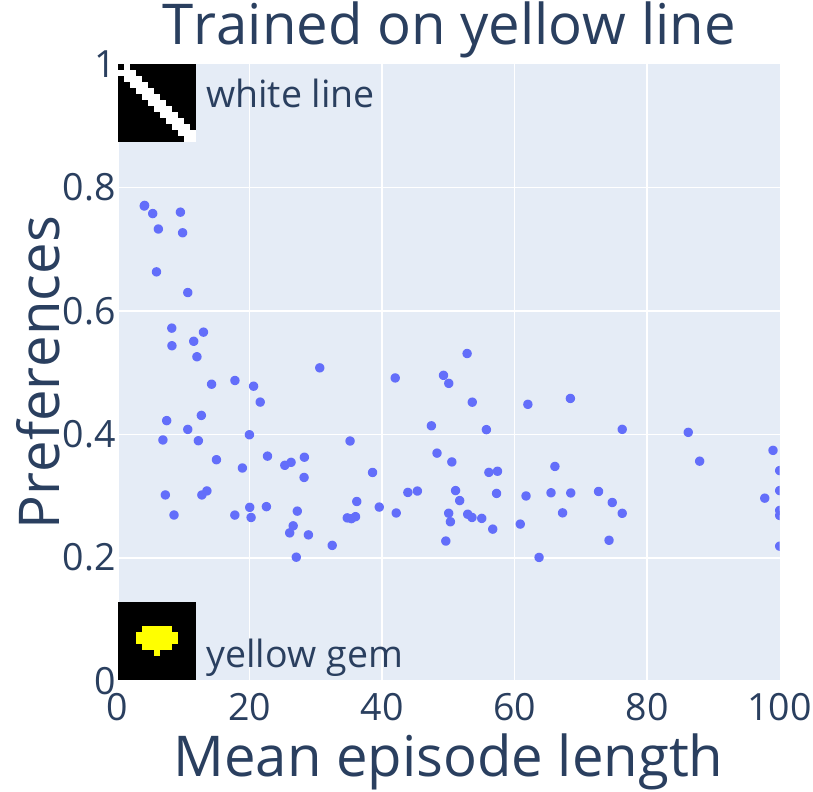}}
  \subfigure{\includegraphics[width=0.3\textwidth]{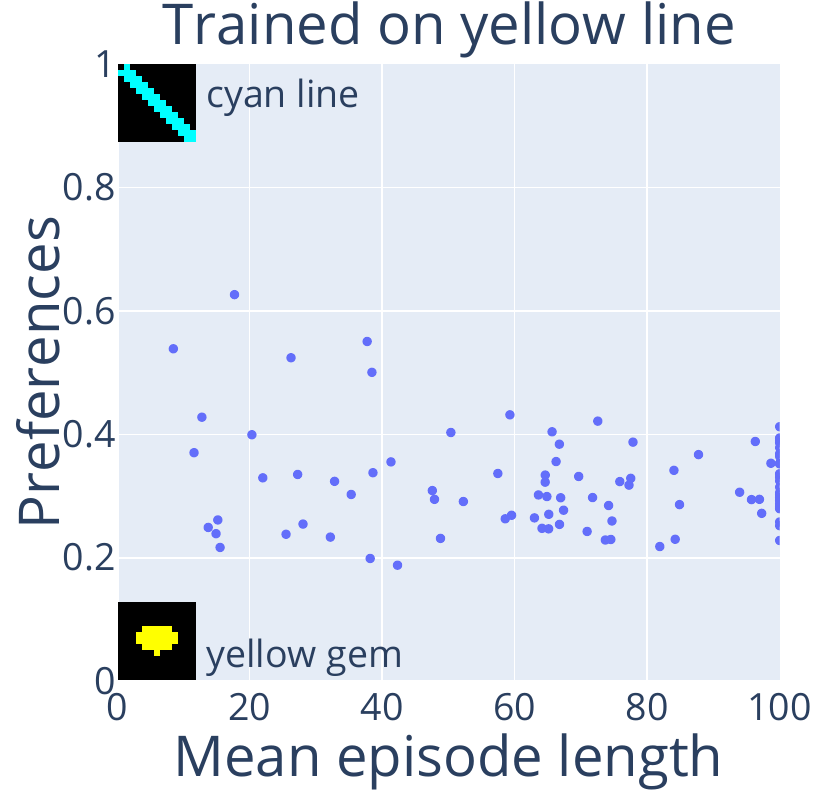}}
  
  \subfigure{\includegraphics[width=0.3\textwidth]{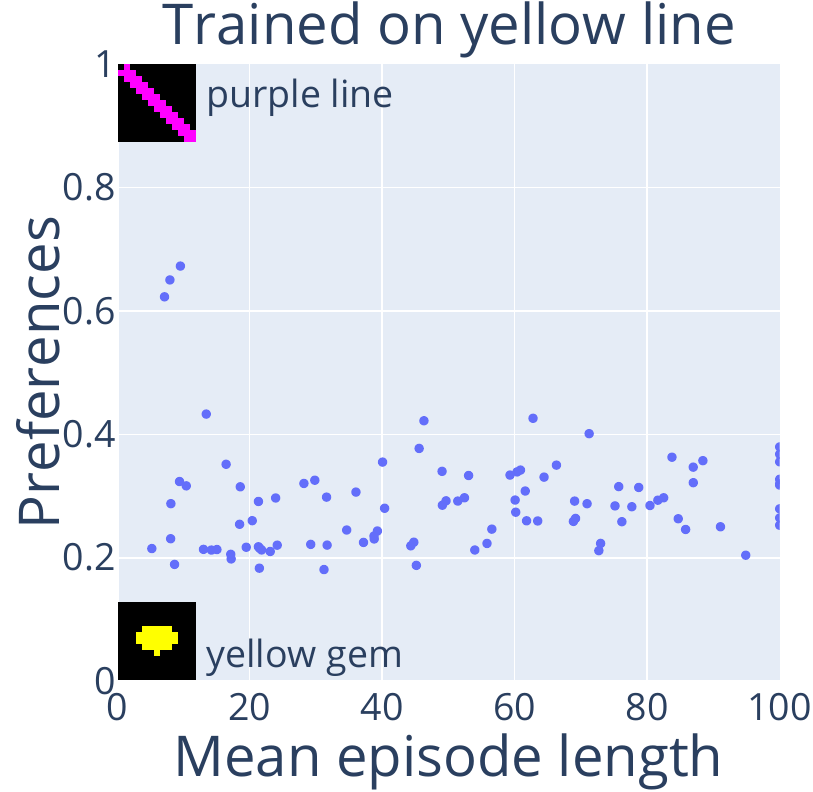}}
  \subfigure{\includegraphics[width=0.3\textwidth]{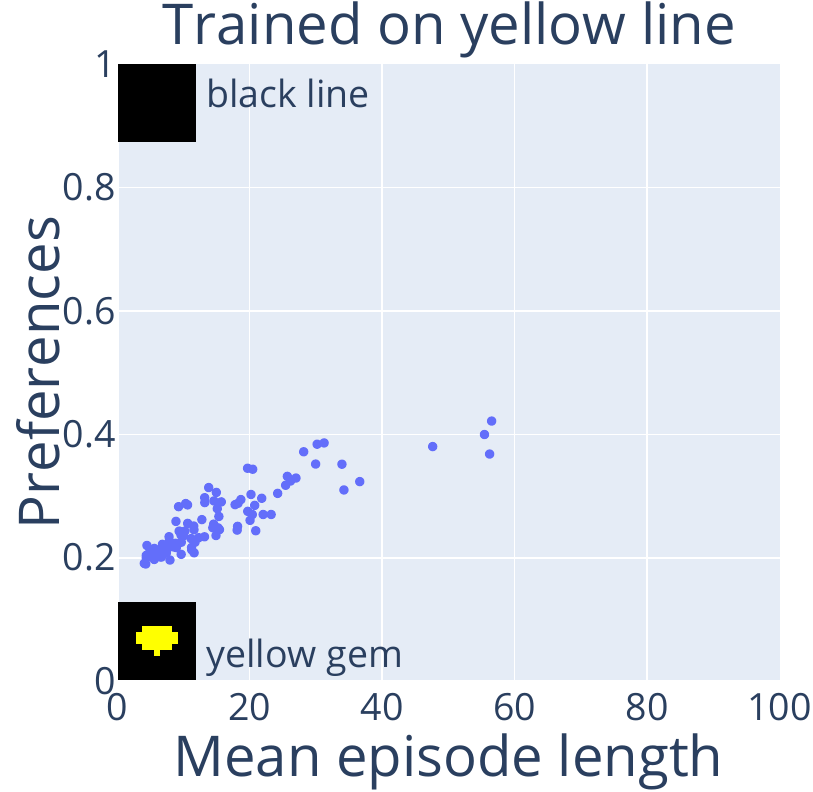}}
  \caption{Trained on the yellow line with black backgrounds, tested on different colour lines versus yellow gem.}
  \label{fig:train-yellow-line-black-background-01}
\end{figure}

\begin{figure}[htbp]
  \centering
  \subfigure{\includegraphics[width=0.3\textwidth]{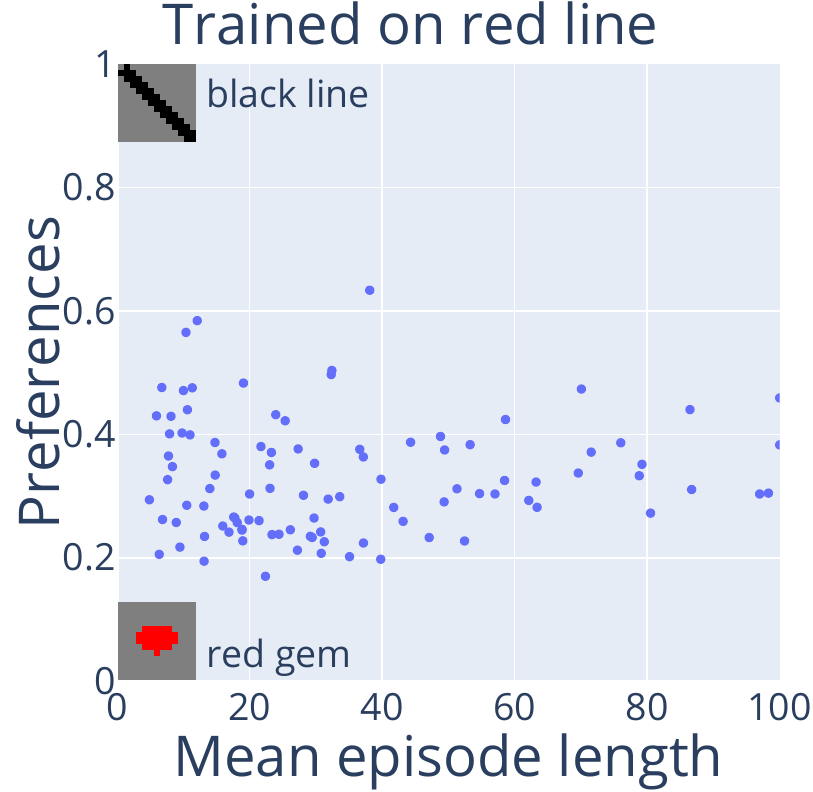}}
  \subfigure{\includegraphics[width=0.3\textwidth]{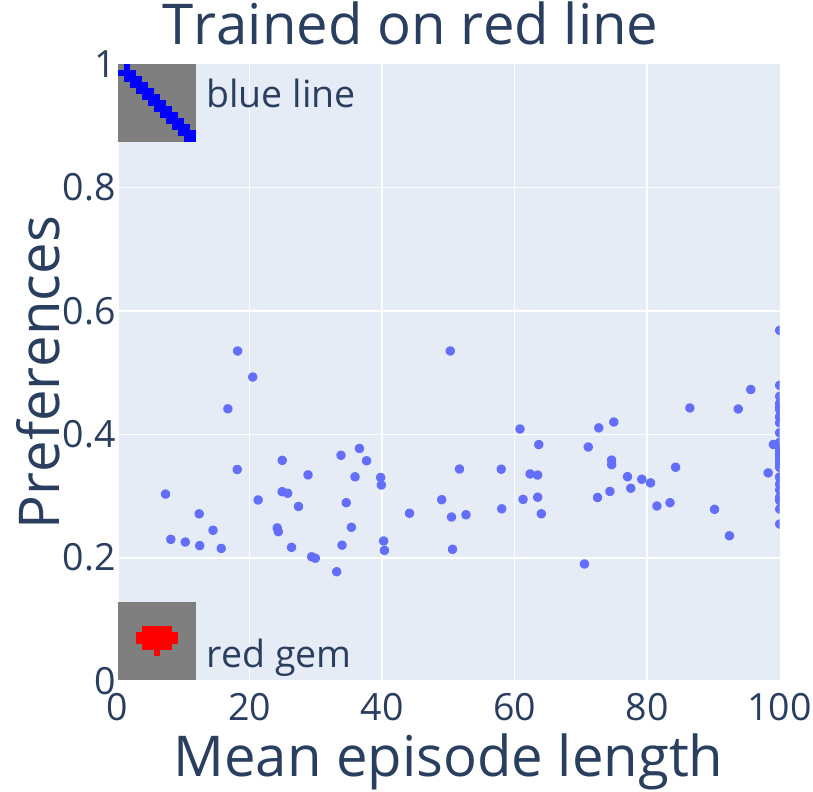}}
  \subfigure{\includegraphics[width=0.3\textwidth]{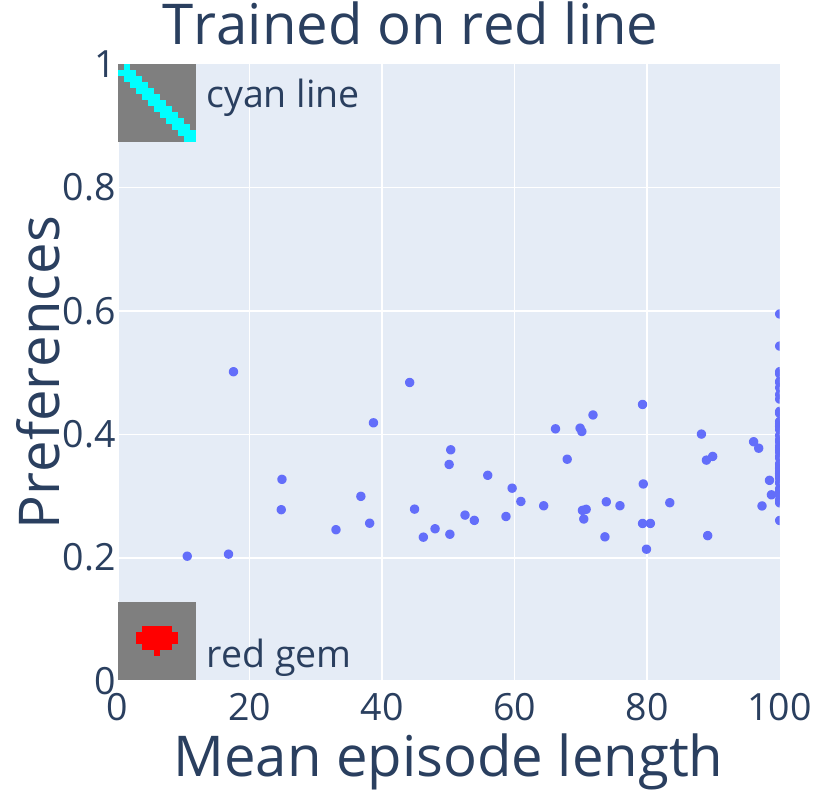}}
  
  \subfigure{\includegraphics[width=0.3\textwidth]{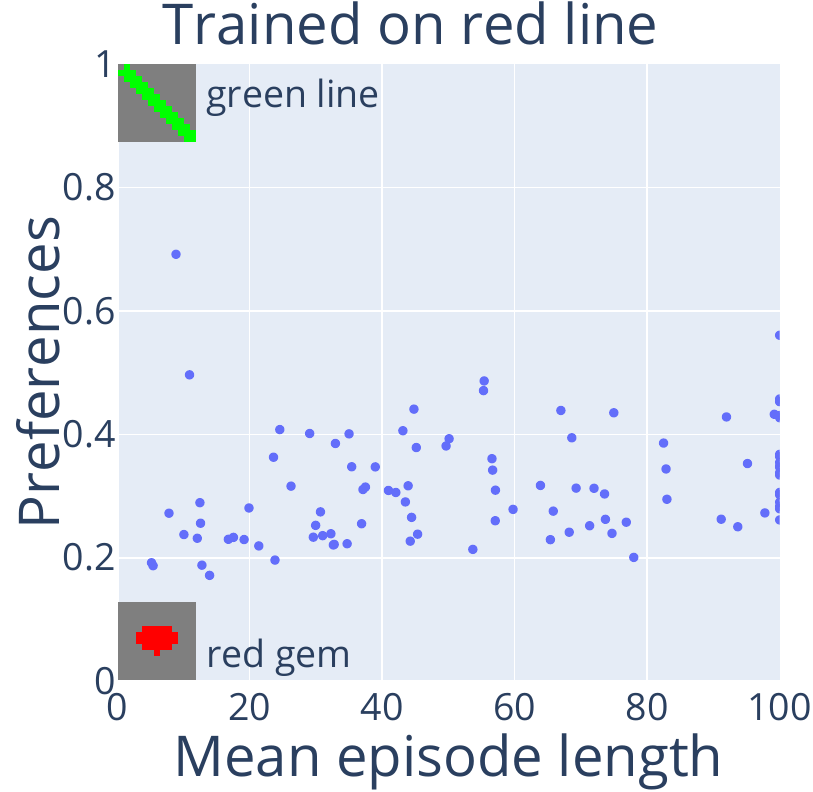}}
  \subfigure{\includegraphics[width=0.3\textwidth]{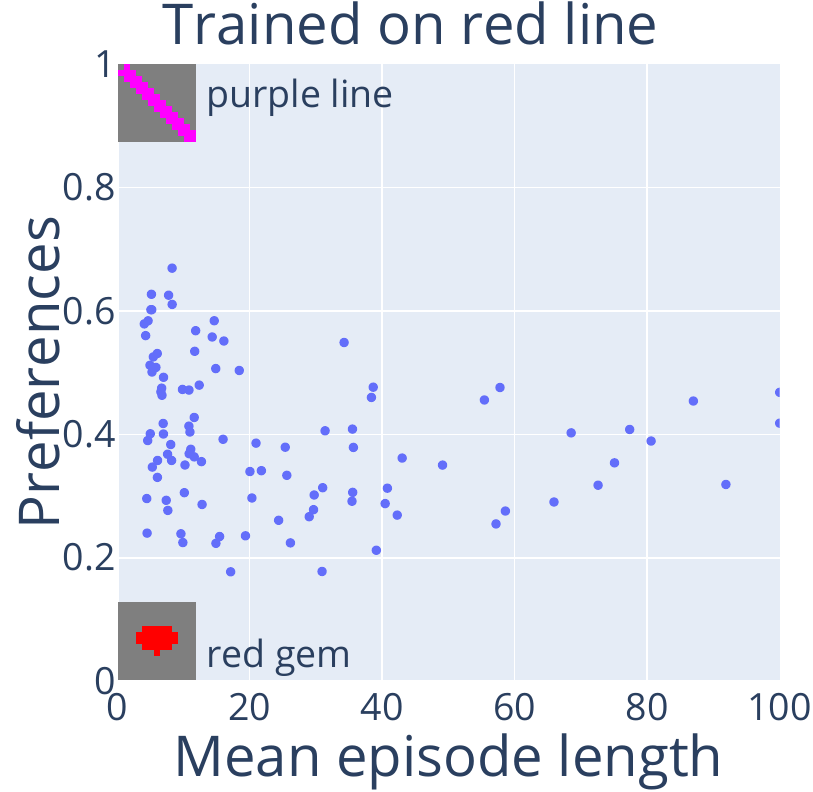}}
  \subfigure{\includegraphics[width=0.3\textwidth]{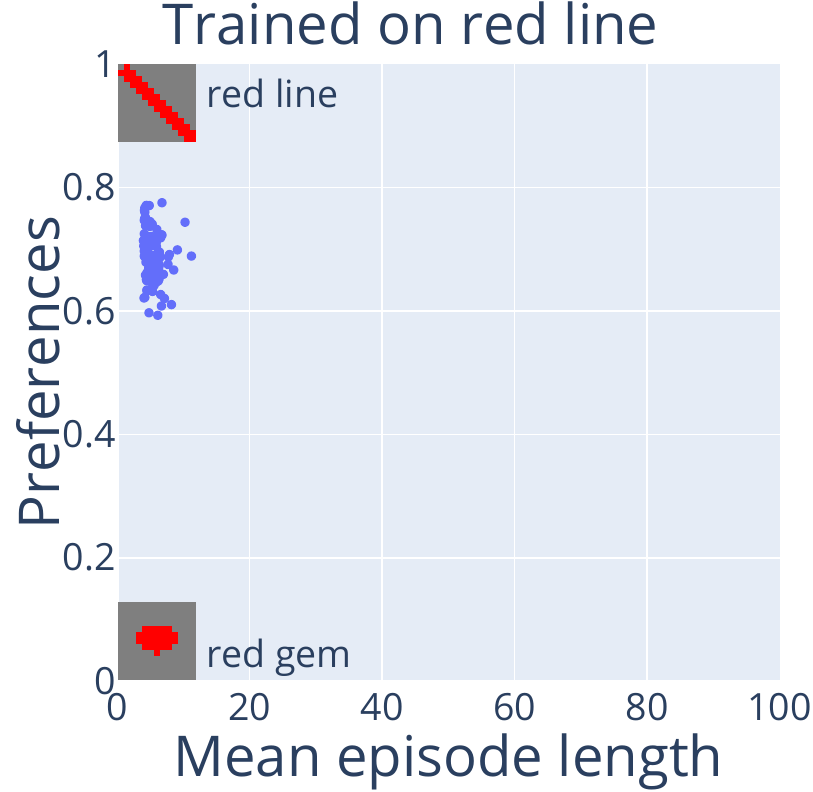}}
  
  \subfigure{\includegraphics[width=0.3\textwidth]{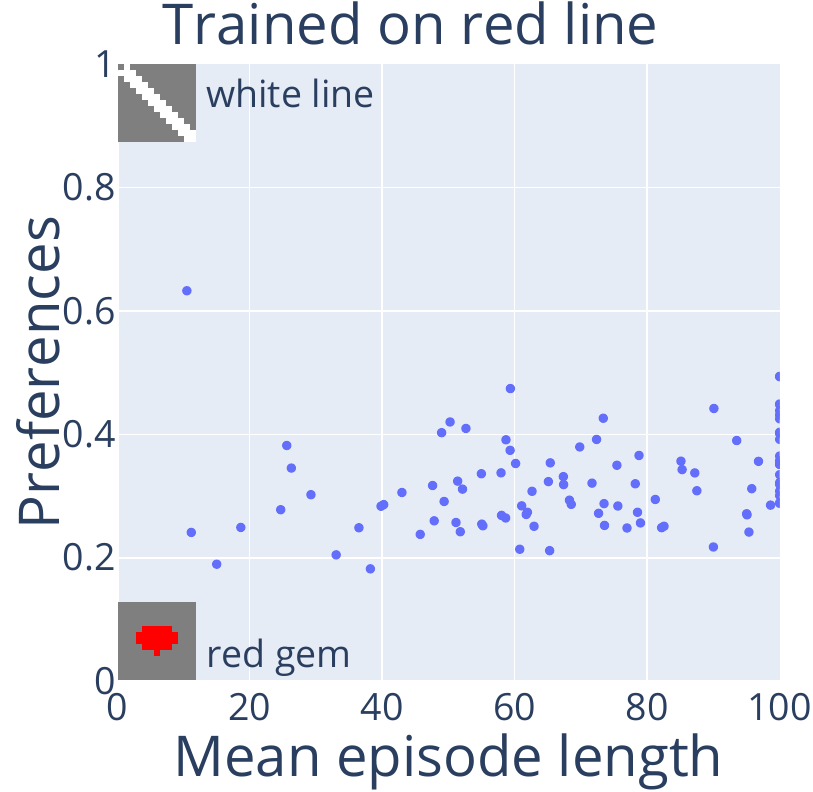}}
  \subfigure{\includegraphics[width=0.3\textwidth]{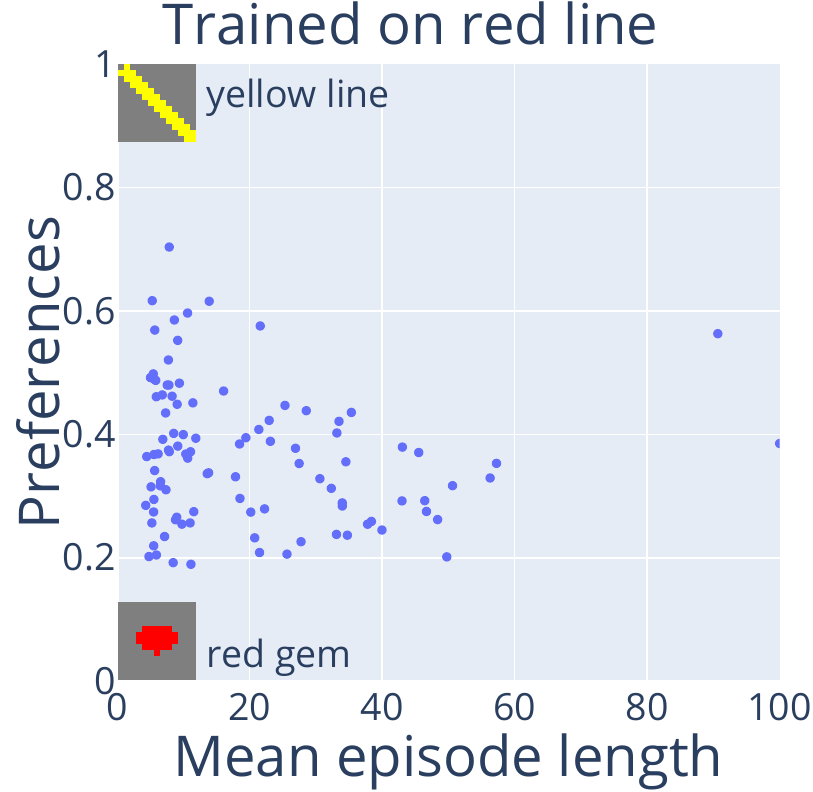}}
  \subfigure{\includegraphics[width=0.3\textwidth]{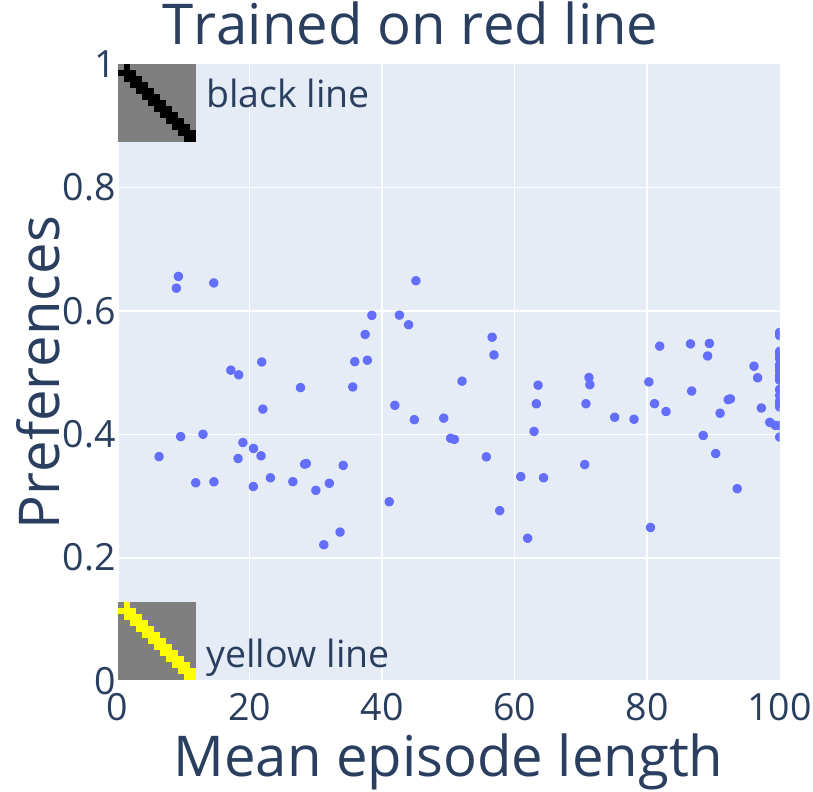}}
  
  \subfigure{\includegraphics[width=0.3\textwidth]{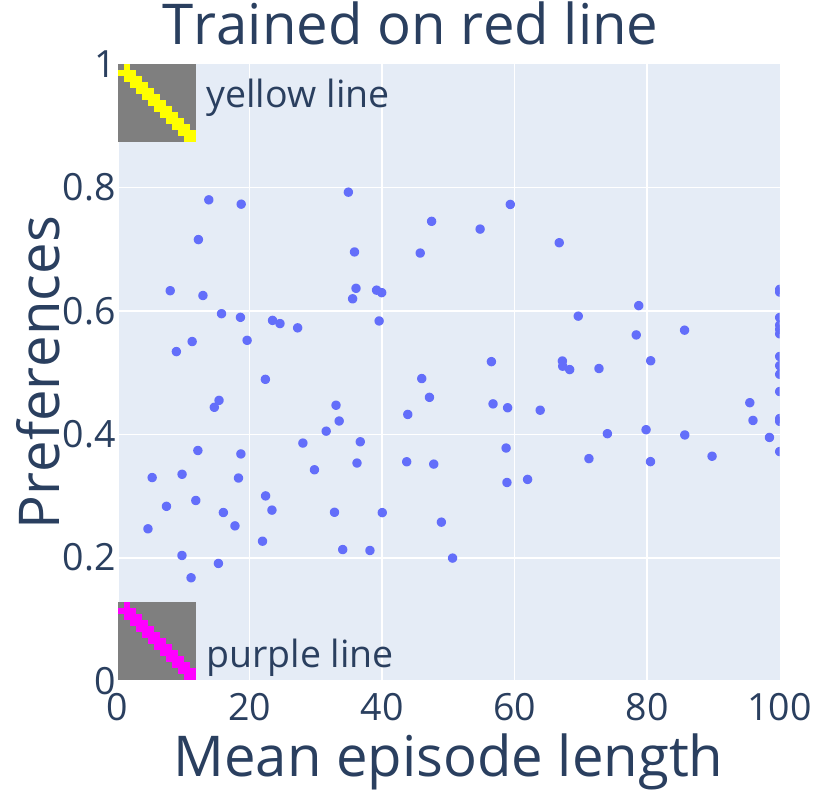}}
  \subfigure{\includegraphics[width=0.3\textwidth]{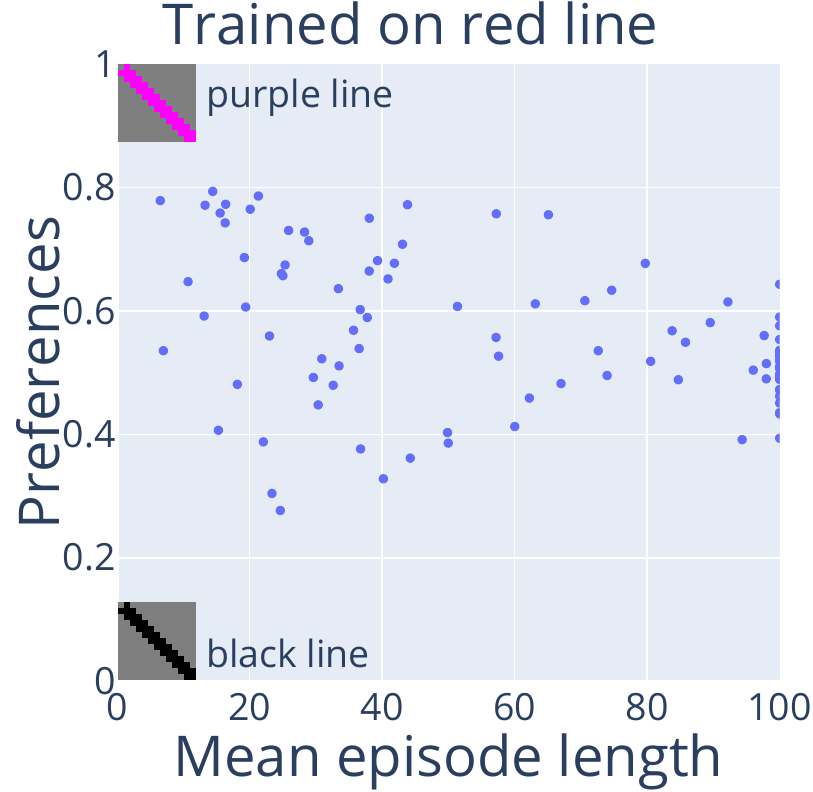}}
  
  \caption{Trained on the red line with grey backgrounds, tested on different colour lines versus red gem. Also, some line versus line tests.}
  \label{fig:train-red-line-grey-background}
\end{figure}

\begin{figure}[htbp]
  \centering
  \subfigure{\includegraphics[width=0.3\textwidth]{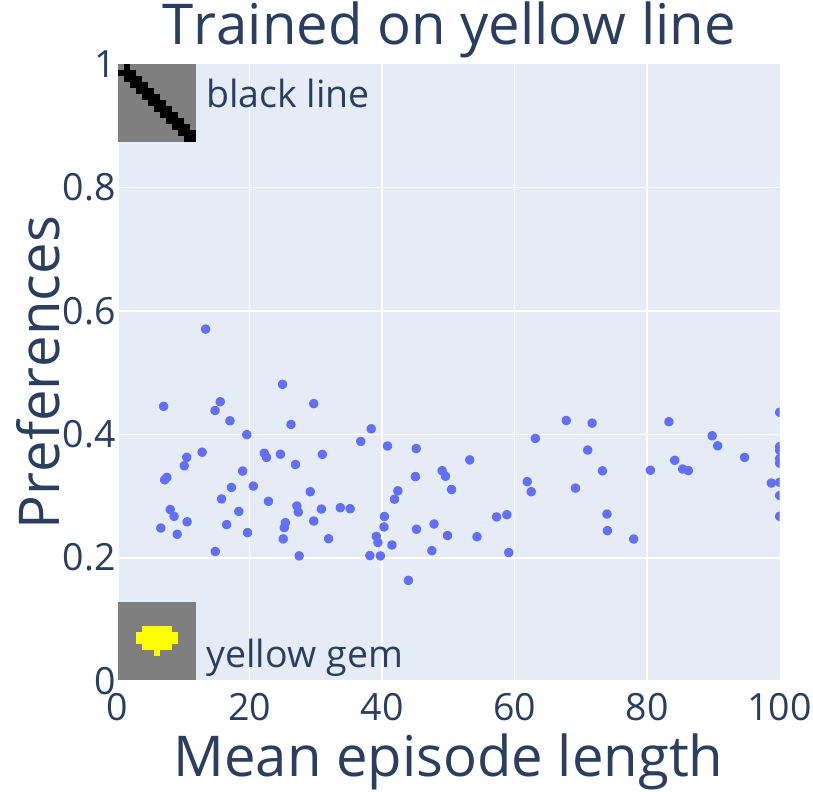}}
  \subfigure{\includegraphics[width=0.3\textwidth]{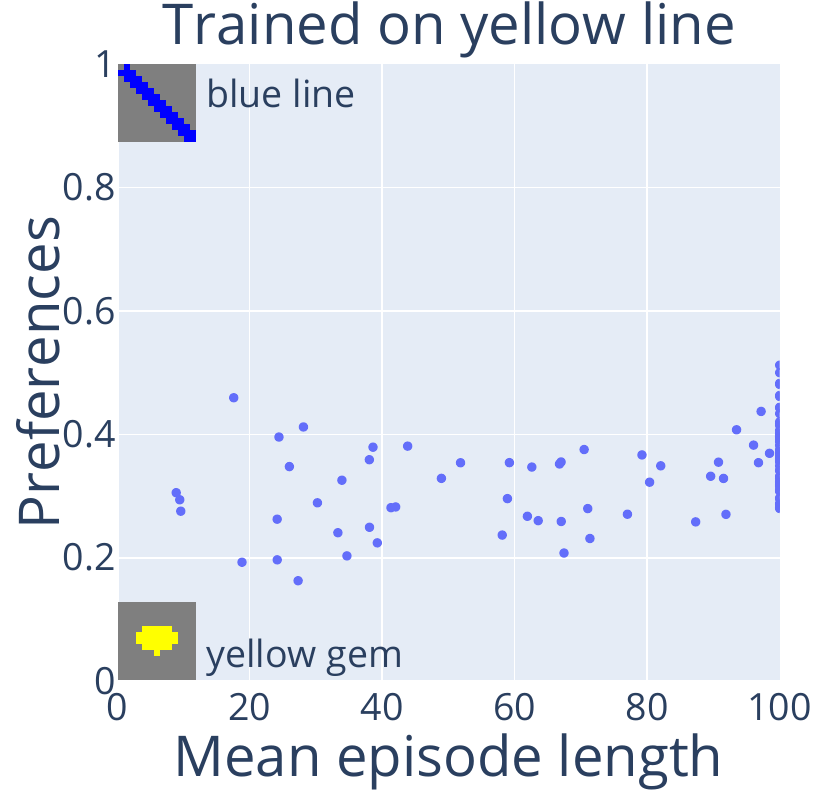}}
  \subfigure{\includegraphics[width=0.3\textwidth]{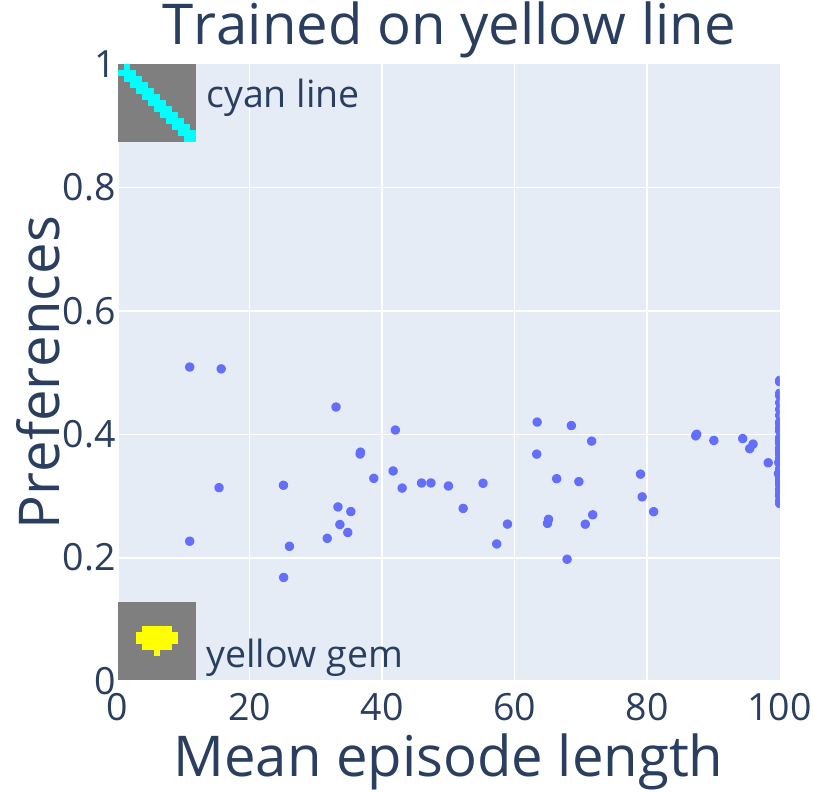}}
  
  \subfigure{\includegraphics[width=0.3\textwidth]{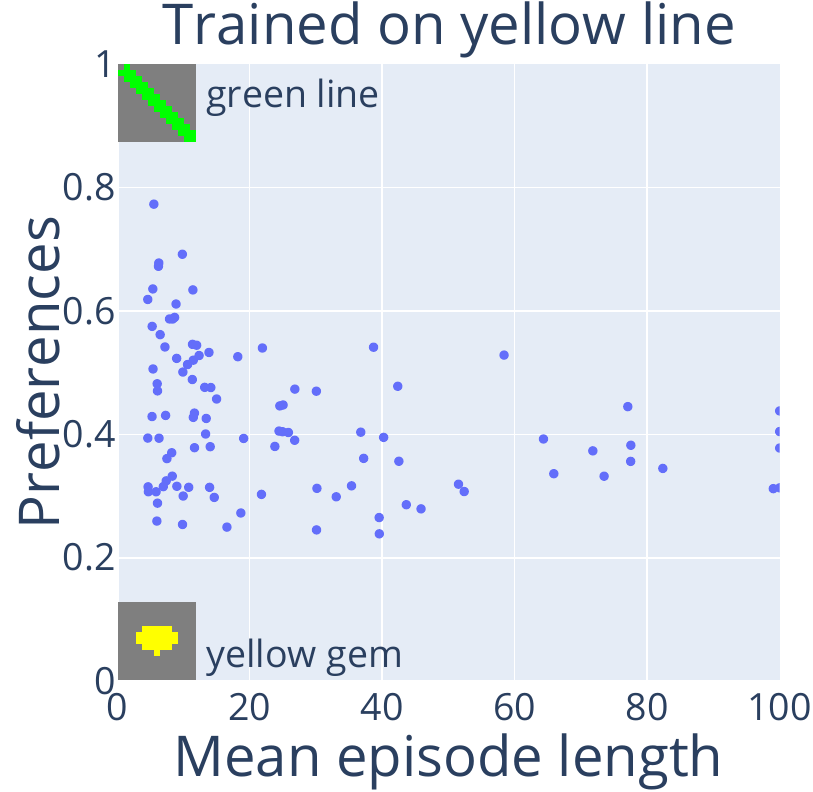}}
  \subfigure{\includegraphics[width=0.3\textwidth]{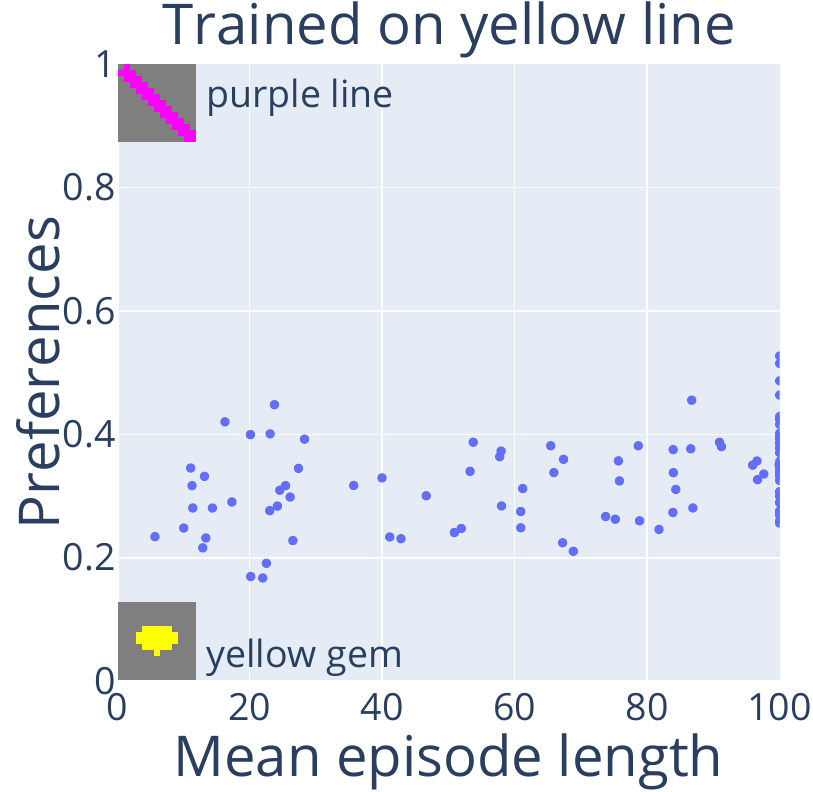}}
  \subfigure{\includegraphics[width=0.3\textwidth]{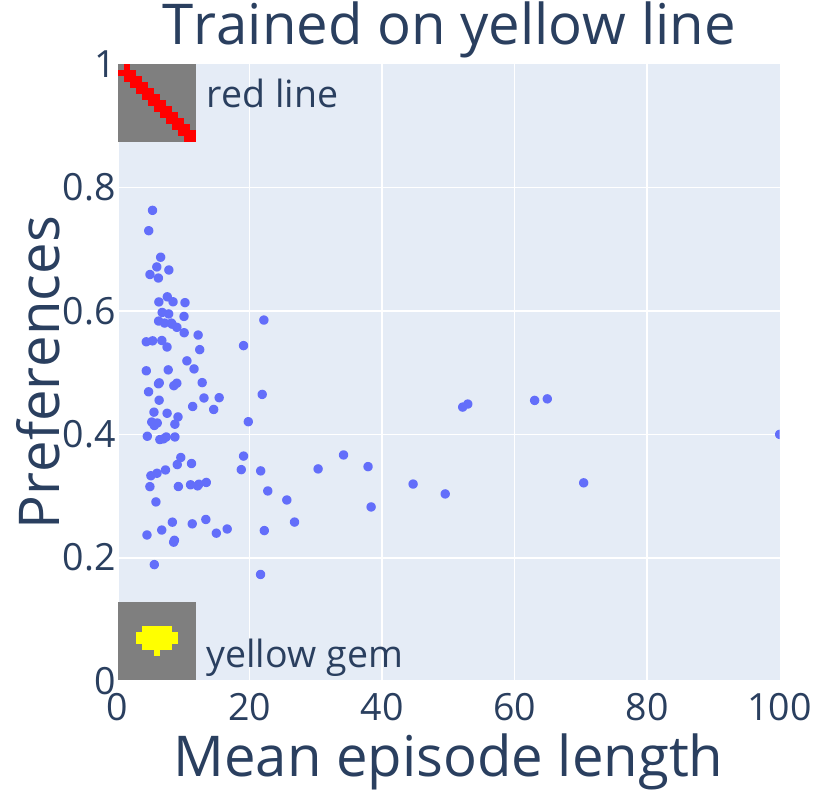}}
  
  \subfigure{\includegraphics[width=0.3\textwidth]{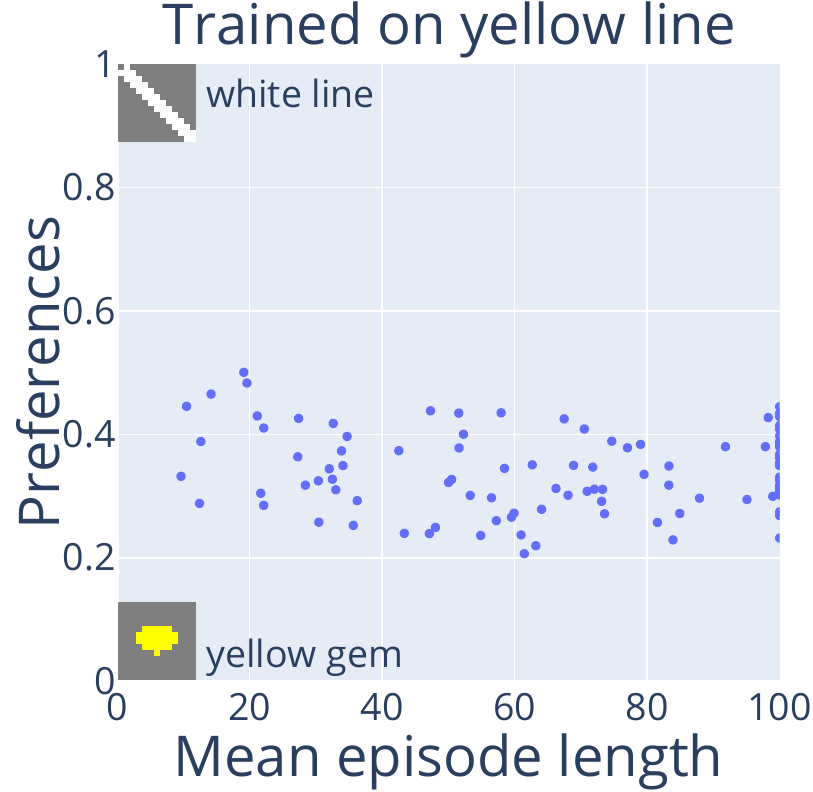}}
  \subfigure{\includegraphics[width=0.3\textwidth]{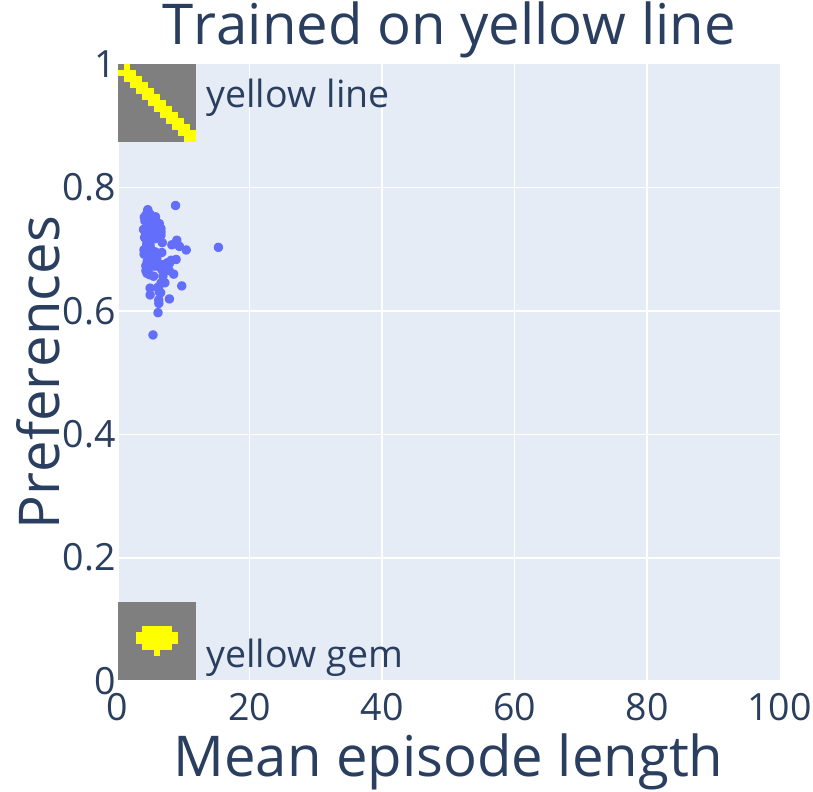}}
  \subfigure{\includegraphics[width=0.3\textwidth]{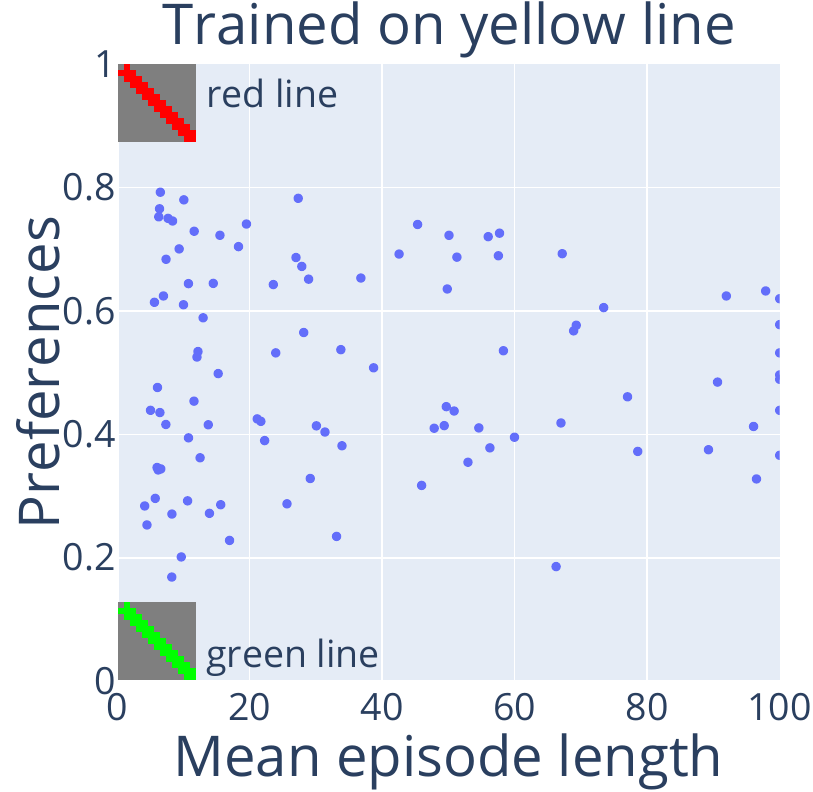}}
  
  \subfigure{\includegraphics[width=0.3\textwidth]{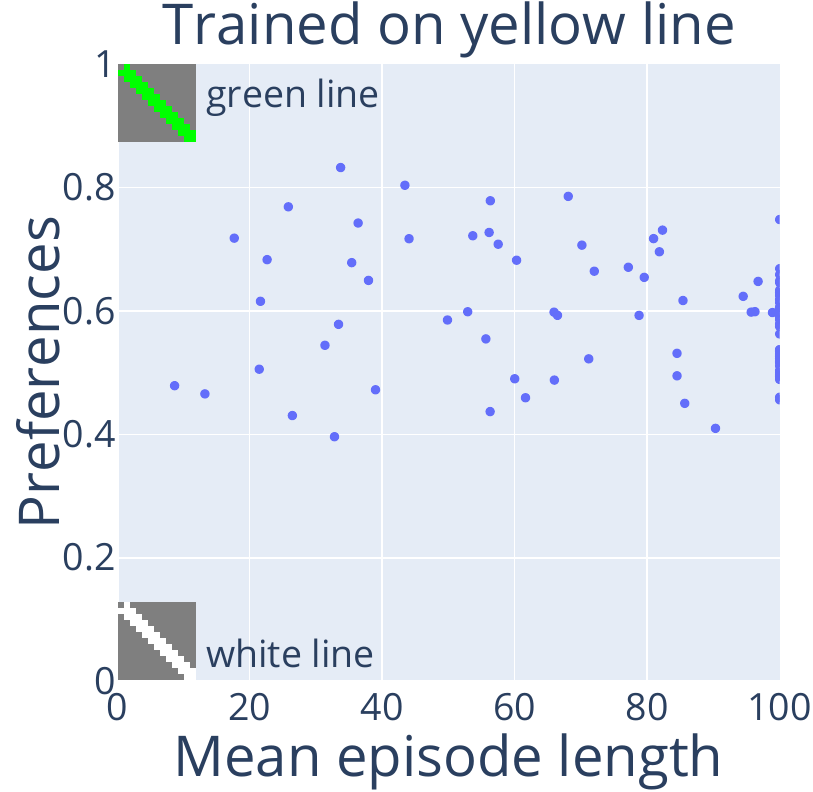}}
  \subfigure{\includegraphics[width=0.3\textwidth]{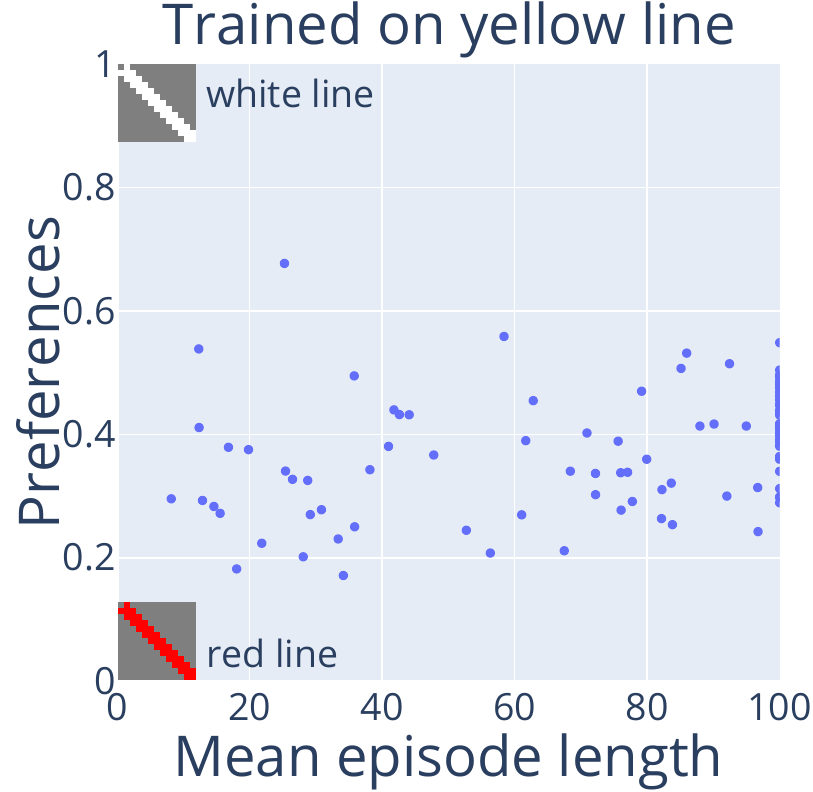}}
  
  \caption{Trained on the yellow line with grey backgrounds, tested on different colour lines versus yellow gem. Also, some line versus line tests.}
  \label{fig:train-yellow-line-grey-background}
\end{figure}

\end{document}